\documentclass{article}

\usepackage{microtype}
\usepackage{graphicx}
\usepackage{subfigure}
\usepackage{booktabs} %
\usepackage{multirow}
\usepackage{wrapfig} 
\usepackage{stfloats}
\usepackage{adjustbox}
\usepackage{caption}
\usepackage{url}
\usepackage{tablefootnote}
\usepackage{enumitem}

\usepackage{hyperref}

\usepackage[accepted]{icml2024}

\usepackage{amsmath}
\usepackage{amssymb}
\usepackage{mathtools}
\usepackage{amsthm}
\usepackage{subcaption}
\usepackage{pifont}

\usepackage{amsmath,amsfonts,bm,dsfont,mathtools}

\newcommand\independent{\protect\mathpalette{\protect\independenT}{\perp}}
\def\independenT#1#2{\mathrel{\rlap{$#1#2$}\mkern2mu{#1#2}}}

\def\<#1,#2>{\left\langle #1,#2 \right\rangle}

\newcommand{\norm}[1]{\|#1\|}

\newcommand{\abs}[1]{|#1|}

\def\eqref#1{equation~\ref{#1}}

\newcommand{\1}[1]{\mathds{1}_{#1}}

\def\rvz{{\mathbf{z}}}

\def\rmA{{\mathbf{A}}}

\def\vone{{\bm{1}}}

\def\vx{{\bm{x}}}

\def\vxi{{\bm{\xi}}}

\def\mA{{\bm{A}}}

\DeclareMathAlphabet{\mathsfit}{\encodingdefault}{\sfdefault}{m}{sl}
\SetMathAlphabet{\mathsfit}{bold}{\encodingdefault}{\sfdefault}{bx}{n}

\newcommand{\prob}{\mathbb{P}}
\newcommand{\E}[2][]{\mathbb{E}_{#1}\left[#2\right]} %

\newcommand{\R}{\mathbb{R}}

\newcommand{\var}[2][]{{ \operatorname{Var}_{#1}\left(#2\right)}}

\DeclareMathOperator*{\argmax}{arg\,max}
\DeclareMathOperator*{\argmin}{arg\,min}

\usepackage[capitalize,noabbrev]{cleveref}

\theoremstyle{plain}
\newtheorem{theorem}{Theorem}[section]
\newtheorem{proposition}[theorem]{Proposition}
\newtheorem{lemma}[theorem]{Lemma}

\theoremstyle{definition}
\newtheorem{definition}[theorem]{Definition}
\newtheorem{assumption}[theorem]{Assumption}
\newtheorem{example}[theorem]{Example}
\theoremstyle{remark}
\newtheorem{remark}[theorem]{Remark}

\crefname{assumption}{Assumption}{Assumptions}
\crefname{proposition}{Proposition}{Propositions}
\crefname{example}{Example}{Examples}
\crefname{remark}{Remark}{Remarks}
\crefname{claim}{Claim}{Claims}
\crefname{conjecture}{Conjecture}{Conjectures}

\newcommand{\cred}[1]{{\color{red}#1}}
\newcommand{\cblue}[1]{{\color{blue}#1}}

\newcommand{\green}[1]{\textcolor[rgb]{0.00,0.50,0.00}{#1}}

\newcommand{\xmark}{\text{\cred{\ding{55}}}}
\newcommand{\NA}{\text{\cred{N/A}}}

\newcommand{\cmark}{\text{\green{\ding{51}}}}
\newcommand{\ok}{\text{\textcolor{yellow}{\ding{108}}}}

\usepackage{xspace}

\newcommand{\HyperAgent}{\hyperref[alg:hyperagent]{\normalfont\texttt{HyperAgent}}\xspace}
\newcommand{\update}{\hyperref[alg:update-dqn]{\normalfont\texttt{update}}\xspace}

\newcommand{\NpS}{\abs{\tilde{\Xi}}}

\hypersetup{
    colorlinks = true,
    linkcolor = {black},
    citecolor = {black}
}

\icmltitlerunning{\HyperAgent - Approximate Posterior Sampling over Q-Star: Simple, Scalable, Efficient}

\begin{document}

\twocolumn[
    \icmltitle{Q-Star Meets Scalable Posterior Sampling: \\Bridging Theory and Practice via HyperAgent}

    \icmlsetsymbol{equal}{*}

    \begin{icmlauthorlist}
        \icmlauthor{Yingru Li}{cuhksz,sribd}
        \icmlauthor{Jiawei Xu}{cuhksz}
        \icmlauthor{Lei Han}{tencent}
        \icmlauthor{Zhi-Quan Luo}{cuhksz,sribd}
    \end{icmlauthorlist}

    \icmlaffiliation{cuhksz}{The Chinese University of Hong Kong, Shenzhen}
    \icmlaffiliation{sribd}{Shenzhen Research Institue of Big Data}
    \icmlaffiliation{tencent}{Tencent AI and Robotics X}

    \icmlcorrespondingauthor{Yingru Li}{\url{szrlee@gmail.com} or \url{yingruli@link.cuhk.edu.cn}}

    \icmlkeywords{Machine Learning, ICML}

    \vskip 0.3in
]

\printAffiliationsAndNotice{}  %

\begin{abstract}
    We propose HyperAgent, a reinforcement learning (RL) algorithm based on the hypermodel framework for exploration in RL. HyperAgent allows for the efficient incremental approximation of  posteriors associated with an optimal action-value function ($Q^\star$) without the need for conjugacy and follows the greedy policies w.r.t. these approximate posterior samples.
    We demonstrate that HyperAgent offers robust performance in large-scale deep RL benchmarks. It can solve Deep Sea hard exploration problems with episodes that optimally scale with problem size and exhibits significant efficiency gains in the Atari suite. Implementing HyperAgent requires minimal code addition to well-established deep RL frameworks like DQN.
    We theoretically prove that, under tabular assumptions, HyperAgent achieves logarithmic per-step computational complexity while attaining sublinear regret, matching the best known randomized tabular RL algorithm.
\end{abstract}

\section{Introduction}
\label{sec:intro}

Practical reinforcement learning (RL) in complex environments faces challenges such as large state spaces and an increasingly large volume of data. The per-step computational complexity, defined as the computational cost for the agent to make a decision at each interaction step, is crucial. Under resource constraints, any reasonable design of an RL agent must ensure bounded per-step computation, a key requirement for \emph{scalability}. If per-step computation scales polynomially with the volume of accumulated interaction data, computational requirements will soon become unsustainable, which is untenable for scalability. \emph{Data efficiency} in sequential decision-making demands that the agent learns the optimal policy with as few interaction steps as possible, a fundamental challenge given the need to balance exploration of the environment to gather more information and exploitation of existing information~\citep{thompson1933likelihood,lai1985asymptotically,thrun1992efficient}. Scalability and efficiency are both critical for the practical deployment of RL algorithms in real-world applications with limited resources. There appears to be a divergence between the development of practical RL algorithms, which mainly focus on scalability and computational efficiency, and RL theory, which prioritizes data efficiency, to our knowledge. This divergence raises an important question:
\begin{center}
    \emph{Can we design a practically efficient RL agent with provable guarantees on efficiency and scalability?}
\end{center}
\subsection{Key Contributions}
\begin{figure}[htbp]
    \centering
        \includegraphics[width=0.8\linewidth]{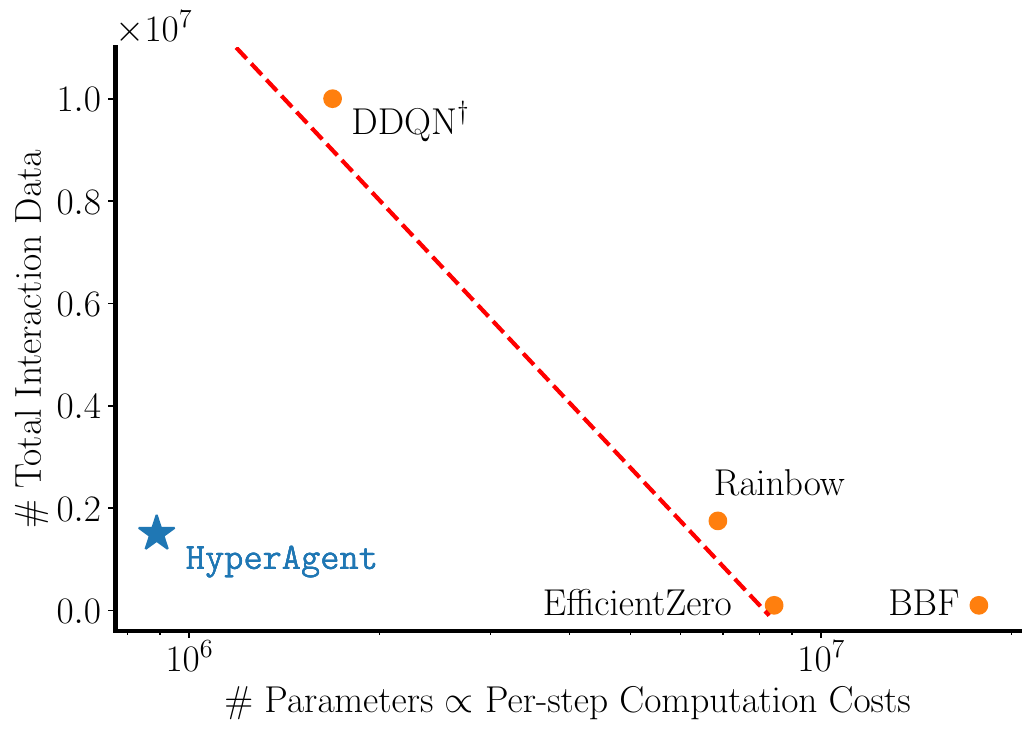}
        \caption{
            This evaluation explores the relationship between the amount of training data required and the model parameters necessary to achieve human-level performance, quantified by a 1.0 IQM score. It is assessed across 26 Atari games using the Interquartile Mean (IQM) metric~\citep{agarwal2021deep} using recent state-of-the-art (SOTA) algorithms.
            The number of parameters is directly proportional to the computational cost, as they predominantly influence the calculation during each SGD update per interaction step.
            \HyperAgent, denoted by {\Large \cblue{$\star$}}, achieves a 1.0 IQM score with a comparatively minimal number of interactions and parameters.
        }
        \label{fig:data-para}
        \vspace{-1.5em}
\end{figure}

This work affirmatively answers the posed question by proposing a novel reinforcement learning algorithm, \HyperAgent, based on the hypermodel framework \citep{li2022hyperdqn,dwaracherla2020hypermodels,osband2023epistemic}.
We highlight the advantages of \HyperAgent below:
\begin{itemize}[leftmargin=*]
    \item \emph{Algorithmic Simplicity}. \HyperAgent's implementation\footnote{We provide the open-source code at \url{https://github.com/szrlee/HyperAgent}.} requires only the addition of a single module, the \textit{last-layer linear hypermodel}, to the conventional DDQN framework and a minor modification for action selection. The added module facilitates efficient incremental approximation and sampling for the posteriors associated with the $Q^\star$ function. 
    Practically, \HyperAgent can be a replacement for the $\varepsilon$-greedy method in most of its usecases.
    This simplicity contrasts with state-of-the-art methods for the Atari benchmarks, which often rely on multiple complex algorithmic components and extensive tuning.
    \item \emph{Practical Efficiency}.
    The \HyperAgent algorithm demonstrates exceptional scalability and efficiency in challenging environments. Notably, in the Deep Sea exploration scenario \citep{osband2019behaviour}, it efficiently solves problems up to $120 \times 120$ in size with optimal episode complexity, as shown in \cref{fig:deepsea_baseline}. Furthermore, \HyperAgent achieves human-level performance on the Atari benchmark suite \citep{bellemare2013arcade}, as detailed in \cref{fig:data-para}, requiring only 15\% of the interaction data ($1.5$M interactions) and 5\% of the network parameters necessary for DDQN$^{\dagger}$ (Double Deep Q-Networks, \citet{van2016deep}) and BBF (Bigger, Better, Faster, \citet{schwarzer2023bigger}), respectively. In contrast, Ensemble+~\citep{osband2019deep}, a randomized exploration method, achieves a mere 0.22 IQM score with $1.5$M interactions but uses double the parameters of our approach.
    \item \emph{Provable Guarantees}.
    We prove that \HyperAgent achieves sublinear regret in a tabular, episodic setting with $\tilde{O}(\log K)$ per-step computation over $K$ episodes. This performance is supported by an incremental posterior approximation argument central to our analysis (\cref{lem:approx}). This argument is proved by a reduction to the sequential random projection~\citep{li2024probability}.
\end{itemize}
\HyperAgent effectively bridges the theoretical and practical aspects of RL in complex, large-scale environments. See \cref{tab:hyperagent-rl} for representative milestones of the RL algorithms.

\begin{table*}[htbp]
\centering
\centering
    \resizebox{0.7\linewidth}{!}{%
        \begin{tabular}{@{}l|ccc|cc@{}}
        \toprule[1.2pt]
         & \multicolumn{3}{c|}{\textbf{Practice} in Deep RL} & \multicolumn{2}{c}{%
           \textbf{Theory} in Tabular RL  }                                                                                                            \\ \midrule
            \textbf{Algorithm}                                  & \textbf{Tractable}                                    & \textbf{Incremental}                                                                                    & \textbf{Efficient}  & \textbf{Regret}
            & \textbf{Per-step Computation}                                \\ 
            \midrule \midrule
            PSRL              & $\xmark$                                  & $\xmark$                                                                                  & $\xmark$ & $\tilde{O}(H^2 \sqrt{SAK}) $   &  $O(S^2A)$                                     \\ %
            RLSVI                 & $\cmark$                                  & $\xmark$                                                                                  & $\xmark$ & $\tilde{O} (H^2 \sqrt{SAK}) $  & $O(S^2A)$                                     \\ %

            Ensemble+              & $\cmark$                                  & $\cmark$                                                                                  & $\ok$    & \NA                                     & \NA                                                 \\
            {{\HyperAgent}}                   & $\cmark$                                  & $\cmark$                                                                                  & $\cmark$ & $\tilde{O}(H^2\sqrt{SAK})$     & 
            $\tilde{O}({\log (K)}{SA} + S^2 A)$ 
            \\ \bottomrule
        \end{tabular}%
    }
    \caption[Efficiency and Scalability Comparison of RL Algorithms]{Milestones of RL algorithms evaluated in both practice
    and theory:
    PSRL~\citep{strens2000bayesian,osband2017posterior}, RLSVI~\citep{osband2016generalization,osband2019deep}, Ensemble+~\citep{osband2018randomized,osband2019deep},
    and our \HyperAgent. In deep RL, $\ok$ indicate intermediate outcomes between $\cmark$ and $\xmark$. In tabular RL, we consider the number of states ($S$), actions ($A$), horizons ($H$), and episodes ($K$).
}
    \label{tab:hyperagent-rl}
    \vspace{-1em}
\end{table*}

\subsection{Related Works}
\label{sec:related-works}

The modern development of practical RL algorithms provides scalable solutions to the challenges posed by large state spaces and the increasing size of interaction data under resource constraints. These algorithms' per-step computation complexity scales sub-linearly with (1) the problem size, thanks to function approximation techniques \citep{bertsekas1996neuro, mnih2015human}; and (2) the increasing size of interaction data, due to advancements in temporal difference learning \citep{sutton2018reinforcement}, Q-learning \citep{watkins1992q}, and incremental SGD with finite buffers \citep{mnih2015human}.
These advances yielded impressive results in simulated environments and attracted significant interest~\citep{mnih2015human,schrittwieser2020mastering}. However, data efficiency remains a barrier to transferring this success to the real world~\citep{lu2023reinforcement}.

To address data-efficiency empirically, recent deep RL algorithms have incorporated increasingly complex heuristic and algorithmic components, such as DDQN \citep{van2016deep}, Rainbow \citep{hessel2018rainbow}, EfficientZero \citep{ye2021mastering}, and BBF \citep{schwarzer2023bigger} whose details are around \cref{tb:tech_simple} in \cref{sec:related}. Moreover, these algorithms lack theoretical efficiency guarantees; for example, BBF employs $\varepsilon$-greedy exploration, which is provably data inefficient, requiring an exponential number of samples \citep{kakade2003sample, strehl2007probably, osband2019deep, dann2022guarantees}. $\varepsilon$-greedy strategy is still popular in practice due to its simplicity in implementation, requiring very few additional lines of code. We aim to develop a simple replacement for $\varepsilon$-greedy by \HyperAgent for practical concerns while achieving data efficient exploration with performance guarantees.

Efficient exploration in reinforcement learning hinges on decisions driven not only by expectations but also by epistemic uncertainty \citep{russo2018tutorial, osband2019deep}. Such decisions are informed by immediate and subsequent observations over a long horizon, embodying the concept of deep exploration \citep{osband2019deep}. Among the pivotal exploration strategies in sequential decision-making is Thompson Sampling (TS), which bases decisions on a posterior distribution over models, reflecting the degree of epistemic uncertainty \citep{thompson1933likelihood, strens2000bayesian, russo2018tutorial}. In its basic form, TS involves sampling a model from the posterior and selecting an action that is optimal according to the sampled model. However, exact posterior sampling remains computationally feasible only in simple environments--like Beta-Bernoulli and Linear-Gaussian Bandits, as well as tabular MDPs with Dirichlet priors over transition vectors--where conjugacy facilitates efficient posterior updates \citep{russo2018tutorial, strens2000bayesian}.

To extend TS to more complex environments, approximations are indispensable \citep{russo2018tutorial}, encompassing both function approximation for scalability across large state spaces and posterior approximation for epistemic uncertainty estimation beyond conjugate scenarios. Randomized Least-Squares Value Iteration (RLSVI) represents another value-based TS approach, aiming to approximate posterior sampling over the optimal value function without explicitly representing the distribution. This method achieves tractability for value function approximation by introducing randomness through perturbations, thus facilitating deep exploration and enhancing data efficiency \citep{osband2019deep}. Despite avoiding explicit posterior maintenance, RLSVI demands significant computational effort to generate new point estimates for each episode through independent perturbations and solving the perturbed optimization problem anew, without leveraging previous computations for incremental updates. Consequently, while RLSVI remains feasible under value function approximation, its scalability is challenged by growing interaction data, a limitation shared by subsequent methods  \citep{ishfaq2021randomized}.

\paragraph{Bridging the Gap.}
The divergence between theoretical and practical realms in reinforcement learning (RL) is expanding, with theoretical algorithms lacking practical applicability and practical algorithms exhibiting empirical and theoretical inefficiencies. Ensemble sampling, as introduced by \citet{osband2016deep, osband2018randomized, osband2019deep}, emerges as a promising technique to approximate the performance of RLSVI. Further attempts such as Incre-Bayes-UCBVI and Bayes-UCBDQN
\citep{tiapkin2022dirichlet} incorporate ensemble-based empirical bootstraps to approximate Bayes-UCBVI, which aims to bridge this gap. These methods maintain multiple point estimates, updated incrementally, essential for scalability. However, the computational demand of managing an ensemble of complex models escalates, especially as the ensemble size must increase to accurately approximate complex posterior distributions \citep{dwaracherla2020hypermodels, osband2023epistemic, li2022hyperdqn, qin2022analysis}.
An alternative strategy involves leveraging a hypermodel \citep{dwaracherla2020hypermodels, li2022hyperdqn} or epistemic neural networks (ENN) \citep{osband2023epistemic, osband2023approximate} to generate approximate posterior samples. This approach, while promising, demands a representation potentially more intricate than simple point estimates. The computational overhead of these models, including ensembles, hypermodels, and ENNs, is theoretically under-explored. 
More discussion of related works can be found in \cref{sec:related}.

\section{Reinforcement Learning \& Hypermodel}
\label{sec:pre}
\def\cM{{\mathcal{M}}}
We consider the episodic RL setting in which an agent interacts with an unknown environment over a sequence of episodes.
We model the environment as a Markov Decision Problem (MDP) $\cM=\left(\mathcal{S}, \mathcal{A}, P, r, s_{\operatorname{terminal}}, \rho\right)$, where $\mathcal{S}$ is the state space, $\mathcal{A}$ is the action space, $\operatorname{terminal} \in \mathcal{S}$ is the terminal state, and $\rho$ is the initial state distribution.
For each episode, the initial state $S_0$ is drawn from the distribution $\rho$. 
At each time step $t=1,2, \ldots$ within an episode, the agent observes a state $S_t \in \mathcal{S}$. If $S_t \neq s_{\operatorname{terminal}}$, the agent selects an action $A_t \in \mathcal{A}$, transits to a new state $S_{t+1} \sim P\left(\cdot \mid S_t, A_t \right)$, with reward $R_{t+1} = r(S_t, A_t, S_{t+1})$.
An episode terminates once the agent arrives at the terminal state. Let $\tau$ be the termination time
, i.e., $S_{\tau} = s_{\operatorname{terminal}}$.
A policy $\pi: \mathcal{S} \rightarrow \mathcal{A}$ maps a state $s \in \mathcal{S}$ to an action $a \in \mathcal{A}$. For each MDP $\cM$ and each policy $\pi$, we define the associated action-value function as
\[
    Q_\cM^{\pi}(s, a):=\mathbb{E}_{\cM, \pi}\left[\sum_{t=1}^\tau R_{t} \mid S_0=s, A_0=a\right],
\]
where the subscript $\pi$ under the expectation indicates that actions over the time periods are selected according to the policy $\pi$. Let $V_\cM^{\pi}(s):=Q_\cM^{\pi}(s, \pi(s))$. We further define the optimal value function $V_\cM^{\star}(s) = \max_{\pi} V_\cM^{\pi}(s)$ for all $s\in \mathcal{S}$ where it takes the maximum on optimal policy $\pi^*$. Optimal policy also corresponds to the optimal action-value function, denoted $Q^\star$ and defined as 
\begin{align}
\label{eq:Q-star}
    Q^\star(s, a) = Q^{\pi^*}_{\cM}(s, a) \quad \forall (s, a) \in \mathcal{S} \times \mathcal{A}.
\end{align}
In the reinforcement learning problem, the agent is given knowledge about $\mathcal{S}, \mathcal{A}, r, s_{\operatorname{terminal}}$, and $\rho$, but is uncertain about transition $P$. The unknown MDP $\cM$, together with the unknown transition function $P$, are modeled as random variables drawn from a prior distribution. As a consequence, the optimal action-value function $Q^\star$ is also a random variable that the agent is uncertain about at the beginning. Thus, the agent needs to explore the environment and gather information to resolve this uncertainty.

\subsection{Hypermodel}
\label{sec:hypermodel}
As maintaining the degree of uncertainty~\citep{russo2018tutorial} is crucial for data-efficient sequential decision-making, we build RL agents based on the hypermodel \citep{li2022hyperdqn,dwaracherla2020hypermodels} framework for epistemic uncertainty estimation.
The hypermodel takes an input \(x \in \R^{d}\) and a random index \(\xi \sim P_{\xi}\) from a fixed reference distribution, producing an output \(f_{\theta}(x, \xi)\) that reflects a sample from the approximate posterior, measuring the degree of uncertainty. 
The variation in the hypermodel's output with \(\xi\) captures the model's degree of uncertainty about \(x\), providing a dynamic and adaptable approach to uncertainty representation. This design, combining a trainable parameter \(\theta\) with a constant reference distribution \(P_{\xi}\), allows the hypermodel to adjust its uncertainty quantification over time, optimizing its performance and decision-making capabilities in dynamic environments.
\textbf{(1)} For example, a special case of a linear hypermodel is $f_{\theta}(x, \xi) = \langle x, \mu + \rmA \xi \rangle$ with $\theta = (\rmA \in \R^{d \times M}, \mu \in \R^d)$ and $P_\xi = N(0, I_M)$, which is essentially the Box-Muller transformation: one could sample from a linear-Gaussian model $N(x^\top\mu, x^\top \Sigma x)$ via a linear hypermodel if $\rmA \rmA^\top = \Sigma$.
\textbf{(2)} Another special case is the ensemble sampling: with a uniform distribution $P_{\xi} = \mathcal{U}(e_1, \ldots, e_M)$ and an ensemble of models $\theta = \rmA = [\tilde{\theta}_{1}, \ldots, \tilde{\theta}_{M}] \in \R^{d \times M}$ such that $\tilde{\theta}_{m} \sim N(\mu, \Sigma)$, one can uniformly sample from these ensembles by a form of hypermodel $f_{\theta}(x, \xi) := \langle x, \rmA \xi \rangle$.
In general, the hypermodel $f_{\theta}(\cdot)$ can be any function approximator, e.g., neural networks, transforming the reference distribution $P_{\xi}$ to an arbitrary distribution.
We adopt a class of hypermodel that can be represented as an additive function of a learnable function and a fixed prior model (additive prior assumption),
\begin{align}
    \label{eq:hypermodel}
    \underbrace{f_{\theta}(x, \xi)}_{\text{``Posterior'' Hypermodel}} = \underbrace{f_{\theta}^L(x, \xi)}_{{\text{Learnable function}}} + \underbrace{f^{P}(x, \xi)}_{{\text{Fixed prior model}}}
\end{align}
The prior model $f^P$ represents the prior bias and prior uncertainty, and it has no trainable parameters. The learnable function is initialized to output values near zero and is then trained by fitting the data. The resultant sum $f_\theta(x, \cdot)$ produces reasonable predictions for all probable values of $\xi$, capturing epistemic uncertainty. This additive prior assumption can be validated under linear-Gaussian model~\citep{osband2018randomized}, also related to  Matheron’s rule~\citep{journel1976mining,hoffman1991constrained,doucet2010note} with applications in Gaussian processes~\citep{wilson2020efficiently}.

\begin{figure}[t]
    \centering
    \includegraphics[width=\linewidth]{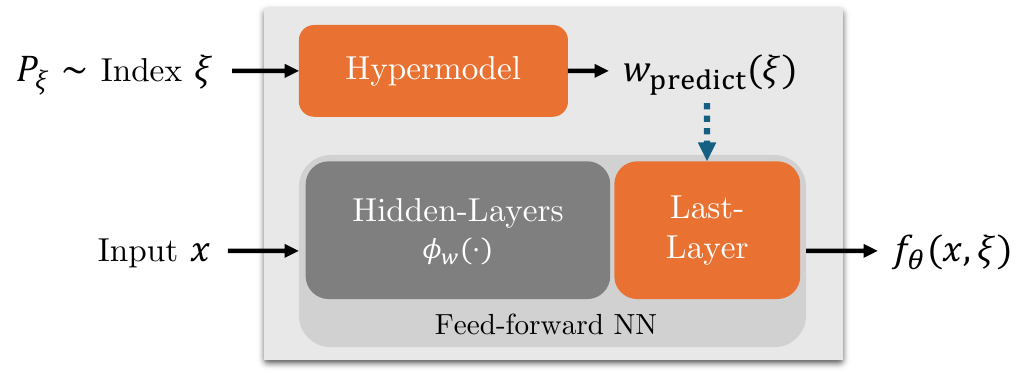}
    \caption[Description of the hypermodel]{Last-layer linear hypermodel.}
    \label{fig:hypremodel}
\end{figure}

As described in \cref{fig:hypremodel}, we design the hypermodel for feed-forward neural networks (NN) with the \textit{last-layer linear hypermodel} assumption: the degree of uncertainty can be approximated by a linear hypermodel $f_{\theta}(\cdot, \xi) = \langle \phi_w(\cdot), {w_{\operatorname{predict}}(\xi)} \rangle$ over the last-layer $w_{\operatorname{predict}}(\xi) = {\rmA \xi + b}$.
This new assumption is unconventional to the literature~\citep{dwaracherla2020hypermodels,li2022hyperdqn,osband2023epistemic}, and can be validated when the hidden-layer feature mapping $\phi_{w}(\cdot)$ is fixed through the learning process, as proved in \cref{sec:analysis}.
We discuss the details of the last-layer linear hypermodel and clarify the critical differences and advantages compared with prior works in \cref{sec:hyperfqi-dnn}.

\section{Algorithm design}
\label{sec:alg}
\begin{algorithm*}[t]
\caption{\texttt{HyperAgent} }
\label{alg:hyperagent}
\begin{algorithmic}[1]
  \STATE \textbf{Input:} 
  Reference $P_{\xi}$. Perturbation $P_{\rvz}$. Buffer $D$.
  Initialize $\theta = \theta^- = \theta_{\operatorname{init}}$, train step $j = 0$.
  \FOR{each episode $k=1,2,\ldots$}
    \STATE \textbf{Sample} an \textit{index mapping} $\vxi_k(\cdot) \sim P_{\xi}$.
    Set $t=0$ and \textbf{observe} $S_{k,0} \sim \rho$
    \REPEAT
      \STATE \textbf{Select} $A_{k,t} = \arg\max_{a \in \mathcal{A}} f_{\theta}(S_{k,t}, a, \vxi_k(S_{k,t}))$
      \STATE \textbf{Observe} $S_{k,t+1}$ from the environment and $R_{k,t+1} = r(S_{k,t}, A_{k,t}, S_{k,t+1})$
      \STATE \textbf{Sample} perturbation random vector $\rvz_{k,t+1} \sim P_{\rvz}$;
      Add $(S_{k,t}, A_{k,t}, R_{k,t+1}, S_{k,t+1}, \rvz_{k,t+1})$ to buffer $D$
      \STATE Increment step counter $t \leftarrow t+1$;
      Compute  $\theta, \theta^{-}, j \leftarrow \update(D, \theta, \theta^-, \vxi^- = \vxi_k, t, j)$ with \cref{alg:update-dqn}
    \UNTIL{$S_{k,t} = s_{\operatorname{terminal}}$}
  \ENDFOR
\end{algorithmic}
\end{algorithm*}
We now describe \HyperAgent, a DQN-type algorithm for large-scale complex environments. \HyperAgent consists of three key components:
\begin{enumerate}[leftmargin=*]
    \item A hypermodel that maintains an approximate distribution over optimal value function $Q^\star$.
    \item An incremental update mechanism to update the hypermodels.
    \item An index sampling scheme that uses the hypermodels for exploration.
\end{enumerate}

In the context of reinforcement learning, we define the action-value function with hypermodel as $f_{\theta}: \mathcal{S} \times \mathcal{A} \times \Xi \rightarrow \mathbb{R}$ parameterized by $\theta$, where $\Xi$ is the index space. As we introduce random index following reference distribution $P_{\xi}$ as an input, the $f_{\theta}$ is essentially a randomized value function. Under the last-layer linear hypermodel assumption, we define the action-value function $f_{\theta}(s, a, \xi)$ as 
\begin{align}
\label{eq:deep-hypermodel}
\underbrace{\langle \rmA^{(a)} \xi + b^{(a)}, \phi_w(s) \rangle}_{\text{Learnable} \ f^L_{\theta}(s, a, \xi)} + \underbrace{\langle \rmA^{(a)}_0\xi + b^{(a)}_0, \phi_{w_0}(s) \rangle}_{\text{Fixed prior} \ f^P(s, a, \xi)}
\end{align}
where $\theta$ includes a set of learnable parameters $\{w, \rmA^{(a)}, b^{(a)}\}$ and fixed parameters $\{w_0, \rmA_0^{(a)}, b_0^{(a)}\}$ for each action $a \in \mathcal{A}$.
The action-value function $f_{\theta}$ based on the hypermodel is then trained by minimizing the loss function motivated by fitted Q-iteration (FQI), a classical method \citep{ernst2005tree} for batch-based function approximation for optimal action-value $Q^\star$, with a famous online extension called DQN~\citep{mnih2015human}.
An important notion we introduce that differentiate \HyperAgent to DQN is the \textit{index mapping} $\vxi^-: \mathcal{S} \rightarrow {\Xi}$.
The \textit{index sampling} procedure is to produce random variables $\vxi^-(s)$ following $P_{\xi}$ for each $s \in \mathcal{S}$ (line 4) and to pergorm greedy action selection (line 7). Intuitively, this procedure introduces noises that are independent across episodes, and induces diverse exploration behavior.

For training, \HyperAgent maintains two hypermodels: one for the main value function $f_{\theta}$ and the other for the target value function
$f_{\theta^-}$ where $\theta^-$ is the target parameters.
It also maintains a buffer of transitions $D = \{(s, a, r, s', \rvz)\}$, where $\rvz \in \mathbb{R}^M$ is the algorithmic perturbation vector sampled from the perturbation distribution $P_{\rvz}$ (described in line 9).
Let $\gamma$ be the discounted factor. We denote the \textit{perturbed temporal difference }(TD) loss for a given index $\xi$ and a transition tuple $d$ as $\ell^{\gamma, \sigma}(\theta; \theta^{-}, \vxi^-, \xi, d )$, defined as
\begin{align*}
[ r + \sigma \xi^\top \rvz + \gamma \max_{a' \in \mathcal{A}} f_{\theta^-}(s', a', \vxi^-(s')) - f_{\theta}(s, a, \xi) ]^2,
\end{align*}
where $\sigma$ is a hyperparameter to control the variance of algorithmic random perturbation.
\HyperAgent updates the hypermodel by minimizing the loss $L^{\gamma, \sigma, \beta}(\theta;\theta^-, \vxi^-, D )$ as
\begin{align}
\label{eq:hyperfqi}
\mathbb{E}_{\xi \sim P_{\xi}}{\left[ \sum_{d \in D} \frac{1}{|D|}
\ell^{\gamma, \sigma}(\theta; \theta^-, \vxi^-, \xi, d) \right]}
+ \frac{\beta}{|D|} \| \theta \|^2,
\end{align}
where $\beta \ge 0$ is for prior regularization. We optimize the loss function \cref{eq:hyperfqi} using SGD with a mini-batch of data $\tilde{D}$ and a batch of indices $\tilde{\Xi}$ from $P_{\xi}$. That is, we take gradient descent w.r.t. the sampled loss $\tilde{L}(\theta; \theta^-, \vxi^-, \tilde{D})$ as
\begin{align}
\label{eq:hyperfqi-sampled}
\frac{1}{|\tilde{\Xi}|} \sum_{\xi \in \tilde{\Xi}} \sum_{d \in \tilde{D}} \frac{1}{|\tilde{D}|} \ell^{\gamma, \sigma}(\theta; \theta^-, \vxi^-, \xi, d) + \frac{\beta}{|D|} \| \theta \|^2.
\end{align}
Then, the target parameters $\theta^-$ are periodically updated to $\theta$. We summarized the \HyperAgent in \cref{alg:hyperagent} where the \update function through~\cref{eq:hyperfqi,eq:hyperfqi-sampled} is described in \cref{alg:update-dqn}.

For practitioners, the primal benefits and motivations include straightforward implementation as a plug-and-play alternative to DQN-type methods. It can also replace the $\varepsilon$-greedy exploration strategy. Extending actor-critic type deep reinforcement learning algorithms to incorporate similar advantages as in \HyperAgent can be readily achieved.

\section{Theoretical insights and analysis}
\label{sec:analysis}
In this section, we provide insights on how perturbed TD loss and index sampling works and why it performs efficient incremental posterior approximation for optimal value $Q^\star$ without reliance on conjugacy and faciliates deep exploration.
For clarity, we focus on tabular representations when $\phi_{w}(s) = \phi_{w_0}(s) = \1{s} \in \R^{\abs{\mathcal{S}}}$ is a fixed one-hot vector.

\textbf{Tabular setups.}
Let us define short notations $m_{sa} = \left(\rmA^{(a)}\right)^{\top} \phi_w(s)$ and $\mu_{sa} = \left(b^{(a)}\right)^{\top} \phi_w(s)$ for ease of exposition. Similarly, define $m_{0, sa}$ and $\mu_{0, sa}$ from $w_{0}, \rmA^{(a)}_0$ and $b^{(a)}_0$, respectively.
Following \cref{eq:deep-hypermodel}, 
\[
f_{\theta}(s, a, \xi) = \underbrace{\mu_{sa} + m_{sa}^\top \xi}_{\text{Learnable}~f^L_{\theta}(s, a, \xi)} + \underbrace{\mu_{0,sa} + m_{0, sa}^\top \xi}_{\text{Fixed Prior}~f^P(s, a, \xi)}
\]
where  
$\theta = (\mu \in \mathbb{R}^{\abs{\mathcal{S}}\abs{\mathcal{A}}}, m \in \mathbb{R}^{\abs{\mathcal{S}}\abs{\mathcal{A}} \times M})$ are the parameters to be learned; $m_{0, sa} := \sigma_0 \rvz_{0, sa}$ where $\rvz_{0, sa} \in \mathbb{R}^M$ is an independent random vector sampled from $P_{\rvz}$ and $\mu_{0, sa}, \sigma_0$ are prior mean and prior variance for each $(s, a) \in \mathcal{S} \times \mathcal{A}$.
The regularizer in \cref{eq:hyperfqi} now becomes $\beta \| \theta \|^2 = \beta \sum_{s, a} \left( \mu_{sa}^2 + \| m_{sa} \|^2 \right)$.

\textbf{History.}
Denote the sequence of observations in episode $k$ by 
$\mathcal{O}_k = \left(S_{k, t}, A_{k, t}, R_{k, t+1}, S_{k, t+1}\right)_{t=0}^{\tau_k - 1}$
where $S_{k, t}, A_{k, t}, R_{k, t+1}$ are the state, action, and reward at the $t$-th time step of the $k$-th episode, and $\tau_k$ is the termination time at episode $k$.
We denote the history of observations made prior to episode $k$ by $\mathcal{H}_k = \left(\mathcal{O}_1, \ldots, \mathcal{O}_{k-1}\right)$.
Without loss of generality, we assume that under any MDP $\cM$ and policy $\pi$, the termination time $\tau<\infty$ is finite with probability 1.
The agent's behavior is governed by the agent policy $\pi_{\operatorname{agent}}= (\pi_k)_{k=1}^K$, which uses the history $\mathcal{H}_k$ to select a policy $\pi_k = \operatorname{agent}(\mathcal{S}, \mathcal{A}, r, \mathcal{H}_k)$ for the $k$-th episode.

Therefore, we define some statistics related to the history $\mathcal{H}_k$. Let $E_k = \{0, 1, \ldots, \tau_k - 1\}$ denote the time index in episode $k$ and $E_{k, sa} = \{ t: (S_{k,t}, A_{k,t}) = (s, a), t \in E_{k} \}$ record the time index the agent encountered $(s, a)$ in the $k$-th episode.
Assume $T = \sum_{k = 1}^K \abs{E_k}$ total interactions encountered within $K$ episodes.
Let $N_{k, sa}=\sum_{\ell =1}^{k-1} \abs{ E_{\ell, sa}}$ denote the counts of visitation for state-action pair $(s, a)$ prior to episode $k$.
For every pair $(s, a)$ with $N_{k, sa}>0$, $\forall s'\in\mathcal{S}$, the empirical transition up to episode $k$ is $$\hat{P}_{k,sa}(s'):= \sum_{\ell=1}^{k-1}\sum_{t\in E_{\ell}}\frac{\1{(S_{\ell, t}, A_{\ell, t}, S_{\ell, t+1})=(s,a,s')}}{N_{k, sa}}.$$
In case $N_{k, sa} = 0$, define $\hat{P}_{k, sa}(s')=1$ for arbitrary $s'$.

\textbf{Stochastic Bellman operator.}
For the ease of explanation, let us use the vector notation $f_{\theta, \vxi}(s, a) := f_{\theta}(s, a, \vxi(s)).$
Initially, let $\theta_0 := (\mu_0 = \mathbf{0}, m_0 = \mathbf{0})$.
At episode $k$, \HyperAgent\ will act greedily w.r.t. the action-value vector $f_{\theta_k, \vxi_k}$ where $\theta_k = (\mu_k, m_k)$. That is, the agent chose policy $\pi_k(s) \in \argmax_{a \in \mathcal{A}} f_{\theta_k, \vxi_k}(s, a)$.

Now we explain how \HyperAgent\ performs incremental updates from $\theta_{k-1}$ to $\theta_k$. Suppose that, at the beginning of episode $k$, it maintains parameters $\theta_{k}^{(0)} := \theta_{k-1} = (\mu_{k-1}, m_{k-1})$, the buffer $D = \mathcal{H}_k$, and index mapping $\vxi^- = \vxi_k$.
By iteratively solving its objective \cref{eq:hyperfqi} with $\beta = \sigma^2/ \sigma_0^2, \theta^- = \theta_{k}^{(i)}$ and obtaining the closed-form solution $\theta_k^{(i+1)}$ from $i=0$ until converging to $\theta_k$, \HyperAgent\ would yield the closed-form iterative update rule (1) $m_{k-1} \rightarrow m_k$ and (2) $\theta_{k}^{(i)} = ( \mu_k^{(i)}, m_k )\rightarrow \theta_{k}^{(i+1)} = ( \mu_k^{(i+1)}, m_k )$ as follows.
Using short notation $\tilde{m}_{k, sa} = m_{k, sa} + \sigma_0 \rvz_{0, sa}, $ for all $(s, a, k)$, we have
{%
\begin{align}
    \label{eq:noise-incremenal}
    & (N_{k, sa} + \beta) \tilde{m}_{k, sa}  - ( N_{k-1, sa} + \beta) \tilde{m}_{k-1, sa} \nonumber \\
    & = \sum_{t \in E_{k-1, sa} } \sigma \rvz_{k-1, t+1}.
\end{align}
More interestingly, the iterative process on $\theta_k^{i},$ for $i=0, 1, \ldots$ can be described using the following equation
\begin{align}
    \label{eq:bellman-iteration}
    &f_{\theta_{k}^{(i+1)}, \vxi_k} = F^{\gamma}_{k} f_{\theta_{k}^{(i)}, \vxi_k},
\end{align}
where $F^{\gamma}_{k}$ can be regarded as a stochastic Bellman operator induced by \HyperAgent\ in episode $k$: i.e., for any $Q$-value function, $F_{k}^{\gamma} Q (s, a)$ is defined as
\begin{align}
    \label{eq:bellman-hyperfqi}
    \frac{\beta \mu_{0, sa} + N_{k, sa}(r_{sa} + \gamma V_Q^\top \hat{P}_{k, sa})}{N_{k, sa} + \beta} + \tilde{m}_{k, sa}^\top \vxi_k(s),
\end{align}
where $V_{Q}(s) := \max_{a} Q(s, a)$ is the greedy value with respect to $Q$.
}
The derivations for \cref{eq:noise-incremenal,eq:bellman-iteration,eq:bellman-hyperfqi} can be found in \cref{sec:hyperfqi-tabular}, which are crucial for understanding.

\textbf{True Bellman operator.} Before digging into \cref{eq:noise-incremenal,eq:bellman-iteration,eq:bellman-hyperfqi}, let us first examine the true Bellman operator $F_{\cM}^{\gamma}$: when applied to fixed and bounded \( Q \), for any $ (s, a)\in \mathcal{S} \times \mathcal{A}$
\begin{align}
    \label{eq:bellman-true}
    F_{\cM}^{\gamma} Q (s, a)
     & = r_{sa} + \gamma V_Q^\top P_{sa}.
\end{align}
Note the true Bellman operator has a fixed point on optimal action-value function, i.e., $Q^\star = F_{\cM}^\gamma Q^{\star}$.
Since the transition $P$ is a random variable in our setup, the true Bellman operator is also a random variable that propagates uncertainty for $Q^\star$. As will be discussed in \cref{lem:so-operator}, conditioned on the history $\mathcal{H}_k$ up to episode $k$, the  posterior variance of then Bellman equation in \cref{eq:bellman-true} is inversely proportion to the visitation counts on $(s,a)$
\begin{align}
\label{eq:true-bellman-variance}
    \var{F_{\cM}^{\gamma} Q (s, a)\mid \mathcal{H}_k } \propto \frac{1}{N_{k, sa} + \beta}.
\end{align}
Intuitively, more experience at $(s,a)$ leads to less epistemic uncertainty and smaller posterior variance on $Q^\star(s,a)$.

\textbf{Understanding \cref{eq:noise-incremenal}.}
This is an incremental update with computational complexity $O(M)$.
A key property of the hypermodel in \HyperAgent\ is that, with logarithmically small $M$, it can approximate the posterior variance sequentially in every episode via incremental update in \cref{eq:noise-incremenal}. This is formalized as the key lemma.
\begin{lemma}[Incremental posterior approximation]
    \label{lem:approx}
    For $\tilde{m}_k$ recursively defined in \cref{eq:noise-incremenal} with $\rvz \sim \mathcal{U}(\mathbb{S}^{M-1})$.
    For any $k \ge 1$, define the good event of $\varepsilon$-approximation
    \begin{align*}
        \mathcal{G}_{k, sa}(\varepsilon) := \bigg\{
        \|\tilde{m}_{k, sa} \|^2  \in \left( \frac{(1-\varepsilon)\sigma^2}{N_{k, sa} + \beta}, \frac{(1+\varepsilon)\sigma^2}{N_{k, sa} + \beta} \right)  \bigg\}.
    \end{align*}
    The joint event $\cap_{(s,a) \in \mathcal{S} \times \mathcal{A}} \cap_{k=1}^K \mathcal{G}_{k, sa}(\varepsilon) $ holds with probability at least $1 - \delta$ if $M \simeq \varepsilon^{-2} \log(\abs{\mathcal{S}} \abs{\mathcal{A}} T/\delta)$.
\end{lemma}
A direct observation from \cref{lem:approx} is larger $M$ results in smaller $\varepsilon$, implying more accurate approximation of posterior variance over $Q^\star(s,a)$ using their visitation counts and appropriately chosen $\sigma$.
As shown in \cref{sec:proof-approx}, the difficulty in proving \cref{lem:approx} arises from the sequential dependence among high-dimensional random variables. It is resolved due to the first probability tool~\citep{li2024probability} for sequential random projection. Additionally, in the regret analysis, we will show constant approximation $\varepsilon = 1/2$ suffices for efficient deep exploration with logarithmic per-step computation costs. To highlight, \cref{lem:approx} also validates our assumption about the \textit{last-layer linear hypermodel} in the special case when the hidden layer feature mapping $\phi_{w}(\cdot)$ is a fixed one-hot mapping. We leave the exploration of general setups for $\phi_{w}(\cdot)$ to future work.

\textbf{Understanding \cref{eq:bellman-hyperfqi}.}
Now, we will argue that the \HyperAgent-induced operator $F_{k}^{\gamma}$ is essentially mimicking the behavior of the true Bellman operator $F^{\gamma}_{\cM}$ conditioned on history $\mathcal{H}_k$, thus producing an approximate posterior over $Q^\star$.
As shown in \cref{lem:contraction}, the stochastic Bellman operator $F_{k}^{\gamma}$ is a contraction mapping and thus guarantees convergence of \cref{eq:bellman-iteration}. It differs from the empirical Bellman iteration $V_Q^\top \hat{{P}}_{k, sa}$ in two ways: (1) there is a slight regularization toward the prior mean $\mu_{0, sa}$, and (2) more importantly, \HyperAgent\ adds noise $w_{k, sa} := \tilde{m}_{k, sa}^\top \vxi_k(s)$ to each iteration. 
For a common choice $\vxi_k(s) \sim P_{\xi} := N(0, I_M)$, the noise $w_{k, sa}$ is Gaussian distributed conditioned on $\tilde{m}_{k, sa}$.
Importantly, as shown by \cref{lem:approx}, the variance of the perturbation noise
\[
    \var[\vxi_k]{w_{k, sa}} = \norm{\tilde{m}_{k, sa}}^2 \propto \frac{1}{N_{k, sa} + \beta}
\]
coincides with the posterior variance of the true Bellman operator in \cref{eq:true-bellman-variance} up to a constant.
Incorporating the Gaussian noise in the Bellman iteration would backpropagate the uncertainty estimates and approximate the posteriors associated with the optimal action-value $Q^\star$, which is essential to incentivize deep exploration behavior.
This is because the injected noise $w$ from later states with less visitation counts (thus larger variance) will be backpropagated to initial state by the \HyperAgent.
A simple illustration on this efficient deep exploration behavior is in \cref{sec:insight}.

\textbf{Regret bound.}
To rigorously justify the algorithmic insights and benefits, we provide theoretical results for \HyperAgent\ with tabular representation and hyperparameters for \update specified in \cref{tab:hyper_parameter_tabular}. 
Denote the regret of a policy $\pi_k$ over episode $k$ by
\(
    \Delta_k := V_{\cM}^{\star}(s_{k, 0}) - V_{\cM}^{\pi_k}(s_{k, 0}).
\)
Maximizing total reward is equivalent to minimizing the expected total regret:
$\operatorname{Regret}(K, \operatorname{agent}) := \mathbb{E} \sum_{k=1}^K \Delta_k.$

\begin{theorem}
\label{thm:regret-HyperAgent}
    Under \cref{asmp:finite-inhomogeneous-mdp,assump:dirichlet-prior} with $\beta \geq 3$, if the Tabular \HyperAgent\ is applied with planning horizon $H$, and parameters with $(M, \mu_0, \sigma, \sigma_0)$ satisfying 
    \[
    M = O(1) \cdot \log(SAH K) = \tilde{O}(\log K),
    \]
    $(\sigma^2/\sigma_0^2) = \beta$, $\sigma \ge \sqrt{6} H$ and $\min_{sa} \mu_{0, sa} \ge H$, then $\forall K \in \mathbb{N}$, $\operatorname{Regret}(K, \HyperAgent)$ is upper bounded by
    \[
        18 H^2 \sqrt{ \beta SA K \log _{+}(1+SAHK)} \log _{+}\left(1+{K}/{SA}\right),
    \]
    where $\log _{+}(x)=\max(1, \log(x))$.
    The per-step computation of \HyperAgent\ is $O(S^2A + SAM)$.
\end{theorem}
\begin{remark}
    The time-inhomogeneous MDP is a common benchmark for regret analysis in the literature, e.g., \citep{azar2017minimax,osband2019deep}. 
    For notation unification with \cref{tab:hyperagent-rl}, as explained in \cref{asmp:finite-inhomogeneous-mdp}, $\abs{\mathcal{S}} = SH, \abs{\mathcal{A}} = A$ and $\tau= H$ almost surely.
    \cref{assump:dirichlet-prior} is common in the literature of Bayesian regret analysis  \citep{osband2019deep,osband2017posterior,lu2019information}.
    The regret bound $\tilde{O}(H^2 \sqrt{SAK})$ of \HyperAgent\ matches the best known Bayesian regret bound, such as those of RLSVI~\citep{osband2019deep} and PSRL~\citep{osband2017posterior}, while \HyperAgent\ provides the computation and scalability benefits that RLSVI and PSRL do not have, as discussed in \cref{sec:intro,tab:hyperagent-rl}.
    On the other hand, most practically scalable algorithms, including recent BBF~\citep{schwarzer2023bigger}, use $\varepsilon$-greedy, which is provably data-inefficient without sublinear regret guarantees in general~\citep{dann2022guarantees}.
    As shown in \cref{tab:hyperagent-rl}, a recent concurrent work claims LMC-LSVI~\citep{ishfaq2024provable} is practical and provable but suffers $\tilde{O}(K)$ per-step computational complexity, which is not acceptable under bounded computation resource constraints. Instead, the per-step computation of \HyperAgent\ is $\tilde{O}(\log K)$. This is due to the choice of $M$ in \cref{sec:proof-approx} when proving the key \cref{lem:approx}.
    These comparisons imply \HyperAgent\ is the first provably scalable and efficient RL agent among practically useful ones.
    Analytical details are in \cref{sec:regret}.
    Extension to frequentist regret without \cref{assump:dirichlet-prior} is direct either using \HyperAgent\ or its variants of optimistic index sampling in \cref{sec:index-sampling-schemes}.
\end{remark}

\section{Empirical studies}
\label{sec:exp_study}

This section assesses the efficiency and scalability of \HyperAgent empirically. In DeepSea hard exploration problems, it is the \textit{only and first} deep RL algorithm that successfully handling large state spaces up to $120 \times 120$ with optimal episodes complexity.
In Atari benchmark suite, it excels in processing continuous state spaces with pixels and achieves human-level performance with comparably minimal total number of interactions. \HyperAgent achieves the best performance in 8 hardest exploration compared to other approximate posterior sampling type deep RL algorithms.
We provide reproduction details in~\cref{sec:reproduce}.

\subsection{Computational results for deep exploration}
\label{sec:exp_deepsea}
\begin{figure*}[htbp]
    \centering
    \includegraphics[width=0.7\linewidth]{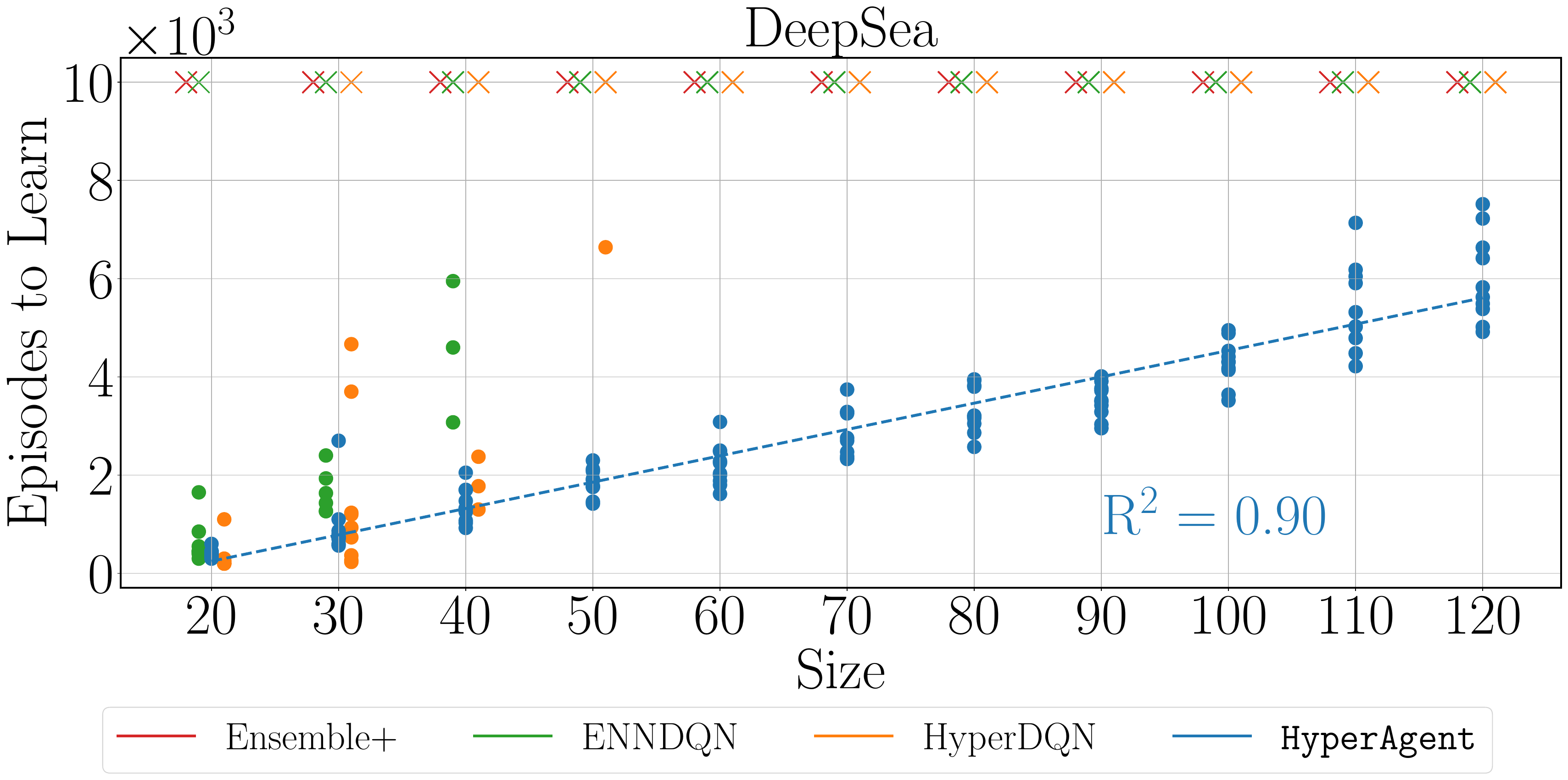} %
    \caption{The metric $\operatorname{Episodes~to~Learn}(N):= \operatorname{avg}\{K|\bar{R}_K \geq 0.99\}$ measures the episodes needed to learn the optimal policy in DeepSea of size $N$, where $\bar{R}_K$ is the return achieved by the agent after $K$ episodes of interaction, averaged over 100 evaluations. The crossmark $\xmark$ denotes the algorithm's failure to solve the problem within $10^4$ episodes. 
    We conduct experiments on each algorithm with 10 different initial random seeds, presenting each result as a distinct point in the figure.
    The dashed line for \HyperAgent, based on linear regression with an $R^2$ of 0.90, illustrates linear scaling in episode complexity, represented by $\Theta(N)$.
    }
    \label{fig:deepsea_baseline}
    \vspace{-0.7em}
\end{figure*}
We demonstrate the exploration effectiveness and scalability of \HyperAgent utilizing DeepSea, a reward-sparse environment that demands deep exploration~\citep{osband2019behaviour,osband2019deep}.
Details for DeepSea are in~\cref{appendix:env_settings}.

\textbf{Comparative analysis.}
Based on the structure of DeepSea with size $N$, i.e., $N \times N$ states, the proficient agent can discern an optimal policy within $\Theta(N)$ episodes~\citep{osband2019deep}, since it learns to move right from one additional cell along the diagonal in each episode.
We compare \HyperAgent with several baselines: {Ensemble+}~\citep{osband2018randomized,osband2019deep}, {HyperDQN}~\citep{li2022hyperdqn}, and {ENNDQN}~\citep{osband2023approximate}, which also claimed deep exploration ability.
As depicted in Figure~\ref{fig:deepsea_baseline}, \HyperAgent outperforms other baselines, showcasing its exceptional data efficiency. To highlight, it is the \textit{only and first} deep RL agent learning the optimal policy with optimal episodes complexity ${\Theta}(N)$.
Moreover, \HyperAgent offers the advantage of computation efficiency as its output layer (hypermodel) maintains \textit{constant parameters} when scaling up the problem size.
In contrast, ENNDQN requires increasing number of parameters as the problem size increases, due to the inclusion of the original state as part of the inputs (see~\cref{appendix:difference} for detailed discussions). For instance, in DeepSea($N=20$), \HyperAgent uses only $5\%$ of the parameters required by ENNDQN. 

Through more ablation studies on DeepSea in \cref{appendix:additional_res_deepsea}, we offer a comprehensive understanding of \HyperAgent, including validation for theoretical insights in \cref{sec:analysis}, index sampling schemes and sample-average approximation in \cref{sec:alg}, and comparison with structures within the hypermodel framework in~\cref{sec:hypermodel}.

\subsection{Results on Atari benchmark}
\label{sec:exp_atari}

\textbf{Baselines.}
We further assess the data and computation efficiency on the Arcade Learning Environment~\citep{bellemare2013arcade} using IQM~\citep{agarwal2021deep} as the evaluation criterion. An IQM score of 1.0 indicates that the algorithm performs on par with humans. We examine \HyperAgent with several baselines: DDQN$^\dag$~\citep{van2016deep}, Ensemble+~\citep{osband2018randomized,osband2019deep}, Rainbow~\citep{hessel2018rainbow}, DER~\citep{van2019use}, HyperDQN~\citep{li2022hyperdqn}, BBF~\citep{schwarzer2023bigger}, and EfficientZero~\citep{ye2021mastering}.
Following the established practice in widely accepted research~\citep{kaiser2019model,van2019use,ye2021mastering}, the results are compared on 26 Atari games.

\textbf{Overall results.}
Figure~\ref{fig:data-para} illustrates the correlation between model parameters and training data for achieving human-level performance. \HyperAgent attains human-level performance with minimal parameters and relatively modest training data, surpassing other methods. Notably, neither DER nor HyperDQN can achieve human-level performance within 2M training data (refer to \cref{appendix:additional_res_atari}).

\begin{table}[htbp]
    \centering
    \resizebox{0.99\linewidth}
    {!}{
        \begin{tabular}{lccc}
            \toprule[1.2pt]
            Method      & IQM               & Median            & Mean              \\
            \midrule \midrule
            DDQN$^\dag$ & 0.13 (0.11, 0.15) & 0.12 (0.07, 0.14) & 0.49 (0.43, 0.55) \\
            DDQN(ours)  & 0.70 (0.69, 0.71) & 0.55 (0.54, 0.58) & 0.97 (0.95, 1.00) \\
            \HyperAgent & 1.22 (1.15, 1.30) & 1.07 (1.03, 1.14) & 1.97 (1.89, 2.07) \\
            \toprule[1.2pt]
        \end{tabular}
    }
    \caption{Performance profiles of \HyperAgent across 26 Atari games with 2M training data. The data in parentheses represent the 95\% confidence interval.}
    \label{tb:atari_pf}
\end{table}

\textbf{Ablation study.}
Table~\ref{tb:atari_pf} displays the comprehensive results of \HyperAgent across 26 Atari games. To demonstrate the superior performance of \HyperAgent stemming from our principled algorithm design rather than the fine-tuning of hyper-parameters, we developed our version of DDQN, referred to as DDQN(ours). This implementation mirrors the hyper-parameters and network structure (except for the last layer) of \HyperAgent.
The comparative result with vanilla DDQN$^\dag$~\citep{hessel2018rainbow} indicates that (1) DDQN(ours) outperforms DDQN$^\dag$ due to hyperparameters adjustments,
and (2) \HyperAgent exhibits superior performance compared to DDQN(ours), owing to the inclusion of an additional hypermodel, index sampling schemes, and incremental mechanisms that facilitate deep exploration. It is worth noting that we also applied an identical set of hyper-parameters across all 55 Atari games (refer to \cref{appendix:additional_res_atari}), where \HyperAgent achieves top performance in 31 out of 55 games, underscoring its robustness and scalability.

\begin{figure*}[htbp]
    \centering
            \includegraphics[width=0.95\linewidth]{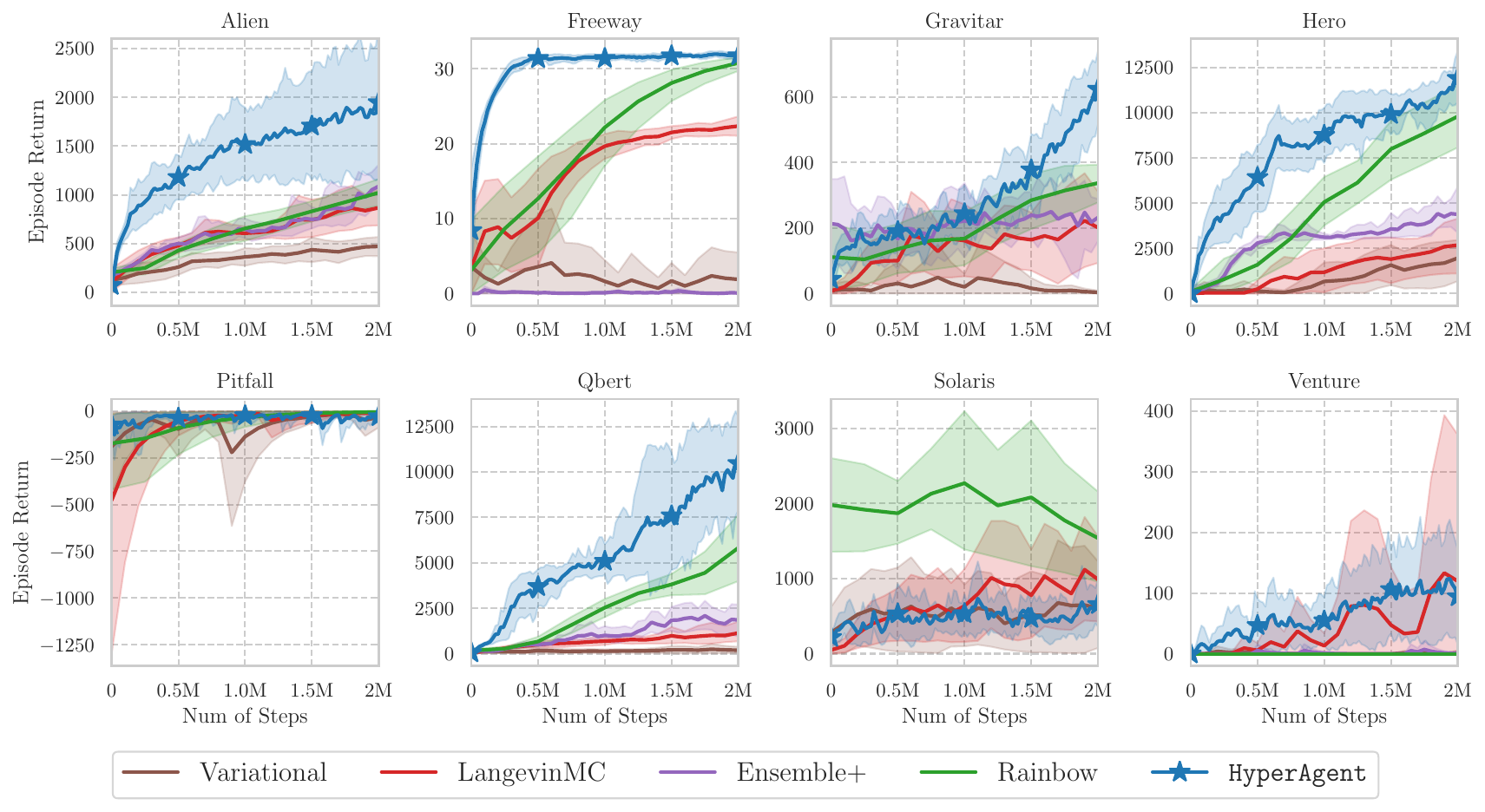}
        \caption{
        Comparison of \HyperAgent on the 8 hardest exploration Atari games with Variational approximation \citep[SANE][]{aravindan2021state}, LangevinMC \citep[AdamLMCDQN][an extension of LMC-LSVI in deep RL]{ishfaq2024provable}, Ensemble+ \citep[][Ensemble sampling with a randomized prior function]{osband2018randomized,osband2019deep} and Rainbow~\citep{hessel2018rainbow}.
    }
    \label{fig:atari_hardest}
    \vspace{-0.5em}
\end{figure*}

\textbf{Exploration on Atari.}
Through a comparison with algorithms related to approximate posterior sampling on the 8 most challenging exploration Atari games~\citep{bellemare2016unifying}, as shown in Figure~\ref{fig:atari_hardest}, \HyperAgent can achieve the best performance in 7 out of 8 games, demonstrating its ability to efficiently track the approximate posteriors over $Q^\star$ and perform deep exploration.

\section{Conclusion and future directions}
\label{sec:conclusion}
We present a reinforcement learning (RL) algorithm, \HyperAgent, that simplifies, accelerates, and scales the learning process across complex environments. With a hypermodel, index sampling, and incremental updates, \HyperAgent efficiently tracks the approximate posterior distribution associated with the optimal value function $Q^\star$ and performs the greedy policy over randomly sampled $Q^\star$ from this distribution to facilitate deep exploration. It achieves significant efficiency in data and computation over baselines. This is demonstrated through superior performance in challenging benchmarks like the DeepSea and Atari suites with minimal computational resources. \HyperAgent's algorithmic simplicity, practical efficiency, and theoretical underpinnings—highlighted by the incremental posterior approximation argument with a novel reduction to sequential random projection~\citep{li2024probability}—establish \HyperAgent as a solution that effectively bridges the gap between theoretical rigor and practical application in reinforcement learning, setting new standards for future RL algorithm design.

Future directions in both practical and theoretical domains are highlighted here. On the practical side, the hypermodel's compatibility with any feedforward neural network architecture offers seamless integration into a wide array of deep reinforcement learning frameworks, including actor-critic structures and transformer-based large models. This flexibility enhances its utility across various applications, such as foundation models, large language models (LLMs), and vision-language models (VLMs). Exploring these integrations could yield significant advancements. 
Theoretically, the prospect of extending our analysis to include linear, generalized linear, and neural function approximations with stochastic gradient descent (SGD) updates opens up a promising field for future studies. This exploration could deepen our understanding of the underlying mechanisms and improve the model's efficacy and applicability in complex scenarios, further bridging the gap.

The study of scalable posterior inference and uncertainty estimation for both exploration and alignment is of great importance.
First, utilizing hypermodel for uncertainty-aware reward modeling is promising for mitigating reward hacking in offline alignment problem and facilitate active feedback query in online alignment problem.
Second, it is possible to derive both more efficient alignment algorithm via \HyperAgent, performing approximate posterior sampling over $Q^\star$.
Extending large foundation models to solve multi-stage sequential decision tasks is also promising.

\clearpage

\section*{Acknowledgements}
The author would like to thank David Janz for reviewing the manuscript and providing feedback on writing.
The work of Y. Li was supported by the Internal Project Fund from Shenzhen Research Institute of Big Data under Grants J00220240001.
The work of Z.-Q. Luo was supported by the Guangdong Major Project of  Basic and Applied Basic Research (No.2023B0303000001), the Guangdong Provincial Key Laboratory of Big Data Computing, and the National Key Research and Development Project under grant 2022YFA1003900.
\section*{Impact statement}
\texttt{HyperAgent} represents a major advancement in reinforcement learning (RL). Its applications span gaming, autonomous vehicles, robotics, healthcare, financial trading, energy production, and more. With its computation- and data-efficient design, businesses can use \texttt{HyperAgent} for real-time decision-making, optimizing efficiency and outcomes under resource constraints.

Education and research in machine learning (ML) will benefit from \texttt{HyperAgent}. Its simplicity allows easy implementation, facilitating learning and experimentation for researchers and students, and accelerating advancements in the field. This tool could become vital in academic research and corporate R\&D, driving discoveries and breakthroughs in AI. The ease of implementation can democratize access to RL algorithms, fostering innovation and growth.

Smaller organizations and startups, often limited by computing resources, could leverage \texttt{HyperAgent}'s scalability and performance, creating a level playing field and encouraging creative innovation.

However, ethical considerations are crucial. The handling of large-scale interaction data requires robust policies to ensure user privacy and data protection. Efficiently managing large data sets heightens privacy concerns. Autonomous decision-making with RL must be monitored to prevent harmful behavior. Ethical implementation and continuous monitoring are essential to ensure fairness, safety, and security.

\bibliography{ref}
\bibliographystyle{icml2024}

\newpage
\appendix
\onecolumn

\newpage
\appendix

\section{Additional discussion on related works}
\label{sec:related}
Our work represents sustained and focused efforts towards developing principled RL algorithms that are practically efficient with function approximation in complex environments.

\paragraph{Discussion on the algorithmic simplicity and deployment efficiency.}
To address data efficiency, recent deep RL algorithms have incorporated increasingly complex heuristic and algorithmic components, such as DDQN \citep{van2016deep}, Rainbow \citep{hessel2018rainbow}, EfficientZero \citep{ye2021mastering}, and BBF \citep{schwarzer2023bigger}. Furthermore, their practical efficiency often falls short due to high per-step computational costs, exemplified by BBF's use of larger networks and more complex components that require careful tuning and may challenge deployment in real-world settings.
Several works~\citep{hessel2018rainbow,van2019use,schwarzer2023bigger} employ a combination of techniques, including dueling networks~\citep{wang2016dueling}, reset strategy~\citep{nikishin2022primacy}, distributional RL~\citep{bellemare2017distributional}, and others, to achieve data efficiency.
Some have demonstrated remarkable performance on Atari games.
However, integrating multiple techniques makes the algorithms
complicated
and challenging to apply to various scenarios. It requires careful selection of hyperparameters for each technique. For example, the reset frequency in the reset strategy needs meticulous consideration. Furthermore, combining multiple techniques results in a significant computational cost. For instance, BBF~\citep{schwarzer2023bigger} designs a larger network with 15 convolutional layers, which has 20 times more parameters than our method.
Model-based RL~\citep{kaiser2019model,schrittwieser2020mastering,ye2021mastering,hafner2023mastering} is a widely used approach for achieving data efficiency.
However, the performance of these methods is contingent upon the accuracy of the learned predictive model. Furthermore, the learning of predictive models can incur higher computational costs, and employing tree-based search methods with predictive models may not offer sufficient exploration.
Other methods~\citep{schwarzer2020data,laskin2020curl,liu2021aps} achieve data efficiency by enhancing representation learning. While these approaches perform well in environments with image-based states, they show poorer performance in environments characterized by simple structure yet requiring deep exploration, as seen in DeepSea.

\begin{table}[H]
  \centering
  \begin{tabular}{l|l}
    \toprule[1.2pt]
    Algorithm          & Components                                                                    \\
    \midrule % \midrule
    DDQN               & Incremental SGD with experience replay and target network                     \\
    \hline
    Rainbow            & (DDQN) + Prioritized replay, Dueling networks, Distributional RL, Noisy Nets. \\
    \hline
    \multirow{2}*{BBF} & (DDQN) + Prioritized replay, Dueling networks, Distributional RL,             \\
                       & Self-Prediction, Harder resets, Larger network, Annealing hyper-parameters.   \\
    \hline
    {\HyperAgent }     & (DDQN) + \textbf{hypermodel}                                                  \\
    \toprule[1.2pt]
  \end{tabular}
  \caption{The extra techniques used in different algorithms, e.g. DDQN \citep{van2016deep}, Rainbow \citep{hessel2018rainbow}, BBF \citep{schwarzer2023bigger} and \HyperAgent.}
  \label{tb:tech_simple}
\end{table}

\paragraph{Other Principled Exploration Approaches.}
Exploration strategies such as "Optimism in the Face of Uncertainty" (OFU) \citep{lai1985asymptotically} and Information-Directed Sampling (IDS) \citep{russo2018learning} also play crucial roles. OFU, efficient in tabular settings, encompasses strategies from explicit exploration in unknown states ($E^3$, \citet{kearns2002near}) to bonus-based and bonus-free optimistic exploration \citep{jaksch2010near,tiapkin2022dirichlet,liang2022bridging}. However, OFU also encounter computational hurdles in RL with general function approximation, leading to either intractability or unsustainable resource demands as data accumulates \citep{jiang2017contextual,jin2021bellman,du2021bilinear,foster2021statistical,liu2023maximize,wang2020reinforcement,agarwal2023vo}. IDS, while statistically advantageous and tractable in multi-armed and linear bandits, lacks feasible solutions for RL problems in tabular settings \citep{russo2018learning}.

\paragraph{Ensemble-based methods.}
\citet{osband2015bootstrapped,osband2016deep} initiated the bootstrapped ensemble methods, an incremental version of randomized value functions~\citep{wen2014efficient}, on bandit and deep RL, maintaining an ensemble of point estimates, each being incremental updated. This algorithm design methodology avoid refit a potentially complex model from scratch in the online interactive decision problems.
Bayes-UCBVI~\citep{tiapkin2022dirichlet} was extended to Incre-Bayes-UCBVI~\citep{tiapkin2022dirichlet} using the exact same idea as Algorithm 5 in \citep{osband2015bootstrapped} and then extended to Bayes-UCBDQN~\citep{tiapkin2022dirichlet} following BootDQN~\citep{osband2016deep}. As reported by the author, Bayes-UCBDQN shares very similar performance as BootDQN~\citep{osband2016deep} but requires addition algorithmic module on artificially generated pseudo transitions and pseduo targets, which is environment-dependent and challenging to tune in practice as mentioned in appendix G.3 of \citep{tiapkin2022dirichlet}.
Ensemble+  \citep{osband2018randomized,osband2019deep} introduces the randomized prior function for controlling the exploration behavior in the initial stages, somewhat similar to optimistic initialization in tabular RL algorithm design, facilitate the deep exploration and data efficiency. This additive prior design principle is further employed in a line of works \citep{dwaracherla2020hypermodels,li2022hyperdqn,osband2023epistemic,osband2023approximate}.
For a practical implmentation of LSVI-PHE~\cite{ishfaq2021randomized}, it utilizes the optimistic sampling (OS) with ensemble methods as a heuristic combination: it maintains an ensemble of $M$ value networks $\{Q_{i}(s, a), i= 1, \ldots, M\}$ and take greedy action according to the maximum value function over $M$ values $Q(s, a) = \max_{i \in [M]}Q_i(s, a)$ for action selection and $Q$-target computation.
As will be discussed in \cref{sec:index-sampling-schemes}, we propose another index sampling scheme called optimistic index sampling (OIS). OIS, OS and quantile-based Incre-Bayes-UCBVI are related in a high level, all using multiple randomized value functions to form a optimistic value function with high probability, thus leading to OFU-based principle for deep exploration.
Critical distinction exists, compared with ensemble-based OS and Incre-Bayes-UCBVI, OIS is computationally much more friendly\footnote{See detailed descriptions and ablation studies in \cref{sec:index-sampling-schemes}.} due to our continuous reference distribution $P_{\xi}$ for sampling as many indices as possible to construct randomized value functions.

% \todo[inline]{practical implementation of LSVI-PHE.}

Theoretical analysis of ensemble based sampling is rare and difficult. As pointed out by the first correct analysis of ensemble sampling for linear bandit problem~\citep{qin2022analysis}, the first analysis of ensemble sampling in 2017 has technical flaws. The results of \citet{qin2022analysis} show Ensemble sampling, although achieving sublinear regret in ($d$-dimensional, $T$-steps) linear bandit problems, requires $\tilde{O}(T)$ per-step computation, which is unsatisfied for a scalable agent with bounded resource.
Because of the potential challenges, there is currently no theoretical analysis available for ensemble-based randomized value functions across any class of RL problems.

\paragraph{Langevin Monte-Carlo.}
Langecin Monte-Carlo (LMC), staring from SGLD~\citep{welling2011bayesian}, has  huge influence in Bayesian deep learning and approximate Bayesian inference. However, as discussed in many literature~\citep{osband2023epistemic}, the computational costs of LMC-based inference are prohibitive in large scale deep learning systems.
Recent advances show the application of LMC-based posterior inference for sequence decision making, such as LMCTS~\citep{xu2022langevin} for contextual bandits as well as
LangevinDQN~\citep{dwaracherla2020langevin} and
LMC-LSVI~\citep{ishfaq2024provable} for reinforcement learning. As we will discuss in the following, these LMC-based TS schemes still suffer scalability issues as the per-step computational complexity would grow unbounded with the increasingly large amount of interaction data.

LMCTS~\citep{xu2022langevin} provides the first regret bound of LMC based TS scheme in ($d$-dimensional, $T$-steps) linear bandit problem, showing $\tilde{O}(d^{3/2}\sqrt{T})$ regret bound with
$\kappa_t \log (3 \sqrt{2 d T \log (T^3 )})$ inner-loop iteration complexity within time step $t$. As discussed in \citep{xu2022langevin}, the conditional number $\kappa_t = O(t)$ in general.
For a single iteration in time step $t$, LMC requires $O(d^2)$ computation for a gradient calculation of loss function:
$\nabla L_{t}(\theta) = 2(V_t \theta - b_t)$ using notations in \cite{xu2022langevin};
and $O(d)$ computation on noise generation and parameters update (line 5 and 6 in Algorithm 1 \cite{xu2022langevin}).
Additional $d A$ computation comes from greedy action selection among action set $\mathcal{A}$ by first computing rewards with inner product and selecting the maximum. Therefore, the per-step computation complexity of LMCTS is $\tilde{O}(d^2T + d A)$, which scales polynomially with increasing number of interactions $T$. LMCTS is not provably scalable under resource constraints.

LMC-LSVI~\citep{ishfaq2024provable} applies similar methodologies and analytical tools as LMCTS~\citep{xu2022langevin}, providing $\tilde{O}(d^{3 / 2} H^{3 / 2} \sqrt{T})$ regret in the linear MDP ($d$-dimensional feature mappings and $H$-horizons and $K$ episodes where $T= KH$).
The inner-loop iteration complexity of LMC within one time step $(k, h)$ of episode $k$ is $2\kappa_k \log(4HKd)$ . Similarly, $\kappa_k = O(k)$ in general.
(1) In the general feature case:
For a single iteration in time step $(k, h)$,
LMC requires $O(d^2)$ computation cost for the gradient calculation as from equation (6) of \citep{ishfaq2024provable} and $O(d)$ computation cost for noise generation and parameter update.
The per-step computational complexity caused by LMC inner-loops is $O(d^2 \kappa_k \log(4HKd)) = O(d^2 k \log (HKd))$.
Additional per-step computation cost in episode $k$ is $\tilde{O}(d^2 A k)$, coming from LSVI as discussed in \citep{jin2020provably}.
Therefore, the per-step computation complexity is $\tilde{O}(d^2 K \log (HKd) + d^2 A K)$, scaling polynomially with increasing number of episodes $K$.
(2) When consider tabular representation where the feature is one-hot vector with $d = SA$, the per-step computational complexity caused by LMC is $O(SA K \log(SAHK))$ as the covariance matrix is diagonal in the tabular setting.
The computation cost by LSVI is now $O(S^2 A)$ with no dependence on $K$ as we can perform incremental counting and bottom-up dynamic programming in tabular setting.
The per-step computational complexity of LMC-LSVI is $O(SA K \log(SAHK) + S^2 A)$, still scaling polynomially with increasing number of episodes $K$. Thus, LMC-LSVI is not provably scalable under resource constraints.

\citet{ishfaq2024provable} also extends their LMC-LSVI to deep RL setting, with a combination of Adam optimization techniques, resulting the AdamLMCDQN. It introduces additional hyper-parameters, such as the bias term and temperature, to tune as shown in Algorithm 2 of \citep{ishfaq2024provable}. As discussed in \citep{ishfaq2024provable}, AdamLMCDQN is sensitive to the bias term and tuned over a set of hyper-parameters for different Atari environments, showing its deployment difficulty. As shown in \cref{fig:atari_hardest}, our \HyperAgent, using a single set of hyper-parameters for all Atari environments, performs much better than AdamLMCDQN (LangevinMC) in all 8 hardest exploration Atari environments.

\paragraph{Heuristics on noise injection.}
For example, Noisy-Net~\citep{fortunato2017noisy} learns noisy parameters using gradient descent, whereas \cite{plappert2017parameter} added constant Gaussian noise to the parameters of the neural network.
SANE~\citep{aravindan2021state} is a
variational Thompson sampling
approximation for DQNs which uses a deep network whose parameters are perturbed by a learned variational noise distribution. Noisy-Net can be regarded as an approximation to the SANE.
While Langevin Monte-Carlo may seem simliar to Noisy-Net~\citep{fortunato2017noisy,plappert2017parameter} or SANE~\citep{aravindan2021state} due to their random perturbations of neural network weights in state-action value and target value computation, as well as action selection.
Critical differences exist, as noisy networks are not ensured to approximate the posterior distribution~\citep{fortunato2017noisy} and do not achieve deep exploration~\citep{osband2018randomized}. SANE~\citep{aravindan2021state} also lacks rigorous guarantees on posterior approximation and deep exploration.

\section{Reproducibility}
\label{sec:reproduce}

In support of other researchers interested in utilizing our work and authenticating our findings, we offer the implementation of \HyperAgent at the link \url{https://github.com/szrlee/HyperAgent}. This repository includes all the necessary code to replicate our experimental results and instructions on usage. 
Following this, we will present the evaluation protocol employed in our experiments and the reproducibility of the compared baselines.

\subsection{Evaluation protocol}
\label{appendix:eval_protocol}

\paragraph{Protocol on DeepSea.}
We replicate all experiments on DeeSea using 10 different random seeds. In each experiment run, we set the maximum episode to 10000 and evaluate the agent 100 times for every 1000 interactions to obtain the average return. The experiment can stop early when the average return reaches 0.99, at which point we record the number of interactions used by the agent. We then collect 10 data points (the number of interactions) for a specific problem size $N$, which are then utilized to generate a scatter plot in~\cref{fig:deepsea_baseline,fig:deepsea_Ms,fig:deepsea_xi,fig:deepsea_ois,fig:deepsea_m16,fig:deepsea_nps,fig:deepsea_sigma,fig:deepsea_net,fig:deepsea_de}. 
In \cref{fig:deepsea_baseline}, we perform linear regression on \HyperAgent's data to establish the dashed line and calculate the $R^2$ value indicating the goodness of fit. In \cref{fig:deepsea_Ms,fig:deepsea_xi,fig:deepsea_ois,fig:deepsea_m16,fig:deepsea_nps,fig:deepsea_sigma,fig:deepsea_net,fig:deepsea_de}, we calculate the average over 10 data points to construct the polyline.

\paragraph{Protocol on Atari.}
For the experiments on Atari, our training and evaluation protocol follows the baseline works \citep{mnih2015human,van2016deep,kaiser2019model,li2022hyperdqn}
During the training process, we assess the agent 10 times after every 20,000 interactions to calculate the average return at each checkpoint. This allows us to obtain 20 data points for each game at every checkpoint, as we repeat the process with 20 different random seeds.
We then compute the mean and 90$\%$ confidence interval range for plotting the learning curve in~\cref{fig:atari_hardest,fig:atari_curve,fig:atari_hardest_more}.

We calculate the best score for each game using the following steps:
(1) for each Atari game, the algorithm is performed with 20 different initial random seeds; 
(2) The program with one particular random seed will produce one best model (the checkpoint with highest average return), leading to 20 different models for each Atari game; (3) We then evaluate all 20 models, each for 200 times;
(4) We calculate the average score from these 200 evaluations as the score for each model associated with each seed.
(5) Finally, we calculate and report the average score across 20 seeds as the final score for each Atari game.
We follow the aforementioned five-step protocol to determine the score for all 55 Atari games outlined in \cref{tab:atari_score,tab:atari_score_other}.
We then utilize the 20 scores for each of the 26 Atari games to compute the Interquartile Mean (IQM), median, and mean score, along with the 95\% confidence interval\footnote{Using the evaluation code from the paper~\citep{agarwal2021deep} available at \url{https://github.com/google-research/rliable}.} for our algorithms, as depicted in \cref{tb:atari_pf,tab:atari_variant}.

\subsection{Reproducibility of baselines}
\label{appendix:repo_baselines}

\paragraph{Experiments on DeepSea.} We present our HyperModel and ENNDQN implementation and utilize the official HyperDQN implementation\footnote{ \url{https://github.com/liziniu/HyperDQN}} to replicate results. We kindly request the implementation of Ensemble+ from the author of \citet{li2022hyperdqn}, and they employ the repository on behavior suite\footnote{\url{https://github.com/google-deepmind/bsuite}} for reproduction.
We perform ablation studies in DeepSea benchmarks for a comprehensive understanding of \HyperAgent in \cref{appendix:additional_res_deepsea}.

\paragraph{Experiments on Atari.} We provide our version of DDQN(ours) and replicated the results using a well-known repository\footnote{\url{https://github.com/Kaixhin/Rainbow}} for DER. We obtained the result data for DDQN$^\dag$ and Rainbow from DQN Zoo\footnote{\url{https://github.com/google-deepmind/dqn_zoo}}. As they were based on 200M frames, we extracted the initial 20M steps from these results to compare them with \HyperAgent.
We acquired the official implementation of BBF\footnote{\url{https://github.com/google-research/google-research/tree/master/bigger_better_faster}} and EfficientZero\footnote{\url{https://github.com/YeWR/EfficientZero}} from their official repositories, and reached out to the author of \citet{li2022hyperdqn} for the results data of Ensemble+ with an ensemble size\footnote{For guidance on selecting the ensemble size, refer to Appendix C.4 in \citep{li2022hyperdqn}.} of 10. For the experiments about exploration on Atari (refer to Figure~\ref{fig:atari_hardest} and Figure~\ref{fig:atari_hardest_more}), we utilized the official implementation to replicate the results of SANE\footnote{\url{https://github.com/NUS-LID/SANE}}. and we obtained the raw result data of Ensemble+\footnote{It shares the same implementation as \citet{li2022hyperdqn}.}, AdamLMCDQN and LangevinAdam from official repository of LMC-LSVI\footnote{\url{https://github.com/hmishfaq/lmc-lsvi}}.
In addition to the results shown in the main article, we also provide fine-grained studies on Atari suite in \cref{appendix:additional_res_atari}.

\subsection{Environment Settings}
\label{appendix:env_settings}

\begin{figure}[htbp]
    \begin{minipage}[h]{0.45\textwidth}
        \centering
        \includegraphics[width=0.62\linewidth]{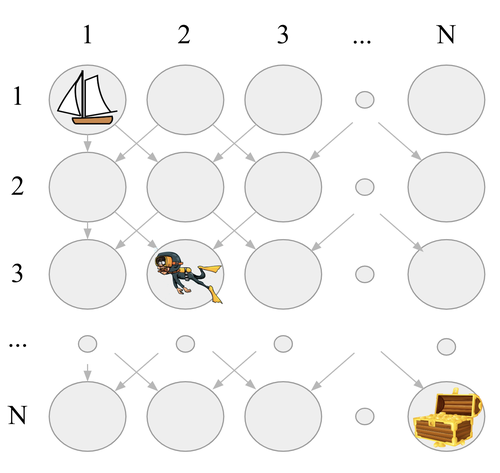}
        \caption{Illustration for DeepSea.}
        \label{fig:deepsea_env}
    \end{minipage}
    \begin{minipage}[h]{0.45\textwidth}
        \centering
        \begin{tabular}{l|r}
            \toprule[1.2pt]
            Hyper-parameters          & Setting  \\
            \midrule \midrule
            Grey scaling              & True     \\
            Sticky action             & False     \\
            Observation down-sampling & (84, 84) \\
            Frames stacked            & 4        \\
            Action repetitions        & 4        \\
            Reward clipping           & [-1, 1]  \\
            Terminal on loss of life  & True     \\
            Max frames per episode    & 108K     \\
            \toprule[1.2pt]
        \end{tabular}
        \captionof{table}{Detailed settings for Atari games}
        \label{tab:atari_setting}
    \end{minipage}
\end{figure}

In this section, we describe the environments used in experiments.
We firstly use the DeepSea~\citep{osband2019behaviour,osband2019deep} to demonstrate the exploration efficiency and scalability of \HyperAgent.
DeepSea (see~\cref{fig:deepsea_env}) is a reward-sparse environment that demands deep exploration~\citep{osband2019deep}.
The environment under consideration has a discrete action space consisting of two actions: moving left or right. During each run of the experiment, the action for moving right is randomly sampled from Bernoulli distribution for each row. Specifically, the action variable takes binary values of 1 or 0 for moving right, and the action map is different for each run of the experiment.
The agent receives a reward of 0 for moving left, and a penalty of $-(0.01/N )$ for moving right, where $N$ denotes the size of DeepSea. The agent will earn a reward of 1 upon reaching the lower-right corner.
The optimal policy for the agent is to learn to move continuously towards the right.
The sparse rewards and states presented in this environment effectively showcase the exploration efficiency of \HyperAgent without any additional complexity.

For the experiments on the Atari games, 
we utilized the standard wrapper provided by OpenAI gym.
Specifically, we terminated each environment after a maximum of 108K steps without using sticky actions. For further details on the settings used for the Atari games, please refer to~\cref{tab:atari_setting}.

\section{\HyperAgent details}
\label{appendix:impl_hyperfqi_deep}
In this section, we describe more details of the proposed \HyperAgent.
First, we describe the general treatment for the incremental {\update} function (in line 11 of \HyperAgent) in the following \cref{alg:update-dqn} of \cref{sec:incremental-update-mechanism}.
Then, we provide the details of index sampling schemes in \cref{sec:index-sampling-schemes}.
Next, in \cref{sec:hyperfqi-dnn}, we provide the implementation details of \HyperAgent with deep neural network (DNN) function approximation.
We want to emphasize that all experiments done in this article is using \textbf{Option} (1) with DNN value function approximation.
In \cref{sec:hyperfqi-tabular}, we describe the closed-form update rule (\textbf{Option} (2)) when the tabular representation of the value function is exploited. Note that the tabular version of \HyperAgent is only for the clarity of analysis and understanding.

\subsection{Incremental \update mechanism of \HyperAgent}
\label{sec:incremental-update-mechanism}
\begin{algorithm}
    \caption{\texttt{update}}
    \label{alg:update-dqn}
    \begin{algorithmic}[1]
        \STATE{{\bfseries Input:} $\operatorname{buffer} D$, $\theta, \theta^-, \vxi^-$, agent step $t$, train step $j$}
        \IF{$t \mod \operatorname{training\_freq} = 0$}
        \REPEAT
        \STATE{ Obtain $\theta$ by optimizing the loss $L^{\gamma, \sigma, \beta}(\theta; \theta^-, \vxi^-, D)$ in ~\cref{eq:hyperfqi}:
            \par
            \textbf{-- Option} (1) with gradient descent w.r.t. the mini-batch sampled loss~\cref{eq:hyperfqi-sampled}; (\HyperAgent)
            \par
            \textbf{-- Option} (2) with closed-form update in~\cref{eq:noise-incremenal,eq:bellman-iteration,eq:bellman-hyperfqi}.} (Tabular \HyperAgent)
        \STATE{Increment $j \leftarrow j + 1$}
        \IF{ $(j \mod \operatorname{target\_update\_freq}) = 0$}
        \STATE{ $\theta^- \leftarrow \theta$ }
        \ENDIF
        \UNTIL{ $(j \mod \operatorname{sample\_update\_ratio} \times \operatorname{training\_freq}) =0$}
        \ENDIF
        \STATE \textbf{Return:} $\theta, \theta^{-}, j$.
    \end{algorithmic}
\end{algorithm}
Notice that in \update, there are three important hyper-parameters $( \operatorname{target\_update\_freq}$, $\operatorname{sample\_update\_ratio}$, $\operatorname{training\_freq})$, which we will specify in \cref{tab:hyper_parameter} the hyper-paramters for practical implementation of \HyperAgent with DNN function approximation for all experimental studies; 
and in \cref{tab:hyper_parameter_tabular} the hyper-parameters only for regret analysis in finite-horizon tabular RL with fixed horizon $H$. To highlight, We have not seen this level of unification of algorithmic \update rules between practice and theoretical analysis in literature!

\begin{table}[h]
    \centering
    \begin{tabular}{l|r|r}
        \toprule[1.2pt]
        Hyper-parameters                                  & Atari Setting & DeepSea Setting \\
        \midrule %
        weight decay $\beta$            & 0.01        &  0\\
        discount factor $\gamma$                          & 0.99          & 0.99            \\
        learning rate                                     & 0.001         & 0.001           \\
        mini-batch size $\abs{\tilde{D}}$                  & 32            & 128             \\
        index dim $M$                                     & 4             & 4               \\
        \# Indices $\NpS$ for approximation               & 20            & 20              \\
        Perturbation std. $\sigma$           & 0.01       &   0.0001\\
        $n$-step target                                   & 5             & 1               \\
        $\operatorname{target\_update\_freq}$ in \update  & 5             & 4               \\
        $\operatorname{sample\_update\_ratio}$ in \update & 1             & 1               \\
        $\operatorname{training\_freq}$ in \update        & 1             & 1               \\
        hidden units                                      & 256           & 64              \\
        min replay size for sampling                      & 2,000 steps   & 128 steps       \\
        memory size                                       & 500,000 steps & 1000000 steps   \\
        \bottomrule[1.2pt]
    \end{tabular}
    \caption{Hyper-parameters of \HyperAgent. Other hyper-parameters used for Atari suite are the same as Rainbow~\citep{hessel2018rainbow}.
Note that we utilize a \textbf{single configuration} for all 55 games from Atari suite and a \textbf{single configuration} for DeepSea with varying sizes.}
    \label{tab:hyper_parameter}
\end{table}

For the approximation of expectation in \cref{eq:hyperfqi}, we sample multiple indices for each transition tuple in the mini-batch and compute the empirical average, as described in \cref{eq:hyperfqi-sampled}.
Recall that $\NpS$ is the number of  indices for each state and we set $\NpS = 20$ as default setting.
We have demonstrated how the number of indices $\NpS$ impacts our method in~\cref{sec:abltaion_implementation}.

\subsection{Index sampling schemes of \HyperAgent}
\label{sec:index-sampling-schemes}
Let us review the loss function.
For a transition tuple $d = (s, a, r, s', \rvz) \in D$ and given index $\xi$, the perturbed temporal difference (TD) loss $\ell^{\gamma, \sigma}(\theta; \theta^{-}, \vxi^-, \xi, d )$ is
\begin{align}
\label{eq:q-loss}
\left( r + \sigma \xi^\top \rvz + \gamma \max_{a' \in \mathcal{A}} f_{\theta^-}(s', a', \vxi^-(s')) - f_{\theta}(s, a, \xi) \right)^2,
\end{align}
where $\theta^-$ is the target parameters, and $\sigma$ is a hyperparameter to control the std of algorithmic perturbation.

We have two options for index sampling schemes for $\vxi_k(s)$:
\begin{enumerate}
    \item \textbf{State-dependent sampling.} As for implementation, especially for continuous or uncountable infinite state space:
          in the interaction, $\vxi_k(s)$ in the line 7 of \HyperAgent is implemented as
          independently sampling $\xi \sim P_{\xi}$ for each encountered state;
          for the target computation in \cref{eq:q-loss}, $\vxi_k(s')$ is implemented as independently sampling $\xi \sim P_{\xi}$ for each tuple $d = (s, a, r, s', \rvz)$ in the every sampled mini-batch.
    \item \textbf{State-independent sampling.} The implementation of state-independent $\vxi_k(s) = \xi_k$ is straightforward as we independently sample $\xi_k$ in the beginning of each episode $k$ and use the same $\xi_k$ for each state $s$ encountered in the interaction and for each target state $s'$ in target computation.
\end{enumerate}

In our implementation by default, \HyperAgent employs the state-independent $\vxi$ for action selection and utilizes state-dependent $\vxi$ for $Q$-target computation. 
The ablation results in~\cref{sec:ablation_sampling_schemes} demonstrate that these distinct index sampling schemes for $\vxi$ yield nearly identical performance.

\paragraph{Optimistic index sampling.}
\label{appendix:optimistic_index_sampling}
To make agent's behavior more optimistic with more aggresive deep exploration, in each episode $k$, we can sample $N_{\operatorname{OIS}}$ indices $\xi_{k,1}, \ldots, \xi_{k, N_{\operatorname{OIS}}}$ and take the greedy action according to the associated hypermodel:
\begin{align}
    \label{eq:OIS}
    a_k = \argmax_{a \in \mathcal{A}} \max_{n \in [N_{\operatorname{OIS}}]} f_{\theta}(s_k, a, \xi_{k, n}),
\end{align}
which we call optimistic index sampling (OIS) action selection scheme.

In the hypermodel training part, for any transition tuple $d = (s, a, r, s', \rvz)$ ,
we also sample multiple indices $\xi^-_{1}, \ldots, \xi^-_{N_{\operatorname{OIS}}}$ independently
and modify the target computation in \cref{eq:q-loss} as
\begin{align}
    \label{eq:q-loss-OIS}
    r + \sigma \xi^\top \rvz + \gamma \max_{a' \in \mathcal{A}} \max_{n \in [N_{\operatorname{OIS}}]} f_{\theta^-}(s', a', \xi_{n}^-(s')).
\end{align}
This modification in target computation boosts the propagation of uncertainty estimates from future states to earlier states, which is beneficial for deep exploration. We call this variant \HyperAgent w. OIS and compare it with \HyperAgent in~\cref{sec:abltaion_implementation}.
\HyperAgent w, OIS can outperform \HyperAgent, and the OIS method incurs minimal additional computation, as we have set $M=4$ and $N_{\operatorname{OIS}}=5$ in empirical studies.
Theoretically, leveraging this optimistic value estimation with OFU-based regret analysis, e.g. UCBVI-CH in \citep{azar2017minimax}, could lead a $O(H^2\sqrt{SAK})$ frequentist regret bound in finite-horizon time-inhomogeneous RL (\cref{asmp:finite-inhomogeneous-mdp}) \textbf{without} using \cref{assump:dirichlet-prior}.

\subsection{Function approximation with deep neural networks}
\label{sec:hyperfqi-dnn}
Here we describe the implementation details of \HyperAgent with deep neural networks and the main difference compared to baselines.

\subsubsection{{Hypermodel architecture in \HyperAgent}}
\label{sec:hypermodel-hyperagent}
\begin{figure}[!htbp]
    \centering
    \includegraphics[width=0.7\linewidth]{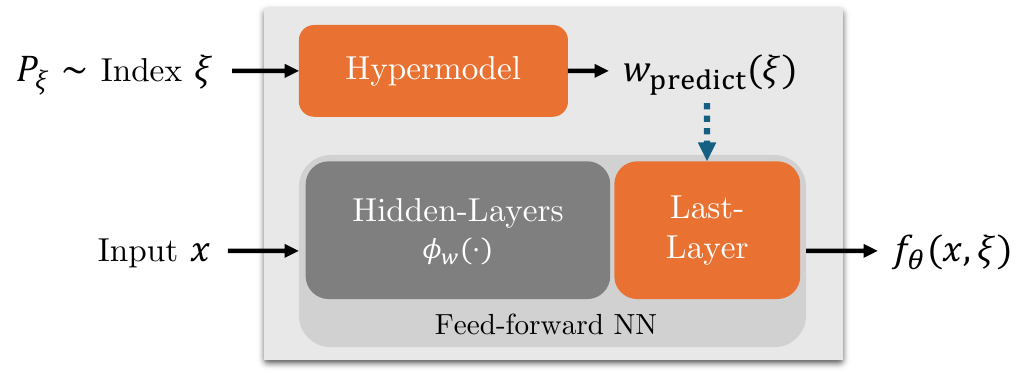}
    \caption[Last-layer linear hypermodel]{(\cref{fig:hypremodel} restated.) Description of the last-layer linear hypermodel: we made an assumption that the injected randomness only from the linear layer is sufficient for uncertainty estimation of feed-forward neural networks.}
\end{figure}

First, we develop a hypermodel for efficient approximate the posterior over the action-value function under neural network function approximation. As illustrated in \cref{fig:hypremodel}, we made assumptions that (1) \textbf{Base-model}: the action-value function is linear in the feature space even when the feature is unknown and needs to be learned through the training of neural network hidden layers; and 
(2) \textbf{Last-layer linear hypermodel}: the degree of uncertainty for base-model can be represented by a linear hypermodel transforming the index distribution to the approximation posterior distribution over the last-layer; and can be used for efficient deep exploration.

The (1) base-model assumption is common in supervised learning and deep reinforcement learning, e.g. DDQN\citep{mnih2015human,van2016deep}, BBF\citep{schwarzer2023bigger}. 

As the explanation of the (2) last-layer linear hypermodel assumption: for example, in \cref{fig:hypremodel}, suppose the hidden layers in neural networks forms the nonlinear feature mapping $\phi_{w}(\cdot)$ with parameters $w$.
    Our last-layer linear hypermodel assumption is formulated in \cref{eq:hyper_decompose}, with trainable $\theta = \{ \mA, b, w\}$ and fixed parameters $\{ \mA_0, b_0, w_0 \}$,
taking the random index $\xi \in \R^{M}$ from reference distribution $P_\xi$ as input and outputs the weights for last-layer.
\begin{align}
    \label{eq:hyper_decompose}
    f_{\theta}(x, \xi)
     & = \underbrace{\langle \mA \xi + b, \phi_{w}(x) \rangle}_{\textbf{Learnable~} f^L_{\theta}(x, \xi)} + \underbrace{\langle \mA_0\xi + b_0, \phi_{w_0}(x) \rangle}_{\textbf{Fixed prior~} f^P(x, \xi)} \nonumber                                                                            \\
     & = \underbrace{\langle \mA \xi, \phi_{w}(x) \rangle}_{\sigma_{\theta}^L(x, \xi)} + \underbrace{\langle \mA_0 \xi, \phi_{w_0}(x) \rangle}_{\sigma^P(x, \xi)} + \underbrace{\langle b, \phi_{w}(x) \rangle}_{\mu_{\theta}^L(x)} + \underbrace{\langle b_0, \phi_{w}(x) \rangle}_{\mu^P(x)}.
\end{align}
It's worth to note that our hypermodel only outputs the weights but not bias for last-layer.

(2) As shown in \cref{lem:approx}, we validate that the linear hypermodel with incremental update can approximate the posterior of action-value function in the sequential decision processes.
We conjecture that last layer linear hypermodel assumption is reasonable under neural network function approximation.
Through our formulation in \cref{eq:hyper_decompose}, \HyperAgent is supposed to accurately estimate the learnable mean $\mu_{\theta}^L(x)$, which relies solely on the original input $x$, and the variation prediction $\sigma^L_{\theta}(x, \xi)$, which is dependent on both the original input $x$ and random index $\xi$.
Since not being influenced by other components that may only depend on the random index $\xi$ like HyperDQN~\citep{li2022hyperdqn}, we conjecture our last-layer linear hypermodel assumption in \cref{eq:hyper_decompose} allows the hypermodel to capture uncertainty better.
Another benefit last-layer linear hypermodel is that this structure will not result in much parameters and provide better expectation estimate.

The fixed prior model also offers prior bias and prior variation through the functions $\mu^P(x)$ and $\sigma^P(x, \xi)$. This prior function is \textit{not} trainable so that it will not bring much computation, and designed to provide better exploration in the early stage of training.
We use Xavier normal initialization for the entire network except for the prior model. For the initialization of prior model, we follow the method described in \citep{li2022hyperdqn,dwaracherla2020hypermodels}. 
In this way, each row of prior function is sampled from the unit hypersphere, which guarantees that the output of prior function can follow a desired Gaussian distribution.

In the context of reinforcement learning, we define the action-value function with hypermodel and DNN approximation as following.
For each action $a \in \mathcal{A}$, there is a set of trainable parameters $\{\rmA^{(a)}, b^{(a)}\}$ and fixed parameters $\{ \rmA_0^{(a)}, b_0^{(a)} \}$, i.e.,
the trainable set of parameters $\theta = \{ w, (\rmA^{(a)}, b^{(a)}): a \in \mathcal{A} \}$ and the fixed one $\{ w_0, (\rmA_0^{(a)}, b_0^{(a)}): a \in \mathcal{A} \}$ with action-value function
\begin{align*}
    f_{\theta}(s, a, \xi) = \underbrace{\langle \rmA^{(a)} \xi + b^{(a)}, \phi_{w}(s) \rangle}_{\textbf{Learnable~} f^L_{\theta}(s, a, \xi)} + \underbrace{\langle \rmA^{(a)}_0\xi + b^{(a)}_0, \phi_{w_0}(s) \rangle}_{\textbf{Fixed prior~} f^P(s, a, \xi)}.
\end{align*}
The last-layer linear hypermodel assumption is further supported by the empirical results \cref{fig:deepsea_baseline,fig:atari_hardest} where hypermodel with incremental updates enables efficient deep exploration in RL.

\subsubsection{Difference compared to prior works}
\label{appendix:difference}
    Several related work can be included in the hypermodel framework introduced in \cref{sec:hypermodel}.
    We will discuss the structural and algorithmic differences under the unified framework in this sections. Furthermore, we performs ablation studies concerning these mentioned differences in \cref{sec:hypermodel_framework_diff}.

\paragraph{Difference with HyperModel~\citep{dwaracherla2020hypermodels}.}
HyperModel~\citep{dwaracherla2020hypermodels} employs hypermodel to represent epistemic uncertainty and facilitate exploration in bandit problem. However, implementing hypermodel across entire based-model results in a significant number of parameters and optimization challenges. As a reuslts, applying HyperModel to tackle large-scale problems, like Atari games, can prove to be exceedingly difficult, as highlighted in \citet{li2022hyperdqn}.
Additionally, HyperModel also encounters challenges in addressing the DeepSea problem due to its substantial state space, even when the size is relatively small, as demonstrated in~\cref{sec:hypermodel_framework_diff}.
In contrast, \HyperAgent offers the advantage of computation efficiency as it only applies hypermodel to the last output layer of base-model, which maintains constant parameters when scaling up the problem.  \HyperAgent aslo demonstrates superior data efficiency compared to HyperModel, as evidenced in~\cref{fig:deepsea_net}.

\paragraph{Structural difference with Ensemble+~\citep{osband2018randomized,osband2019deep}.} Ensemble+ applies the bootstrapped ensemble method to the entire based-model, which maintains an ensemble of $M$ value networks $\{Q_{i}(s, a), i= 1, \ldots, M\}$.
When combined with prior network, it has demonstrated effective exploration in chain environments with large sizes. 
Nevertheless, to achieve effective exploration, Ensemble+ demands a relatively large ensemble size $M$~\cite{osband2018randomized}, which raises challenges analogous to those faced by the HyperModel. This involves managing numerous parameters and optimization issues. 
We also evaluated Ensemble+ with a larger $M=16$, and it still proved ineffective in solving the DeepSea, as depicted in Figure~\ref{fig:deepsea_ois}, highlighting the superior data efficiency of \HyperAgent.

\paragraph{Structural difference with hypermodel in HyperDQN~\citep{li2022hyperdqn}.}
HyperDQN shares a similar structure with \HyperAgent and has demonstrated promising results in exploration. Nevertheless, it struggles to handle the DeepSea, which requires deep exploration, as depicted in~\cref{fig:deepsea_baseline}.
We enhanced HyperDQN by simplifying the hypermodel, as demonstrated in \cref{eq:hyper_decompose}. \HyperAgent estimates the mean $\mu$ exclusively from the original input $x$ and estimates the variation $\sigma$ using both the original input $x$ and the random index $\vxi$. 
In the implementation of HyperDQN, there are  two linear hypermodel $f_{\theta_1}(\xi)$ and $f_{\theta_2}(\xi)$.
\begin{align*}
    f_{\theta_1}(\xi)     & = \mA_{1} \xi + b_1;                    & f_{\theta_2}(\xi)     & = \mA_{2} \xi + b_2.            \\
    f_{\theta_1^{P}}(\xi) & =  \mA^{P}_{1} \xi  + b^P_1; & f_{\theta_2^{P}}(\xi) & =  \mA^{P}_{2}  \xi  +  b^P_2.
\end{align*}
The hypermodel $f_{\theta_1}(\xi)$ outputs weights for last output layer of base-model, and hypermodel $f_{\theta_2}(\xi)$ outputs bias for last output layer of base-model. The functions $f_{\theta^P_1}(\xi)$ and $f_{\theta^P_2}(\xi)$ are the prior network corresponding to trainable linear hypermodel.
These linear hypermodel contain trainable $\theta_1 = \{ \mA_1, b_1\}$, $\theta_2 = \{ \mA_2, b_2\}$ and fixed parameters $\theta^P_1 = \{ \mA^P_1, b^P_1\}$, $\theta^P_2 = \{ \mA^P_2, b^P_2\}$.
Therefore, the implementation of HyperDQN can be formulated by
\begin{align}
    f^{\operatorname{HyperDQN}}(x, \xi) & = \langle f_{\theta_1}(\xi), \phi_w(x) \rangle + f_{\theta_2}(\xi) + \langle f_{\theta_1^P}(\xi), \phi^P_w(x) \rangle  + f_{\theta^P_2}(\xi)  \nonumber \\  
            & = \underbrace{ \langle b_1, \phi_w(x) \rangle }_{\mu^L(x)} + 
            \underbrace{ \langle {b^P_1}, \phi^P_w(x) \rangle }_{\mu^P(x)} + \underbrace{\langle \mA_1 \xi, \phi_w(x) \rangle}_{\sigma_1^L(, \xi)} + \underbrace{\langle {\mA^P_1} \xi, \phi^P_w(x) \rangle}_{\sigma_1^P(, \xi)} + \nonumber \\  
            & \quad \quad \underbrace{\mA_2 \xi}_{\sigma_2^L(\xi)} + \underbrace{\mA^P_2 \xi }_{\sigma_2^P(\xi)} + \underbrace{{b_2}}_{\mu^L_3} + \underbrace{{b^P_2}}_{\mu^P_3}.
    \label{eq:hyperdqn_decompose}
\end{align}

As demonstrated in~\cref{eq:hyperdqn_decompose}, HyperDQN utilizes the hypermodel to generate both weights and bias for the output layer, leading to redundant components, such as functions ($\sigma_2^L(\xi)$, $\sigma_2^P(\xi)$) that rely solely on the random index $\vxi$, or functions ($\mu^L_3$, $\mu^P_3$) that do not depend on any inputs. These components lack a clear semantic explanation.
We also found that initializing the hypermodel with Xavier Normal can improve optimization. 
These modifications are some of the factors leading to \HyperAgent outperforming HyperDQN on both DeepSea and Atari games, as demonstrated in~\cref{sec:exp_study}.

\paragraph{Structural difference with epinet in ENNDQN~\citep{osband2023epistemic,osband2023approximate}.}
ENNDQN~\cite{osband2023approximate}, leveraging the epinet~\cite{osband2023epistemic} structure, exhibits potential in capturing epistemic uncertainty and has showcased effectiveness across diverse tasks.
A notable difference is that ENNDQN use ``stop gradient" between feature layers and epinet. This indicates that the error feedback from epinet will not be back-propagated to the feature layers.
Another difference is about epinet structure where the original input $x$, feature $\phi_{w}(x)$, and random index $\vxi$ are concatenated as the input of epinet. 
An ensemble prior function with size $M$ is used for the output layer, but a separated prior feature network is not present.
This network structure results in larger parameters when handling tasks at a large scale, creating notable computation and optimization challenges. 
For instance, in the case of DeepSea with a size of 20, the parameters of epinet are nearly 20 times larger than those of \HyperAgent. 
This is due to the raw state input $x \in \R^{N^2}$ for DeepSea with size $N$, whose dimension $N^2$ is too large for the epinet to effectively process.
As as results, epinet struggles with larger scale of the problem, as evidenced in~\cref{fig:deepsea_net}.
In contrast, \HyperAgent takes only a random index $\vxi$ and feature $\phi_w(x)$ as input for output layer, resulting in more efficient computation with fewer constant parameters.

\paragraph{Algorithmic difference with Ensemble+, HyperDQN and ENNDQN.}
For a transition tuple $d = (s, a, r, s', \rvz) \in D$ and given a single index $\xi$, with main hypermodel parameters $\theta$ and target hypermodel parameters $\theta^-$, the loss in these works can be represented using our notion of hypermodel:
\begin{itemize}
\item The temporal difference (TD) loss for Ensemble+~\citep{osband2018randomized,osband2019deep} inherits the BootDQN~\citep{osband2016deep},
\begin{align}
    \label{eq:q-loss-ensemble} \ell^{\gamma}_{\operatorname{Ensemble+}}(\theta; \theta^{-}, \xi, d )
    & = (\rvz^\top \xi) \left( f^{\operatorname{Ensemble+}}_{\theta}(s, a, \xi) - (r + \gamma \max_{a' \in \mathcal{A}} f^{\operatorname{Ensemble+}}_{\theta^-}(s', a', \xi) )\right)^2,
\end{align}
where $f^{\operatorname{Ensemble+}}$ is the ensemble network structure, $\xi \sim P_{\xi} := \mathcal{U}(\{e_1, \ldots, e_M\})$ where $e_i$ is the one-hot vector in $\R^M$ and $\rvz \sim P_{\rvz}$, where $\rvz_i$ sampled from $2 \cdot \operatorname{Bernoulli}(0.5)$ independently across entries $i \in [M]$.
\item The TD loss for HyperDQN~\citep{li2022hyperdqn} is
\begin{align}
    \label{eq:q-loss-hyperdqn} \ell^{\gamma, \sigma}_{\operatorname{HyperDQN}}(\theta; \theta^{-}, \xi, d )
    & = \left( f^{\operatorname{HyperDQN}}_{\theta}(s, a, \xi) - (r + \sigma \rvz^\top \xi \gamma + \max_{a' \in \mathcal{A}}  f^{\operatorname{HyperDQN}}_{\theta^-}(s', a', \xi) )\right)^2,
\end{align}
where $f^{\operatorname{HyperDQN}}$ is the network structure of HyperDQN, $\xi \sim P_{\xi} := \mathcal{U}(\mathbb{S}^{M-1})$  and $\rvz \sim P_{\rvz} := \mathcal{U}(\mathbb{S}^{M-1})$.
\item 
The TD loss for ENNDQN~\citep{osband2023approximate} is
\begin{align}
    \label{eq:q-loss-enn} \ell^{\gamma}_{\operatorname{ENNDQN}}(\theta; \theta^{-}, \xi, d )
    & = \left( f^{\operatorname{epinet}}_{\theta}(s, a, \xi) - (r + \gamma \max_{a' \in \mathcal{A}} f^{\operatorname{epinet}}_{\theta^-}(s', a', \xi) )\right)^2,
\end{align}
where $f^{\operatorname{epinet}}$ is the epinet structure used in ENNDQN, $\xi \sim P_{\xi} := N(0, I_M)$.
\end{itemize}
As discussed in this section, $f^{\operatorname{Ensemble+}}, f^{\operatorname{HyperDQN}}, f^{\operatorname{epinet}}$ can all be represented within our hypermodel framework and our construction of hypermodel in \HyperAgent has a few mentioned advantages.
The key difference in these loss functions \cref{eq:q-loss-ensemble,eq:q-loss-hyperdqn,eq:q-loss-enn} is that the index $\xi$ used for target computation is the same one as in main network;
while in \cref{eq:q-loss} of \HyperAgent, we choose a index mapping $\vxi^-$ in the target computation that is independent of $\xi$ used in main network. 

The choice for independent $\vxi^-$ is \textbf{critical} for theoretical analysis as closed-form solution under setting with fixed feature mapping can be derived with \cref{eq:q-loss} of \HyperAgent but can \textit{not} be derived with \cref{eq:q-loss-ensemble,eq:q-loss-hyperdqn,eq:q-loss-enn} of any previous related works.
The empirical difference for this issue will be discussed in ablation studies in \cref{sec:hypermodel_framework_diff}.
Another issue is that whether we need a additive perturbation.
As stated in the incremental update~\cref{eq:noise-incremenal}, the std of artificial perturbation $\sigma$ is important for our posterior approximation argument~\cref{lem:approx}.
However, in some practical problems with deterministic transitions, do we really need this level of perturbations i.e., $\sigma =0$ or $\sigma>0$? If we need $\sigma>0$, how large should it be?
This issue would be address in \cref{sec:abltaion_implementation}.

\subsection{Tabular representations}
\label{sec:hyperfqi-tabular}
To understand and analyze the behavior of \HyperAgent, we specify the algorithm in the tabular setups.
Notice that the closed-form iterative update rule derived here is general and can be applied for infinite-horizon and finite horizon problems.

Hypermodel in \HyperAgent with tabular representation would be 
\[
f_{\theta}(s, a, \xi) = \underbrace{\mu_{sa} + m_{sa}^\top \xi}_{\textbf{Learnable}~f^L_{\theta}(s, a, \xi)} + \underbrace{\mu_{0,sa} + \sigma_0 \rvz_{0, sa}^\top \xi}_{\textbf{Fixed prior}~f^P(s, a, \xi)}
\]
where  
$\theta = (\mu \in \mathbb{R}^{\abs{\mathcal{S}}\abs{\mathcal{A}}}, m \in \mathbb{R}^{\abs{\mathcal{S}}\abs{\mathcal{A}} \times M})$ are the parameters to be learned, 
    and $\rvz_{0, sa} \in \mathbb{R}^M$ is a independent random vector sampled from $P_{\rvz}$ and $\mu_{0, sa}, \sigma_0$ is a prior mean and prior variance for each $(s, a) \in \mathcal{S} \times \mathcal{A}$.
The tabular representation is related to the last-layer linear hypermodel assumption in \cref{eq:hyper_decompose} when the hidden layer maps each state $s$ to a fixed one-hot feature $\phi_w(s) = \phi_{w_0}(s) = \1{s} \in \R^{\abs{\mathcal{S}}}$.
The regularizer in \cref{eq:hyperfqi} then becomes $\beta \| \theta \|^2 = \beta \sum_{s, a} \left( \mu_{sa}^2 + \| m_{sa} \|^2 \right)$.

\paragraph{Derivations of \cref{eq:noise-incremenal,eq:bellman-iteration,eq:bellman-hyperfqi} in \cref{sec:analysis}.}
The derivation is mainly from the separability of optimization problem in tabular setup.
Let $\theta_{sa} = (\mu_{sa}, m_{sa})$ be the optimization variable for specific $(s, a) \in \mathcal{S} \times \mathcal{A}$.
As mentioned in \cref{sec:analysis}, at the beginning of episode $k$, with $D = \mathcal{H}_{k}$ and target noise mapping $\vxi^- = \vxi_k$, we iterative solve \cref{eq:hyperfqi} by taking target parameters $\theta^- = \theta^{(i)}_k$ as previous solved iterate starting from $\theta^{(0)}_{k} = \theta_{k-1} = (\mu_{k-1}, m_{k-1})$ as the solution for previous episode $k-1$.

Let the optimal solution in $(i)$-th iteration be $\theta^{(i+1)}_{k} = \argmin_{\theta} {L}^{\gamma, \sigma, \beta}(\theta; \theta^- = \theta^{(i)}_k, \vxi^- = \vxi_k, \mathcal{H}_k)$.
In tabular setting, by the separability of the objective function in \cref{eq:hyperfqi}, we have $\theta_{k, sa} = (\mu_{k, sa}, m_{k, sa}) =
    \argmin_{\theta_{sa}} L_{sa}(\theta_{sa}; \theta^- = \theta^{(i)}_k, \vxi^- = \vxi_k, \mathcal{H}_k))$ where $\argmin_{\theta_{sa}}L_{sa}(\theta_{sa}; \theta^-, \vxi^-, \mathcal{H}_k)$ is defined as
\begin{align*}
     \argmin_{\theta_{sa}} & \E[\xi \sim P_{\xi}]{ \sum_{\ell =1}^{k-1} \sum_{t \in E_{\ell, sa} }( f_{\theta}(S_{\ell, t}, A_{\ell, t}, \xi) - (  \sigma \xi^\top \rvz_{\ell, t+1} + y_{\ell, t+1}(\theta^-, \vxi^-) ) )^2 } + {\beta}(\mu_{sa}^2 + \norm{m_{sa}}^2) \\
    = \argmin_{\theta_{sa}} & \E[\xi \sim P_{\xi}]{ \sum_{\ell =1}^{k-1} \sum_{t \in E_{\ell, sa} }( f_{\theta}(s, a, \xi)- (  \sigma \xi^\top \rvz_{\ell, t+1} + y_{\ell, t+1}(\theta^-, \vxi^-) ) )^2 } + {\beta}(\mu_{sa}^2 + \norm{m_{sa}}^2) \\
    = \argmin_{(\mu_{sa}, m_{sa})} & \E[\xi \sim P_{\xi}]{ \sum_{\ell =1}^{k-1} \sum_{t \in E_{\ell, sa} }( (\mu_{sa} + \mu_{0, sa}) + (m_{sa} + \sigma_0 \rvz_{0, sa})^\top \xi - (  \sigma \xi^\top \rvz_{\ell, t+1} + y_{\ell, t+1}(\theta^-, \vxi^-) ) )^2 } \\
    & \quad \quad + {\beta}(\mu_{sa}^2 + \norm{m_{sa}}^2)
\end{align*}
where the target value is
\[
    y_{\ell, t+1} (\theta^-, \vxi^-) = R_{\ell,t+1} + \gamma \max_{a' \in \mathcal{A}} f_{\theta^-}(S_{\ell,t+1}, a', \vxi^-(S_{\ell, t+1})).
\]
With some calculations, the closed form solution of $\theta^{(i+1)}_k = (\mu_{k,sa}^{(i+1)}, m_{k, sa})$ is
\begin{align*}
    m_{k, sa} + \sigma_0 \rvz_{0, sa} 
    & = \frac{ \sigma \sum_{\ell = 1}^{k-1} \sum_{ t \in E_{\ell, sa} } \rvz_{\ell, t+1} + \beta \sigma_0 \rvz_{0, sa} }{N_{k, sa} + \beta} \\
    & = \frac{ (N_{k-1, sa} + \beta)(m_{k-1,sa} + \sigma_0 \rvz_{0, sa}) + \sigma \sum_{ t \in E_{k-1, sa} } \rvz_{k-1, t+1} }{N_{k, sa} + \beta},
\end{align*}
which derives the incremental update in \cref{eq:noise-incremenal}, and 
\begin{align}
    \label{eq:mu-iterative}
    \mu_{k, sa}^{(i+1)} = \frac{ \sum_{\ell = 1}^{k-1} \sum_{t \in E_{\ell, sa}} y_{\ell, t+1}(\theta^- = \theta_k^{(i)}, \vxi^- = \vxi_k) + \beta \mu_{0, sa} }{N_{k, sa} + \beta}.
\end{align}
Recall some short notations:
(1) $V_Q$ is greedy value with respect to $Q$ such that $V_{Q}(s) = \max_{a\in \mathcal{A}} Q(s, a)$ for all $s \in \mathcal{S}$;
(2) $f_{\theta, \vxi}(s, a)= f_{\theta}(s, a, \vxi(s)), \forall (s,a) \in \mathcal{S} \times \mathcal{A}$.
Recall the stochastic bellman operator $F_{k}^{\gamma}$ induced by \HyperAgent,
\begin{align*}
    F_{k}^{\gamma} Q (s, a)
     & := \frac{\beta \mu_{0, sa} + N_{k, sa}(r_{sa} + \gamma V_Q^\top \hat{P}_{k, sa})}{N_{k, sa} + \beta} + m_{k, sa}^\top \vxi_k(s), \quad \forall (s, a) \in \mathcal{S} \times \mathcal{A}.
\end{align*}
With the following observation,
\begin{align*}
    \sum_{\ell = 1}^{k-1} \sum_{t \in E_{\ell, sa}} y_{\ell, t+1}(\theta^-, \vxi^-)
     & = N_{k, sa} ( r_{sa} +  \gamma \sum_{s' \in \mathcal{S}} \hat{P}_{k, sa}(s') (\max_{a' \in \mathcal{A}} f_{\theta^-}(s', a', \vxi^-(s'))) ),
\end{align*}
from \cref{eq:mu-iterative}, we have for all pairs $(s, a)$
\begin{align*}
    f_{\theta_k^{(i+1)}}(s, a, \vxi_k) & = \mu_{k, sa}^{(i+1)} + m_{k, sa}^\top \vxi_k(s) = \frac{\beta \mu_{0, sa} + N_{k, sa} ( r_{sa} + \gamma V_{f_{\theta_k^{(i)}, \vxi_k}}^\top \hat{P}_{k, sa} ) }{N_{k, sa} + \beta} + m_{k, sa}^\top \vxi_k(s),
\end{align*}
which is essentially bellman iteration under stochastic bellman operator induced by \HyperAgent,
\begin{align*}
    f_{\theta_{k}^{(i+1)}, \vxi_k} = F^{\gamma}_{k} f_{\theta_{k}^{(i)}, \vxi_k}.
\end{align*}
\begin{lemma}[Contraction mapping]
    \label{lem:contraction}
    Let $B(\mathcal{S} \times \mathcal{A})$ be the space of bounded functions $Q: \mathcal{S} \times \mathcal{A} \rightarrow \mathbb{R}$.
    Let $\rho$ be the distance metric $\rho(Q, Q') = \sup_{ (s, a) \in \mathcal{S} \times \mathcal{A} } | Q(s, a) - Q'(s, a) | $.
    For all $k \in \mathbb{Z}_{++}$ the Bellman operator of \HyperAgent $F^{\gamma}_{k}: B(\mathcal{S} \times \mathcal{A}) \rightarrow B(\mathcal{S} \times \mathcal{A}) $ is a contraction mapping with modulus $\gamma \in [0, 1)$ in metric space $(B(\mathcal{S} \times \mathcal{A}), \rho)$.
\end{lemma}
By \cref{lem:contraction}, since contraction mapping, the bellman operator of \HyperAgent $F^{\gamma}_{k}$ has a unique fixed point and the iterative process in \cref{eq:bellman-iteration} can converge to a unique fixed point $\theta_{k}$. Essentially, due the algorithmic randomness introduced in the iterative process, $f_{\theta_k, \vxi_k}$ is a randomized state-action value function.
\begin{proof}[Proof of \cref{lem:contraction}]
    By Blackwell’s sufficient conditions, we need to show that $F^{\gamma}_{k}$ satisfies the following two conditions:
    \par
    1. Monotonicity: for all $Q, Q' \in B(\mathcal{S} \times \mathcal{A})$,
    if $Q(s, a) \le Q'(s, a)$ for all $(s, a) \in \mathcal{S} \times \mathcal{A}$, then
    \begin{align*}
        F_{k}^{\gamma} Q (s, a) \le F_{k}^{\gamma} Q'(s, a)
    \end{align*}
    2. Discounting: for all $(s, a) \in \mathcal{S} \times \mathcal{A}$, all $c \ge 0$ and $Q \in B(\mathcal{S} \times \mathcal{A})$,
    \begin{align*}
        [F^{\gamma}_{k} (Q + c)] (s, a) - [F^{\gamma}_{k} Q] (s, a)
        = \left( \frac{N_{k, sa} }{ N_{k, sa} + \beta } \right) \gamma c \le \gamma c
    \end{align*}
\end{proof}

\paragraph{True Bellman Operator.}
For any MDP $\cM = (\mathcal{S}, \mathcal{A}, r, P, \rho, s_{\operatorname{terminal}})$, consider a function \( Q \in B(\mathcal{S} \times \mathcal{A}) \).
The true Bellman operator, when applied to \( Q \), is defined as follows:
\begin{align*}
    F_{\cM}^{\gamma} Q (s, a)
     & = r_{sa} + \gamma V_Q^\top P_{sa}, \quad \forall (s, a)\in \mathcal{S} \times \mathcal{A}.
\end{align*}
As will be introduced later, under Dirichlet prior~\cref{assump:dirichlet-prior}, 
given the randomness in $P$, $F_{\cM}^{\gamma}$ is essentially stochastic in nature. When $F_{\cM}^{\gamma}$ acts upon a state-action value function \( Q \in B(\mathcal{S} \times \mathcal{A}) \), the result is a randomized state-action value function.

\section{Insight and Theoretical analysis of \HyperAgent}
\subsection{Sequential posterior approximation argument in \cref{lem:approx}}
\label{sec:proof-approx}
We use short notation for $[n] = \{1, 2, \ldots, n\}$ and $\mathcal{T} = \{ 0, 1, \ldots, T\} = \{ 0 \} \cup [T] $.
Before digging into the details of proof of our key lemma, let us first introduce some useful probability tools developed for random projection recently.
\begin{lemma}[Distributional JL lemma \citep{johnson1984extensions}]
  \label{lem:djl}
  For any $0<\varepsilon, \delta<1 / 2$ and $d \geq 1$ there exists a distribution $\mathcal{D}_{\varepsilon, \delta}$ on $\mathbb{R}^{M \times d}$ for $M=O\left(\varepsilon^{-2} \log (1 / \delta)\right)$ such that for any $\vx \in \mathbb{R}^d$
  \[
    \underset{\Pi \sim \mathcal{D}_{\varepsilon, \delta}}{\mathbb{P}}\left(\|\Pi \vx\|_2^2 \notin\left[(1-\varepsilon)\|\vx\|_2^2,(1+\varepsilon)\|\vx\|_2^2\right]\right)<\delta
  \]
\end{lemma}

\begin{theorem}[Sequential random projection in adaptive process \citep{li2024probability}]
  \label{thm:one-dimension-dependent-sequence}
  Let $\varepsilon \in (0, 1)$ be fixed and $(\mathcal{F}_{t})_{t\ge 0}$ be a filtration.
 Let $\rvz_0 \in \R^M$ be an $\mathcal{F}_0$-measurable random vector satisfies $\E{\norm{\rvz_0}^2} = 1$ and $\abs{\norm{\rvz_0}^2 - 1} \le (\varepsilon/2)$.
  Let $(\rvz_t)_{t \ge 1} \subset \R^M$ be a stochastic process adapted to filtration $(\mathcal{F}_{t})_{t \ge 1}$ such that it is $\sqrt{c_0/M}$-sub-Gaussian and each $\rvz_t$ is unit-norm.
  Let $(x_t)_{t\ge 1} \subset \R$ be a stochastic process adapted to filtration $(\mathcal{F}_{t-1})_{t \ge 1}$ such that it is $c_x$-bounded. Here, $c_0$ and $c_x$ are absolute constants.
  For any fixed $x_0 \in \R$, if the following condition is satisfied
    \begin{align}
    \label{eq:M-selection}
        M \ge \frac{16 c_0 (1+ \varepsilon)}{\varepsilon^2}  \left(\log \left( \frac{1}{\delta} \right) + \log \left(1 + \frac{c_x T}{x_0^2} \right) \right),
    \end{align}
  we have, with probability at least $1 - \delta$
  \begin{align}
  \label{eq:seq-rand-proj}
    \forall t \in \mathcal{T}, \quad
    (1 - \varepsilon) \left( \sum_{i=0}^t x_i^2 \right) \le \norm{\sum_{i=0}^t x_i \rvz_i}^2 \le (1 + \varepsilon) \left(\sum_{i=0}^t x_i^2 \right).
  \end{align}
\end{theorem}
\begin{remark}
\label{rem:seq-rand-proj}
  \citet{li2024probability} claims this is a
  ``sequential random projection'' argument because one can relate \cref{thm:one-dimension-dependent-sequence} to the traditional random projection~(\cref{lem:djl}) setting where $\Pi_t = (\rvz_0, \ldots, \rvz_t) \in \R^{M \times t+1}$ is a random projection matrix and $\vx_t = (x_0, \ldots, x_t)^\top \in \R^{t+1}$ is the vector to be projected.
    The argument in \cref{eq:seq-rand-proj} translates to
    \begin{align}
    \label{eq:seq-rand-proj-2}
        \forall t \in \mathcal{T}, \quad (1-\varepsilon) \norm{\vx_t}^2 \le \norm{ \Pi_t \vx_t }^2 \le (1+ \varepsilon) \norm{\vx_t}^2.
    \end{align}
    When assuming independence between $\vx_t$ and $\Pi_t$ for all $t \in \mathcal{T}$, by simply applying union bound over time index $t \in \mathcal{T}$ with existing JL analysis, we can derive that the required dimension $M = O( \varepsilon^{-2} \log (T/\delta))$ is of the same order in \cref{eq:M-selection}.
    \textbf{However}, as discussed in \citep{li2024probability}, existing JL analytical techniques are \textit{not} able to handle the sequential dependence in our setup as $\vx_t$ is statistically dependent with $\Pi_t$ for $t \in \mathcal{T}$. In short, the fundamental difficulties for adapting existing JL techniques in our setup are \textbf{(1)} when conditioned on $x_t$, the random variables $(\rvz_s)_{s<t}$ \textbf{loss} their independence; \textbf{(2)} there is \textbf{no} characterization on the conditional distribution $P_{(\rvz_s)_{s<t} \mid x_t}$.
\end{remark}

\begin{proof}[Proof of \cref{lem:approx}]
  \textbf{Step 1: Prior approximation.}
  We first show the prior approximation by the fixed prior model, i.e. the event $\mathcal{G}_{1, sa}(\varepsilon/2) = \{ \abs{\norm{\sigma_0 \rvz_{0, sa}}^2 - \sigma_0^2} \le \frac{\varepsilon}{2} \sigma_0^2 \}$ holds for any $(s, a)$.
  This is obvious true even for $\varepsilon=0$ due to the fact $\norm{\rvz_{0, sa}} = 1$ for all $(s, a) \in \mathcal{S} \times \mathcal{A}$ as by the choice of the perturbation distribution $P_{\rvz} := \mathcal{U}(\mathbb{S}^{M-1})$.

  \textbf{Step 2: Posterior approximation}.
  Recall that $\beta = \sigma^2/\sigma_0^2$. To handle the posterior approximation, we first define a sequence of indicator variables
  \[
    x_{\ell, t} = \1{t \in E_{\ell, sa}},
  \]
  where $E_{\ell, sa}$ is the collection of time steps in episode $\ell$ encountering state-action pair $(s, a)$. We also define auxiliary notations
  $\rvz_0 := \rvz_{0, sa}$ and  $x_0 = \sqrt{\beta}$. 
  Immediately, we could rewrite \cref{eq:noise-incremenal} as
  \begin{align}
    \label{eq:proof-approx-1}
    \frac{(N_{k, sa} + \beta)}{\sigma} \tilde{m}_{k, sa}
     & = x_0 \rvz_{0} + \sum_{\ell = 1}^{k-1} \sum_{t \in E_{\ell}} x_{\ell, t} \rvz_{\ell, t+1}
  \end{align}
    The reorganization in \cref{eq:proof-approx-1} is essential to reduce \cref{lem:approx} to the following argument:
    \begin{remark}[Reduction]
    In the following, we are going to prove
        with probability $1- \delta$, the \cref{eq:proof-approx-2} holds for all $(s, a) \in \mathcal{S} \times \mathcal{A}$ and $k \in [K]$ simultaneously:
          \begin{align}
            \label{eq:proof-approx-2}
            (1-\varepsilon) ( {x_0^2 + \sum_{\ell = 1}^{k-1} \sum_{t \in E_{\ell}} x_{t, \ell}^2} )  \le \norm{{x_0 \rvz_0 + \sum_{\ell = 1}^{k-1} \sum_{t \in E_{\ell}} x_{\ell, t} \rvz_{\ell, t+1}}}^2 \le (1+\varepsilon) ( {x_0^2 + \sum_{\ell = 1}^{k-1} \sum_{t \in E_{\ell}} x_{\ell, t}^2} )
          \end{align}
    \end{remark}
  Recall the notations $E_{\ell}$ for the collection of time steps in episode $\ell$ and $E_{\ell, sa}$ for the collection of time steps in episode $\ell$.
  Importantly, the sequential dependence structure in \HyperAgent when interacting with environment is that
  \begin{itemize}
    \item $x_{\ell, t} := \1{t \in E_{\ell, sa}}$ is {\textbf{dependent}} on the environmental and algorithmic randomness in all previous time steps:
          \[
            \rvz_0, (x_{1, t'}, \rvz_{1, t'+1})_{t' \in E_1}, (x_{2, t'}, \rvz_{2, t'+1})_{t' \in E_2}, \ldots, (x_{\ell, t'}, \rvz_{\ell, t'+1})_{t' < t};
          \]
    \item $\rvz_{\ell, t+1}$ is {\textbf{independent}} of the environmental and algorithmic randomness in all previous time steps:
          \[
            \rvz_0, (x_{1, t'}, \rvz_{1, t'+1})_{t' \in E_1}, (x_{2, t'}, \rvz_{2, t'+1})_{t' \in E_2} \ldots, (x_{\ell, t'+1}, \rvz_{\ell, t'+1})_{t' < t}, x_{\ell, t},
          \]
  \end{itemize}
  The difficulty of posterior approximation comes from the above dependence structure as we can not directly use argument conditioning on the entire history $\mathcal{H}_\ell$ at once.
  This is because the conditional distributions of $(\rvz_{\ell', t'})$ given $\mathcal{H}_\ell$ are changed from the unconditional one, without clear characterization. Besides, the random variables $((\rvz_{\ell', t'})$ are not conditionally independent given the history $\mathcal{H}_\ell$.
  \begin{center}
  \emph{
      These difficulties calls for the innovation on fundamental tools in probability with martingale analysis, as shown in \cref{thm:one-dimension-dependent-sequence} and \cref{rem:seq-rand-proj}.
      }
  \end{center}
  Prior approximation guarantees the initial condition $\norm{\rvz_0}^2 = 1$ for applying \cref{thm:one-dimension-dependent-sequence}.
  Observe that for all $(k, s, a)$, we have $x_{\ell, t}:= \1{t \in E_{\ell, sa}} \le 1$ bounded and $\sum_{\ell = 1}^{k-1} \sum_{t \in E_{\ell} } x_{\ell, t}^2 =  \sum_{\ell = 1}^{k-1} \sum_{t \in E_{\ell} } \1{t \in E_{\ell, sa}} = N_{k, sa}$.
  Also, as proved in \citep{li2024simple,li2024probability}, at each time step $(\ell, t)$, the perturbation random vector $\rvz_{\ell, t} \sim P_{\rvz} := \mathcal{U}(\mathbb{S}^{M-1})$ is the $\frac{1}{\sqrt{M}}$-sub-Gaussian random variable in $\R^M$.
  Now, we are ready to apply \cref{thm:one-dimension-dependent-sequence} to the RHS of \cref{eq:proof-approx-1} with sequence $(\rvz_{\ell, t})$ and $(x_{\ell, t})$. This yields the results $\prob ( \cap_{k = 1}^{K+1} \mathcal{G}_{k, sa}(\varepsilon) ) \ge 1 - \delta$
  if
  \begin{align*}
    M \ge \frac{16 (1+ \varepsilon)}{\varepsilon^2}  \left(\log \left( \frac{1}{\delta} \right) + \log \left(1 + \frac{T}{\beta} \right) \right),
  \end{align*}
  where $\sum_{k = 1}^{K} \abs{E_k} = T$ almost surely.
  Then, by union bound over the set $ \mathcal{S} \times \mathcal{A}$, we conclude that
  if
  \begin{align}
    \label{eq:M-selection-general-gamma}
    M \ge \frac{16 (1+ \varepsilon)}{\varepsilon^2}  \left(\log \left( \frac{\abs{\mathcal{S}} \abs{\mathcal{A}} }{\delta} \right) + \log \left(1 + \frac{T}{\beta} \right) \right),
  \end{align}
  then
  \(
  \prob \left( \cap_{(s, a) \in \mathcal{S} \times \mathcal{A}} \cap_{k=1}^{K+1} \mathcal{G}_{k, sa}(\varepsilon) \right) \ge 1 - \delta.
  \)
\end{proof}
\begin{remark}
  According to the proof, the above sequential posterior approximation argument~\cref{lem:approx} holds for any tabular MDP. It does \textit{not} rely on any assumptions made latter for regret analysis.
  If the tabular MDP additionally satisfies \cref{asmp:finite-inhomogeneous-mdp}, the same result in \cref{lem:approx} holds when
  \begin{align}
    \label{eq:M-selection-tabular}
        M \ge M(\varepsilon) := \frac{16 (1+ \varepsilon)}{\varepsilon^2}  \left(\log \left( \frac{ \abs{\mathcal{S}} \abs{\mathcal{A}} }{\delta} \right) + \log \left(1 + \frac{K}{\beta} \right) \right),
  \end{align}
  where $\abs{\mathcal{S}} = SH$.
  The difference between \cref{eq:M-selection-tabular} and \cref{eq:M-selection-general-gamma} is in the $\log (1 + K/ \beta)$ term. This difference is due to the fact we apply \cref{thm:one-dimension-dependent-sequence} for random variables only in a single stage $t$ across episode $\ell = 1, \ldots, K$ under \cref{asmp:finite-inhomogeneous-mdp} since the visitation counts $N_{k, (t, x), a}$ only takes the historical data in stage $t$ into considerations for all $(t, x) \in \mathcal{S}_t$, $a \in \mathcal{A}$ and $t \in \{0, \ldots, H-1\}$.
\end{remark}

\subsection{Insight: How does \HyperAgent drives efficient deep exploration?}
\label{sec:insight}
In this section, we highlight the key components of \HyperAgent that enable efficient deep exploration.
We consider a simple example (adapted from \citep{osband2019deep}) to understand the \HyperAgent's learning rule in \cref{eq:hyperfqi,eq:hyperfqi-sampled} and the role of hypermodel, and how they together drive efficient deep exploration.
\begin{example}
    \label{example:uncertainty-propagation}
    Consider a fixed horizon MDP $\mathcal{M}$ with four states $\mathcal{S}=\{1,2,3,4\}$, two actions $\mathcal{A}=\{up, down\}$ and a horizon of $H=6$. 
    Let us consider the scenario when the agent is at the beginning of $k$-th episode. Let $\mathcal{H}_k$ be the history of all transitions observed prior to episode $k$, and let $\mathcal{H}_{k, sa}=\left(\left(\hat{s}, \hat{a}, r, s^{\prime}\right) \in \mathcal{H}^k:(\hat{s}, \hat{a})=(s, a) \right)$ contain the transitions from state-action pair $(s, a)$ encountered before episode $k$. 
    Suppose $\abs{\mathcal{H}^k_{4, down}}=1$, while for every other pair $(s, a) \neq (4, {down}),\left|\mathcal{D}_{s, a}\right|$ is very large, virtually infinite.
    Hence, we are highly certain about the expected immediate rewards and transition probabilities except for $(4, {down})$.
\end{example}

\begin{figure}[!htbp]
    \centering
    \includegraphics[width=0.5\linewidth]{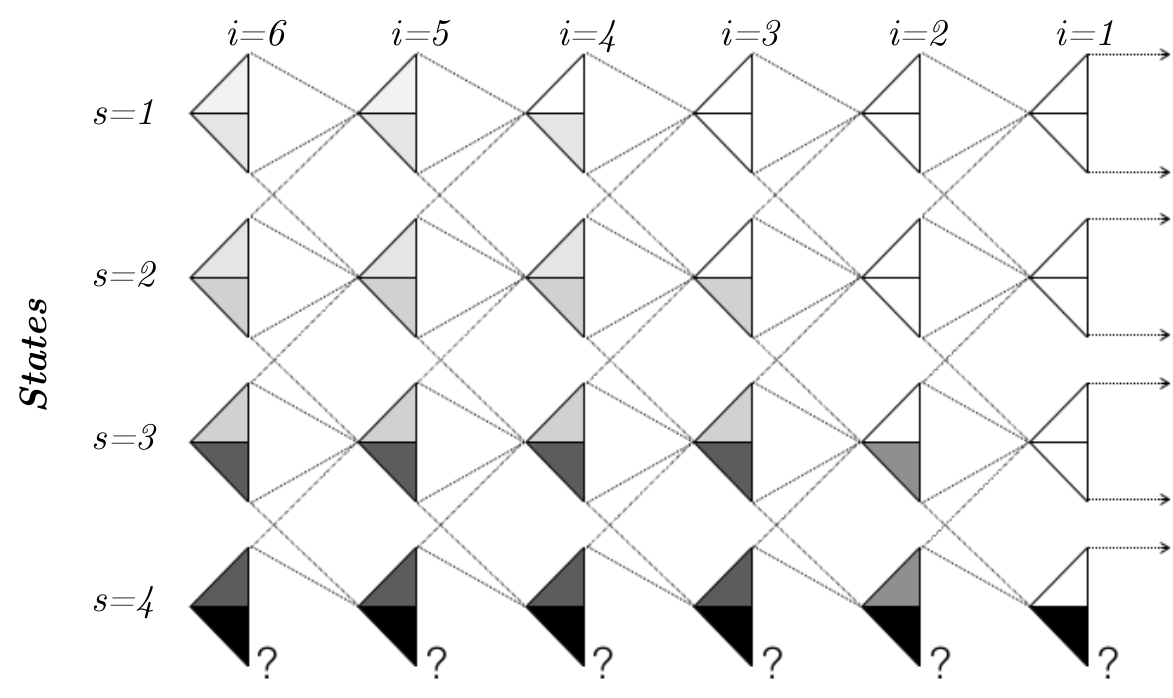}
    \caption{Example to illustrate how \HyperAgent achieves deep exploration. We can see the propagation of uncertainty from later time period to earlier time period in the figure. Darker shade indicates higher degree of uncertainty.}
    \label{fig:uncertainty-propagation}
\end{figure}

From \cref{eq:noise-incremenal,eq:bellman-iteration,eq:bellman-hyperfqi}, \HyperAgent produces a sequence of action-value functions $Q^{(i)}:= f_{\theta_{k}^{(i)}}$ for $i = 0, 1, 2, \ldots, 6$ as shown in \cref{fig:uncertainty-propagation} by iteratively solving \cref{eq:hyperfqi} starting from 
$Q^{(0)} = \mathbf{0}$,
\begin{align*}
    Q^{(i+1)}(s, a)
    & = F^{\gamma}_{k} Q^{(i)} (s, a)
    = \frac{\beta \mu_{0, sa} + N_{k, sa}(r_{sa} + \gamma V_{Q^{(i)}}^\top \hat{P}_{k, sa})}{N_{k, sa} + \beta} + \tilde{m}_{k, sa}^\top \vxi_k(s),
\end{align*}
In \cref{fig:uncertainty-propagation}, each triangle in row $s$ and column $t$ contains two smaller triangles that are associated with action-values of $up$ and $down$ actions at state $s$. The shade on the smaller triangle shows the uncertainty estimates in the $Q^{(i)}(s, a)$, specifically the variance. 
The dotted lines show plausible transitions, except at $(4, down)$. Since we are uncertain about $(4, down)$, any transition is plausible.

As shown in the \cref{fig:uncertainty-propagation}, when $i = 1$, since data size at $(s, a) \neq (4, \text{down})$ tends to infinity, $\norm{\tilde{m}_{k, sa}}^2 \approx 0$ from our \cref{lem:approx}, thus almost zero variance and white colored at $(s, a) \neq (4, \text{down})$;
while $(s,a) = (4, \text{down})$ exhibits large variance, thus dark share, due to the lack of data $\abs{\mathcal{H}_{k, (4, \text{down}) }} = 1$ and \cref{lem:approx}.
By performing bellman update, the noise term ${\tilde{m}_{k, sa}}^\top \vxi_k(s)$ at $(s, a) = (4, \text{down})$ is back-propagated to other states consecutively by the iterative process $i = 2, 3, 4, 5$.
This is because other state-action pairs may lead to the transition into $(4$, down$)$, thus being influenced by the variance introduced by the estimation of $\mathcal{Q}^{(1)}(4$, down$)$. As traced in \cref{fig:uncertainty-propagation}, it's clear that the propagation of noise affects the value estimates in a pattern depicted by the figure's shading. 

This dynamic is crucial for fostering deep exploration. In essence, a sample $\mathcal{Q}^{(6)}(s, a)$ with high variance may appear excessively optimistic in certain episodes, incentivizing the agent to explore that action. Such exploration is justified by the agent's uncertainty about the true optimal value $Q^\star(s, a)$ across the planning horizon. This incentive extends beyond just the immediate reward and transition uncertainty. As depicted in \cref{fig:uncertainty-propagation}, the spread of uncertainty through the system creates incentives for the agent to seek out information, potentially over several time steps, to make a informative observation. This process underlines the core principle of deep exploration.

\subsection{Provable scalability and efficiency of \HyperAgent}
\label{sec:regret}
The design goal of RL algorithm is to maximize the expected total reward up to episode $K$: $\E{\sum_{k=1}^K \sum_{t=1}^{\tau_k} R_{k, t} \mid \cM, \pi_{\operatorname{agent}}}$, which is equivalent to $\E{\sum_{k=1}^K V_{\cM}^{\pi_k}(s_{k, 0}) \mid \cM, \pi_{\operatorname{agent}}}$.
Note that the expectations are over the randomness from stochastic transitions $P(\cdot \mid \cdot)$ under the given MDP $\cM$, and the algorithmic randomization introduced by $\operatorname{agent}$. The expectation in the former one is also over the random termination time $\tau_k$.

Next, we give some specific assumptions under which we simplify the exposition on the analysis.
\subsubsection{Assumptions}
\begin{assumption}[Finite-horizon time-inhomogeneous MDPs]
\label{asmp:finite-inhomogeneous-mdp}
    We consider a problem class that can be formulated as a special case of the general formulation in \cref{sec:pre}. Assume the state space factorizes as $\mathcal{S}=\mathcal{S}_0 \cup \mathcal{S}_1 \cup \mathcal{S}_2 \cup \cdots \cup \mathcal{S}_{H-1}$ where the state always advances from some state $s_t \in \mathcal{S}_t$ to $s_{t+1} \in \mathcal{S}_{t+1}$ and the process terminates with probability 1 in period $H$, i.e.,
    \[
    \sum_{s^{\prime} \in \mathcal{S}_{t+1}} {P}_{s a}\left(s^{\prime}\right)= 1 \quad \forall t \in\{0, \ldots, H-2\}, s \in \mathcal{S}_t, a \in \mathcal{A}
    \]
    and
    \[
    \sum_{s^{\prime} \in \mathcal{S}} {P}_{s a}\left(s^{\prime}\right)=0 \quad \forall s \in \mathcal{S}_{H-1}, a \in \mathcal{A} .
    \]
    For notational convenience, we assume each set $\mathcal{S}_0, \ldots, \mathcal{S}_{H-1}$ contains an equal number of  $S$ elements. 
    Each state $s \in \mathcal{S}_t$ can be written as a pair $s = (t, x)$ where $t \in \{0, \ldots, H-1\}$ and $ x \in \mathcal{X} =\{1, \ldots, S\} $ . That is, $\abs{\mathcal{S}} = SH$.
\end{assumption}
Our notation can be specialized to this time-inhomogenous problem, writing transition probabilities as $P_{t, xa}\left(x^{\prime}\right) \equiv {P}_{(t, x), a}\left(\left(t+1, x^{\prime}\right)\right)$.
For consistency, we also use different notation for the optimal value function, writing
\[
V_{\cM, t}^\pi(x) \equiv V_{{\cM}}^\pi((t, x))
\]
and define $V_{{\cM}, t}^*(x):=\max _\pi V_{{\cM}, t}^\pi(x)$. Similarly, we can define the state-action value function under the MDP at timestep $t \in\{0, \ldots, H-1\}$ by
\[
Q_{{\cM}, t}^*(x, a)=\mathbb{E}\left[r_{t+1}+V_{{\cM}, t+1}^*\left(x_{t+1}\right) \mid \mathcal{\cM}, x_t=x, a_t=a\right] \quad \forall x \in \mathcal{X}, a \in \mathcal{A} .
\]
This is the expected reward accrued by taking action $a$ in state $x$ and proceeding optimally thereafter.
Equivalently, the process applying true bellman operator $F_{\cM,t} Q_{\cM,t+1}^*(\cdot, \cdot) := F_{\cM}^\gamma Q_{{\cM}}^*((t+1, \cdot), \cdot)$ when $\gamma=1$, where $F_{\cM}^{\gamma}$ is defined in \cref{eq:bellman-true}
yields a series of optimal state-action value functions, fulfilling \( Q_{\cM, H}^* = 0 \) and the Bellman equation \( Q_{\cM, t}^* = F_{\cM, t} Q_{\cM, t+1}^* \) for \( t < H \).

\begin{table}[!htbp]
    \centering
    \begin{tabular}{l|r}
        \toprule[1.2pt]
        Hyper-parameters                       & Finite MDP with Horizon $H$ \\
        \midrule %
        reference distribution $P_{\xi}$          &   $N(0, I_M)$   \\
        perturbation distribution $P_{\rvz}$         &    $\mathcal{U}(\mathbb{S}^{M-1})$    \\
        level of perturbation $\sigma^2$  &  $6H^2$
        \\
        prior variance $\sigma_0^2$ &  $6H^2/\beta$
        \\
        prior regularization $\beta$ & $\beta$ in \cref{assump:dirichlet-prior}
        \\
        {index dimension $M$} & $M(1/2)$ in \cref{eq:M-selection-tabular}
        \\ 
        discount factor $\gamma$               & 1                           \\
        $\operatorname{target\_update\_freq}$  & 1                           \\
        $\operatorname{sample\_update\_ratio}$ & 1                           \\
        $\operatorname{training\_freq}$        & H                           \\
        \bottomrule[1.2pt]
    \end{tabular}
    \caption{Hyper-parameters of our Tabular-\HyperAgent and corresponding \update rules.}
    \label{tab:hyper_parameter_tabular}
\end{table}
The agent designer's prior beliefs over MDPs $M$ is formalized with mild assumptions.
\begin{assumption}[Independent Dirichlet prior for outcomes]
    \label{assump:dirichlet-prior}
    For each $(s, a) \in \mathcal{S} \times \mathcal{A}$, the outcome distribution is drawn from a Dirichlet prior
    \[
        P_{sa} \sim \operatorname{Dirichlet}(\alpha_{0, sa})
    \]
    for $\alpha_{0, sa} \in \mathbb{R}^{\mathcal{S}}_{+}$ and each $P_{sa}$ is drawn independently across $(s, a)$. Assume there is $\beta \ge 3$ such that $\vone^\top \alpha_{0, sa} = \beta$ for all $(s, a)$.
\end{assumption}

\subsubsection{Bayesian analysis}

Denote the short notation $[n] = \{1, \ldots, n \}$.
Let us define a filtration $(\mathcal{Z}_k)_{k \ge 1}$ on the algorithmic random perturbation that facilitates the analysis
\[
    \mathcal{Z}_{k} = \sigma( (\rvz_{0, sa})_{(s, a)\in \mathcal{S} \times \mathcal{A}}, (\rvz_{\ell, t}: \ell \in [k-1], t \in E_{\ell}) ). 
\]
Specifically from this definition, $\tilde{m}_{k, sa}$ is $(\mathcal{H}_k, \mathcal{Z}_k)$-measurable.
As derived in \cref{eq:bellman-hyperfqi}, the perturbation $\tilde{m}_{k, s, a}^\top \vxi_{k}(s)$ injected to the Bellman update is conditionally Gaussian distributed
\begin{align}
\label{eq:noise-gaussian-distributed}
    \tilde{m}_{k, s, a}^\top \vxi_{k}(s) \mid \mathcal{H}_k, \mathcal{Z}_k \sim N(0, \norm{\tilde{m}_{k, s, a}}^2),
\end{align}
due to the fact that for all $s \in \mathcal{S}$, $\vxi_k(s)$ is independent of $\mathcal{H}_k, \mathcal{Z}_k$ and is Normal distributed.

Meanwhile, using notation $s = (t, z)$ in the time-inhomogeneous setting where $\abs{E_\ell} = H$ for all $\ell$,
these injected perturbation $\tilde{m}_{k, (t, x), a}^\top \vxi_{k}((t, x))$ are conditionally independent across $(t, x) \in ([H-1] \cup \{0\}) \times \mathcal{X}$ given $\mathcal{H}_t, \mathcal{Z}_t$.
From \cref{lem:approx} using $M(1/2)$ in \cref{eq:M-selection-tabular}, the noise level of $\tilde{m}_{k, (t, x), a}^\top \vxi_{k}((t, x))$ is a $(1/2)$-approximation of the noise level of $\omega_{k}(t, x, a)$ through out the entire learning process for all tuples $(k, (t, x), a)$.
Therefore, the role of the injected perturbation $\tilde{m}_{k, (t, x), a}^\top \vxi_{k}((t, x))$ in the Bellman update of \HyperAgent is essentially the same as the Gaussian noise $\omega_{k}(t, x, a)$ injected in the Bellman udpate of RLSVI. 

Basically, as long as the sequential approximation argument in \cref{lem:approx} is established, the regret analysis follows analysis of RLSVI in the Section 6 of \citep{osband2019deep}.
Still, we want to emphasize a key argument that enables efficient deep exploration is the stochastic optimism of \HyperAgent by a selection on $\sigma^2 = 6H^2$ which is a double of the number of $\sigma^2$ selected in \citep{osband2019deep}. The rest of the analysis follows Section 6.4 (Optimism and regret decompositions) and Section 6.6 (Analysis of on-policy Bellman error) in \citep{osband2019deep}.

\begin{definition}[Stochastic optimism]
    A random variable $X$ is stochastically optimistic with respect to another random variable $Y$, written $X \geq_{S O} Y$, if for all convex increasing functions $u: \mathbb{R} \rightarrow \mathbb{R}$
    \[
        \mathbb{E}[u(X)] \geq \mathbb{E}[u(Y)].
    \]
\end{definition}
We show that \HyperAgent is stochastic optimistic in the sense that it overestimates the value of each state-action pairs in expectation. This is formalized in the following proposition.
\begin{proposition}
\label{prop:so-value}
    If \cref{asmp:finite-inhomogeneous-mdp,assump:dirichlet-prior} hold and tabular \HyperAgent is applied with planning horizon $H$ and parameters parameters $(M, \mu_0, \sigma, \sigma_0)$ satisfying $M = M(1/2)$ defined in \cref{eq:M-selection-tabular}, $(\sigma^2/\sigma_0^2) = \beta$, $\sigma \ge \sqrt{6} H $ and $\min_{s, a}\mu_{0, s, a} \ge H$,
    \begin{align}
        \label{eq:optimism}
        f_{\theta_k}(s, a, \vxi_k(s)) \mid (\mathcal{H}_{k}, \mathcal{Z}_k) \succeq_{S O} Q_{{\cM}}^*(s, a) \mid (\mathcal{H}_{k}, \mathcal{Z}_k), \quad \forall (s, a) \in \mathcal{S} \times \mathcal{A}
    \end{align}
    holds for all episode $k \in \{0, \ldots, K\}$ simultaneously with probability at least $(1 - \delta)$.
\end{proposition}
The following \cref{lem:so-operator} about the stochastic optimistic operators is also built upon our sequential posterior approximation argument in \cref{lem:approx}. 
As long as \cref{lem:so-operator} is established, the proof of \cref{prop:so-value} follows the proof of Corollary 2 in \citep{osband2019deep}.
\begin{lemma}[$\delta$-approximate stochastically optimistic operators]
    \label{lem:so-operator}
    If \cref{assump:dirichlet-prior} holds and \HyperAgent is applied with $(M, \mu_0, \sigma, \sigma_0)$ satisfying
    $M = M(1/2)$ defined in \cref{eq:M-selection-general-gamma} or \cref{eq:M-selection-tabular}, $\sigma^2/\sigma_{0}^2 = \beta$, $\sigma \ge \sqrt{6} \gamma \operatorname{Span}(V_{Q})$ and $\min_{s, a} \mu_0(s, a) \ge \gamma \max_{s\in \mathcal{S}} V_Q(s) + 1$,
    \begin{align}
        F_{k}^{\gamma} Q (s, a) \mid (\mathcal{H}_{k}, \mathcal{Z}_k) \succeq_{SO} F_{\cM}^{\gamma} Q (s, a) \mid (\mathcal{H}_k, \mathcal{Z}_k), \quad \forall (s, a) \in \mathcal{S} \times \mathcal{A}.
    \end{align}
    holds for all episode $k \in \{0, \cdots, K\}$ simultaneously with probability at least $(1 - \delta)$.
\end{lemma}

\begin{remark}
    The \cref{lem:so-operator} with $M(1/2)$ in \cref{eq:M-selection-general-gamma} does not require time-inhomogeneity \cref{asmp:finite-inhomogeneous-mdp} and holds with any fixed $\gamma \ge 0$.
    In the time-inhomogeneous problem, one can set $M(1/2)$ in \cref{eq:M-selection-tabular}, $\gamma = 1$ and $\operatorname{Span}(V_Q) \le \max_{s} V_{Q} \le H-1$ as $V_Q$ is from the case that $\{ t = 1, \ldots, H-1 \}$.
\end{remark}
\begin{proof}[Proof of \cref{lem:so-operator}]
Recall from \cref{eq:bellman-true}, we have
    \begin{align*}
        F^{\gamma}_{\cM} Q (s, a) = r_{sa} + \gamma V_{Q}^\top P_{sa} = (r_{sa} \vone + \gamma V_Q)^\top P_{sa},
    \end{align*}
where $\vone$ is the all one vector in $\R^{\abs{\mathcal{S}}}$.
Notice that by \cref{assump:dirichlet-prior}, 
for each pair $(s, a)$, the transition vector $P_{sa} \independent \mathcal{Z}_k \mid \mathcal{H}_k$ and the conditional distribution is
\[
    \prob( P_{sa} \in \cdot \mid (\mathcal{H}_{k}, \mathcal{Z}_k) ) = \prob( P_{sa} \in \cdot \mid \mathcal{H}_{k} ) = \operatorname{Dirichlet}(\alpha_{k, sa}),
\]
where $\alpha_{k, sa} = \alpha_{0, sa} + N_{k, sa} \hat{P}_{k, sa} \in \mathbb{R}^{S}$.
Define the posterior mean of the transition vector as
\begin{align}
    \bar{P}_{k,sa} & := \E{ P_{sa} \mid \mathcal{H}_k } 
    = \frac{ \alpha_{0, sa} + N_{k, sa} \hat{P}_{k,sa}}{ \beta + N_{k, sa} } \in \R^{\abs{\mathcal{S}}},
\end{align}
which can be interpreted as the empirical transition $\hat{P}_{s, a}$ but with a slight regularization towards the prior mean $\mu_{0,sa}$ by weights $\beta = \vone^\top \alpha_{0, sa}$.

As mentioned in \cref{sec:analysis,eq:noise-gaussian-distributed}, from \cref{eq:bellman-hyperfqi,}
\begin{align*}
    F^{\gamma}_{k} Q (s, a) \mid (\mathcal{H}_{k}, \mathcal{Z}_k) \sim N(\mu_{k,sa}, \norm{\tilde{m}_{k, sa}}^2)
\end{align*}
where the conditional mean $\mu_{k, sa}$ is
\begin{align}
\label{eq:bellman-hyperfqi-mean}
    \mu_{k, sa}
     & = (r_{sa} \vone + \gamma V_Q)^\top \bar{P}_{k, sa} + \frac{\beta \mu_{0, sa} - \beta r_{sa} - \gamma V_Q^\top \alpha_{0, sa}}{\beta + N_{k, sa}}.
\end{align}
From our established \cref{lem:approx}, when $M = M(\varepsilon)$, the joint event $\cap_{(k, s, a)\in [K]\times\mathcal{S}\times \mathcal{A}}G_{k, sa}(\varepsilon)$ holds with probability $1 - \delta$.
Specifically, $G_{k, sa}(1/2) \in \sigma(\mathcal{H}_k, \mathcal{Z}_k)$ implies that
\[
    \norm{\tilde{m}_{k, sa}}^2 \ge (1/2)\sigma^2/(\beta + N_{k, sa}).
\]
Under the above established event, the result follows from \cref{lem:gaussian-dirichlet-optimism} if we establish on conditional variance $\norm{\tilde{m}_{k, sa}}^2 \ge 6 (\vone^\top \alpha_{k, sa})^{-1} \operatorname{Span} ( r_{sa}\vone + \gamma V_Q )^2 $ and on conditional mean $\mu_{k, sa} \ge ( \vone^\top \alpha_{k, sa} )^{-1}(r_{sa}\vone + \gamma V_Q)^\top \alpha_{k, sa} = (r_{sa}\vone + \gamma V_Q)^\top \bar{P}_{k, sa}$ since $\vone^\top \alpha_{0, sa} = \beta$ for all $(s, a)$.
That is, for conditional variance
\begin{align*}
    \frac{3\cdot \text{Span}(r_{sa} \vone + \gamma V_Q)^2 }{(\vone^\top \alpha_{k, sa})} \le \frac{3\gamma^2\cdot \text{Span}(V_Q)^2}{\beta + N_{k, sa}} \stackrel{(1)}{\le} \frac{(1/2)\sigma^2}{\beta + N_{k, sa}} \stackrel{(2)}{\le} \norm{\tilde{m}_{k, sa}}^2,
\end{align*}
where $(1)$ is from the condition of $\sigma$ in \cref{lem:so-operator} and $(2)$ holds true under the event $G_{k, sa}(1/2)$.
Next we have the conditional mean $\mu_{k, sa}$ dominating the posterior mean of true bellman update $\E{F_{\cM}^{\gamma}Q \mid \mathcal{H}_k} = (r_{sa}\vone + \gamma V_Q)^\top \bar{P}_{k, sa}$,
\begin{align*}
    \mu_{k, sa} - (r_{sa}\vone + \gamma V_Q)^\top \bar{P}_{k, sa}
     & = \frac{\beta \mu_{0, sa} - \beta r_{sa} - \gamma V_Q^\top \alpha_{0, sa}}{\beta + N_{k, sa}} \\
     & \ge \frac{\beta \mu_{0, sa} - \beta (\gamma \max_{s \in \mathcal{S}} V_Q(s) + 1)}{\beta + N_{k, sa}}\\
     & \ge 0
\end{align*}
because of the condition in \cref{lem:so-operator} that $\min_{sa}\mu_{0, sa} \ge \gamma \max_{s} V_Q(s) + 1$.
\end{proof}

\begin{lemma}[Gaussian vs Dirichlet optimism, Lemma 4 in \cite{osband2019deep}]
    \label{lem:gaussian-dirichlet-optimism}
    Let $Y=p^\top v$ for $v \in \mathbb{R}^n$ fixed and $p \sim \operatorname{Dirichlet}(\alpha)$ with $\alpha \in \mathbb{R}_{+}^n$ and $\sum_{i=1}^n \alpha_i \geq 3$. Let $X \sim N\left(\mu, \sigma^2\right)$ with $\mu \geq (\sum_{i=1}^n \alpha_i)^{-1}{\sum_{i=1}^n \alpha_i V_i}, \sigma^2 \geq 3\left(\sum_{i=1}^n \alpha_i\right)^{-1} \operatorname{Span}(V)^2$, then $X \succeq_{S O} Y$. Thus, $p^\top v$ is sub-Gaussian with parameter $\sigma^2$.
\end{lemma}

\clearpage
\section{Understanding \HyperAgent via comprehensive studies on DeepSea}
\label{appendix:additional_res_deepsea}
In this section, we present insightful experiments to demonstrate the impact of critical components on \HyperAgent. Specifically, we employ DeepSea to validate the theoretical insights of our method, discuss the sampling schemes for $Q$-target computation and action selection, introduce methods for achieving accurate posterior approximation, and compare with additional baselines within the hypermodel framework as discussed in \cref{appendix:difference}.

\subsection{Validating theoretical insights through experimentation}

\paragraph{Empirical validation of theoretical insights on the index dimension $M$.}
According to the theoretical analysis in \cref{lem:approx}, increasing the index dimension $M$ can enhance the approximation of posterior covariance under closed-form solution of \cref{eq:hyperfqi}, where its expectation is precisely computed, thereby facilitating deep exploration.

\begin{figure}[htbp]
    \centering
    \includegraphics[width=0.6\linewidth]{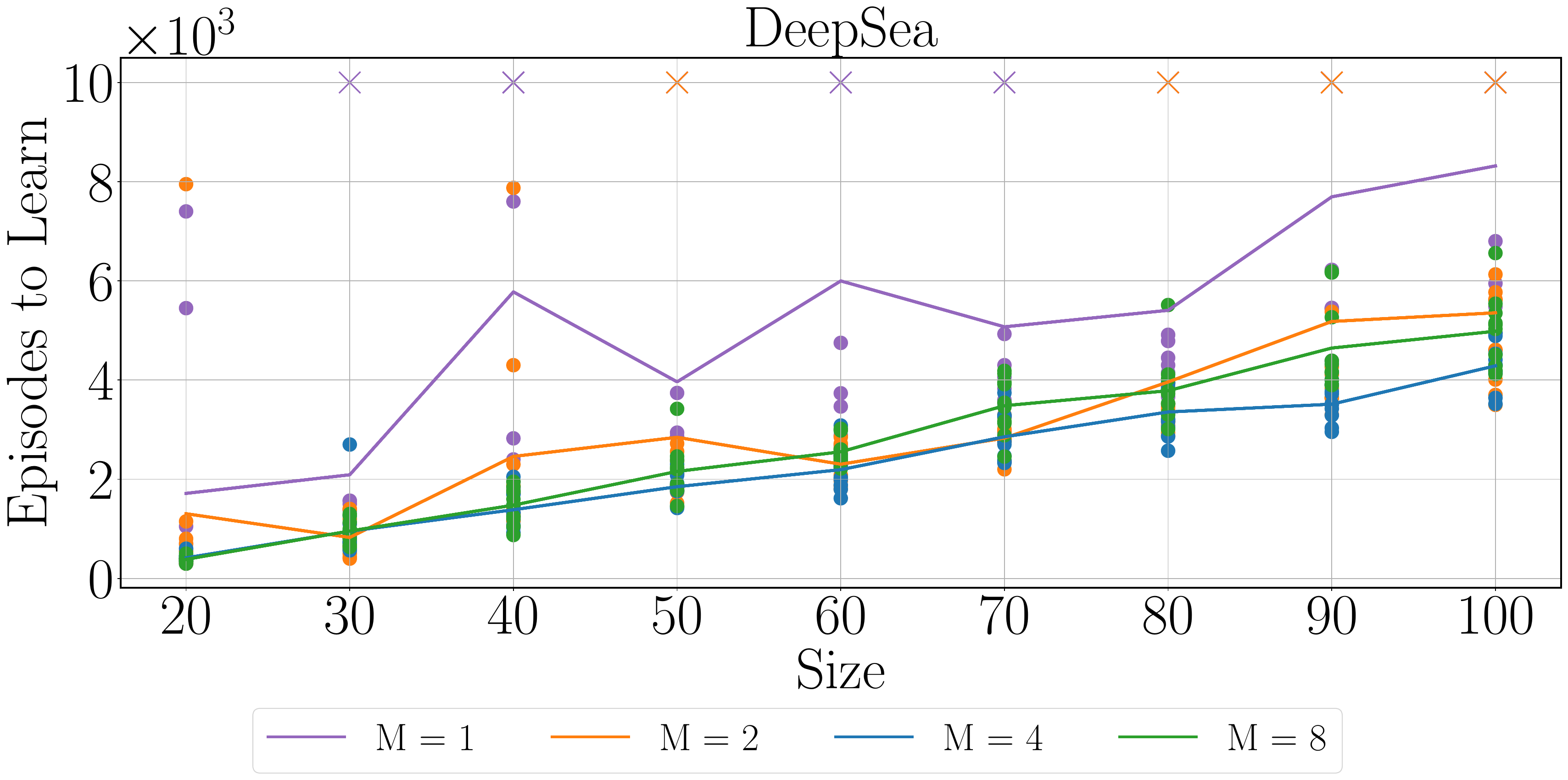}
    \caption{Ablation results on the impact of index dimension $M$. \HyperAgent exhibits improved performance when $M$ exceeds 1, aligning with our theoretical analysis.}
    \label{fig:deepsea_Ms}
\end{figure}

In this experiment, we set the number of indices $\NpS$ of \cref{eq:hyperfqi-sampled} to 20 for all index dimensions $\{1, 2, 4, 8\}$.
Let us first discuss the case $M = 1$, it can be regarded as an incremental updated RLSVI point estimate without resampling the perturbation and re-solving the least-square for the entire history.
As evidenced by the results in Figure~\ref{fig:deepsea_Ms}, the incremental version of RLSVI cannot fully solve DeepSea tasks starting from size of 30, indicating dimension $M > 1$ is necessary and increasing a bit on $M$ benefits deep exploration, possibly from more accurate posterior approximation by larger $M$.

\begin{figure}[htbp]
    \centering
    \includegraphics[width=0.6\linewidth]{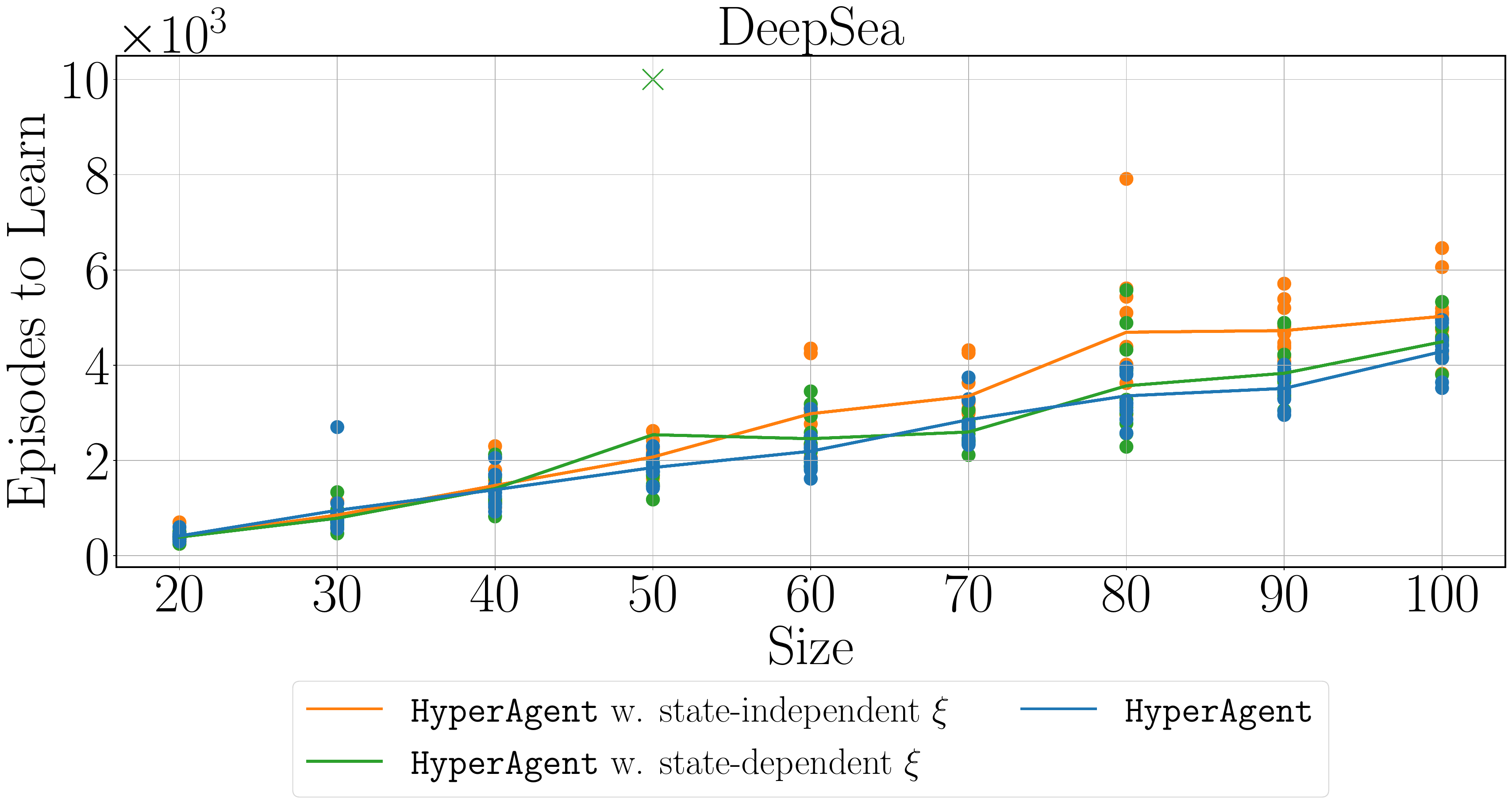}
    \caption{Ablation results for the sampling schemes used in $Q$-target computation. Both sampling schemes achieve similar performance.}
    \label{fig:deepsea_xi}
\end{figure}

\paragraph{Ablation experiment on different sampling schemes for $\vxi$.}
\label{sec:ablation_sampling_schemes}
We have two sampling schemes for $\vxi$ in the computation of the $Q$-target and action selection: state-independent $\vxi$ sampling and state-dependent $\vxi$ sampling, see detailed description in Appendix~\ref{sec:index-sampling-schemes}.
In our implementation by default, \HyperAgent employs the state-independent $\vxi$ for action selection, where we use
the same index $\vxi_k(S) = \xi_k$ for each state within episode $k$. Whereas, \HyperAgent utilizes state-dependent $\vxi$ for $Q$-target computation, where we sample independent $\xi \sim P_{\xi}$ when computing the $Q$-target for each transition tuple in mini-batches. 
To systematically compare the index sampling schemes on action selection and $Q$-target computation, two variants are created. 
\begin{itemize}
    \item \HyperAgent \textit{with state-independent $\vxi$}: applying state-independent $\vxi$ sampling to both $Q$-target computation and action selection.
    \item \HyperAgent \textit{ with state-dependent $\vxi$}: applying state-dependent $\vxi$ sampling to both $Q$-target computation and action selection.
\end{itemize}   
As illustrated in Figure~\ref{fig:deepsea_xi}, these distinct index sampling schemes for $\vxi$ exhibit nearly identical performance.

\begin{figure}[htbp]
    \centering
    \subfigure[We set $M=4$ for all algorithms. \HyperAgent w. OIS achieves the best performance.]{
    \includegraphics[width=0.48\linewidth]{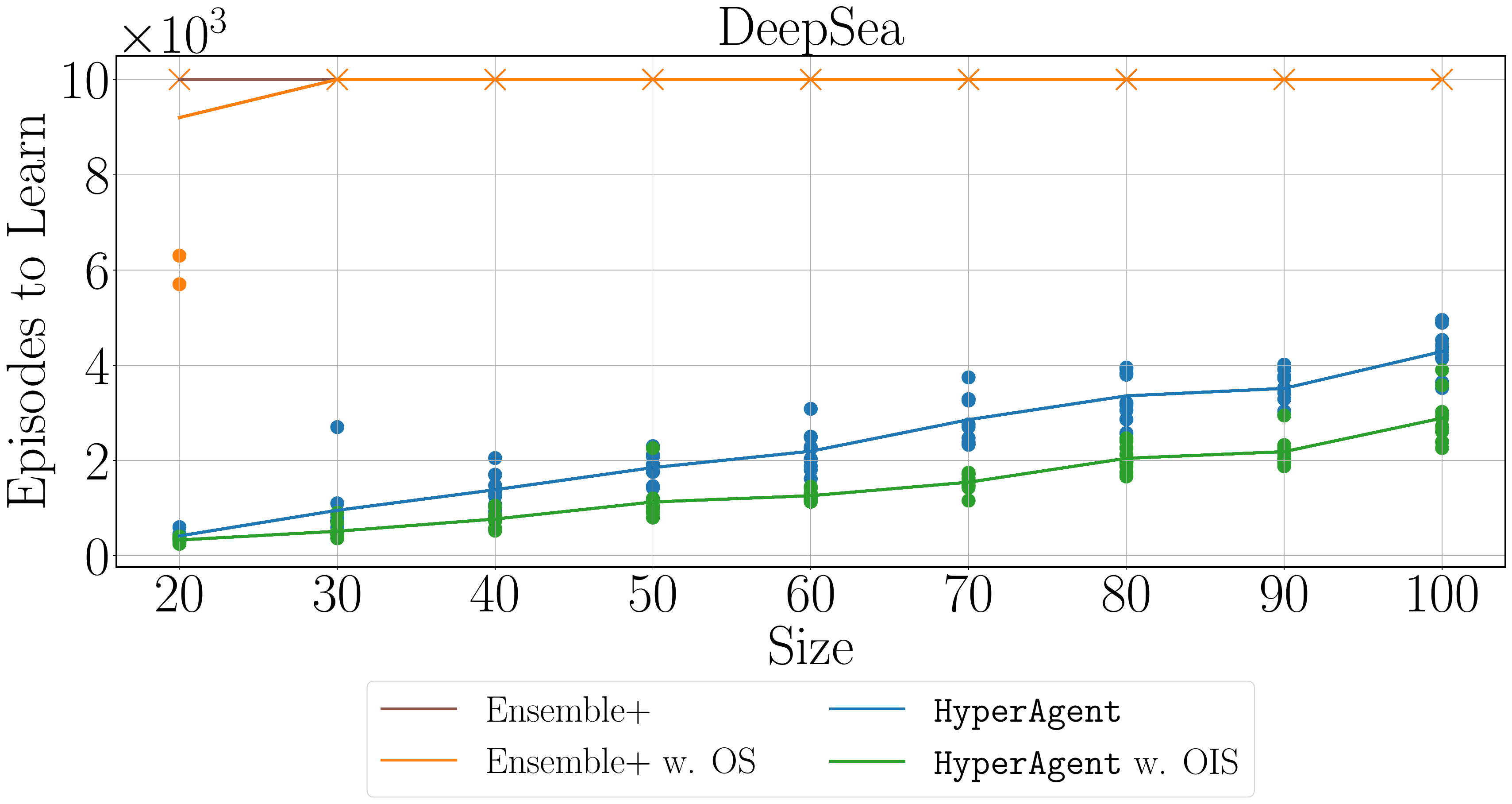}}
    \subfigure[We set $M=16$ for algorithms with ensemble network, and keep $M=4$ for \HyperAgent.]{
    \includegraphics[width=0.48\linewidth]{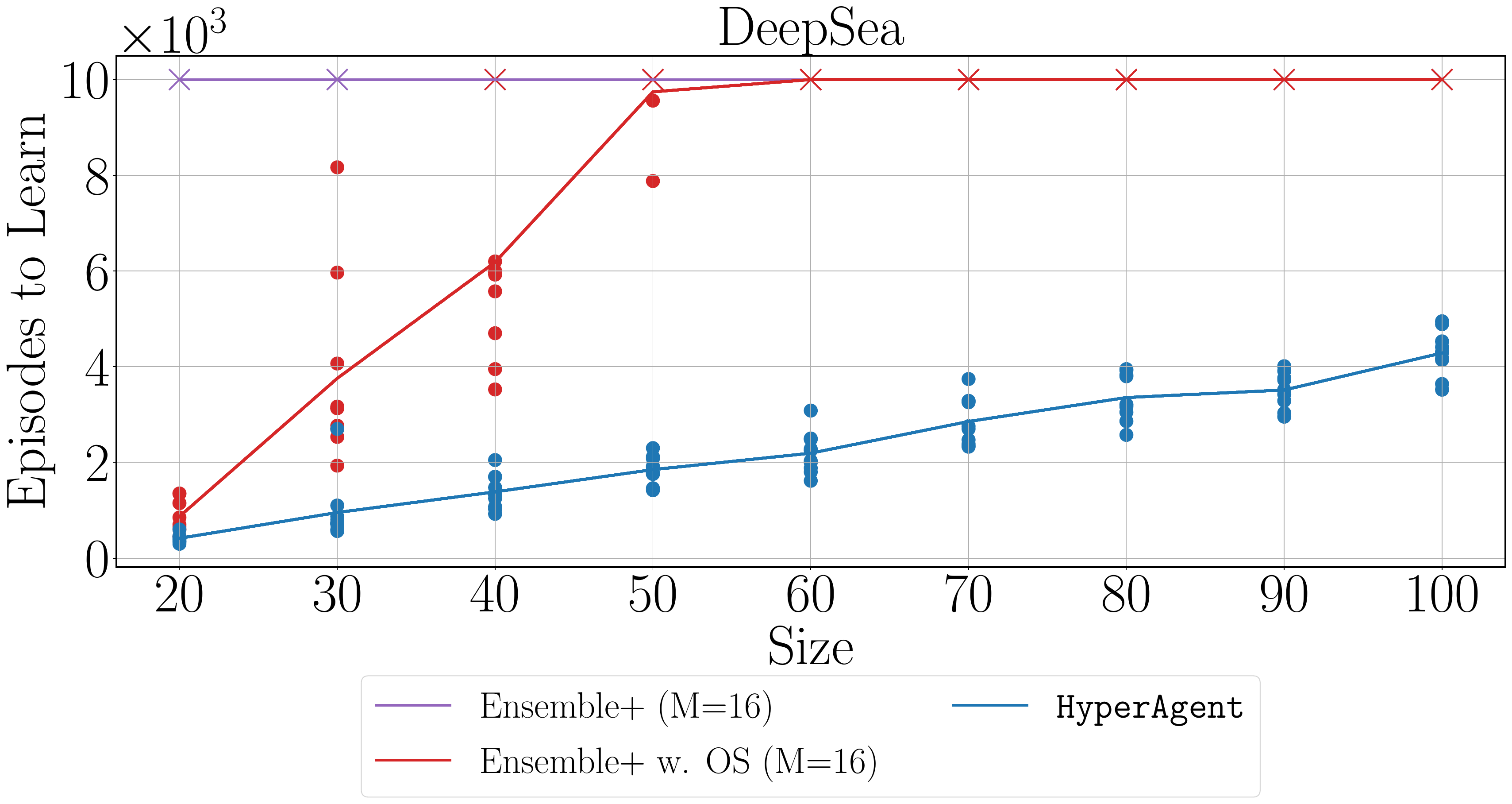}}
    \caption{Results on different action selection schemes. The OIS method can achieve better performance due to the optimistic estimates.}
    \label{fig:deepsea_ois}
\end{figure}

\paragraph{Comparative results on optimistic index sampling.}
Introduced in \ref{appendix:optimistic_index_sampling}, optimistic index sampling (OIS) for action selection and $Q$-target computation is another index sampling scheme in \HyperAgent framework, naming \HyperAgent w. OIS.
As shown in Figure~\ref{fig:deepsea_ois}(a), \HyperAgent w. OIS outperforms \HyperAgent by leveraging the OIS method to generate more optimistic estimates, thereby enhancing exploration. The OIS method incurs minimal additional computation, as we have set the dimension $M$ to 4.

The empirical implementation of LSVI-PHE~\citep{ishfaq2021randomized} also utilizes the similar optimistic sampling (OS) method with ensemble networks as described in~\cref{sec:related}.
Since no official implementation of LSVI-PHE is available, we apply the OIS to the Ensemble+~\citep{osband2018randomized,osband2019deep}, denoted as Ensemble+ w. OS, representing the LSVI-PHE.
A critical difference between LSVI-PHE and \HyperAgent w. OIS is that LSVI-PHE performs maximum value over $M$ separately trained ensemble models, while \HyperAgent w. OIS has the computation advantage that we can sample as many as indices from reference distribution $P_{\xi}$ to form the optimistic estimation of value function even when the index dimension $M$ is small.

When $M = 4$, both Ensemble+ and Ensemble+ w. OS demonstrates subpar performance.
According to~\citet{osband2018randomized} and the observation in \cref{fig:deepsea_ois}, a larger ensemble size ($M =16$) can lead to improved performance.
When $M = 16$, both Ensemble+ w. OS (LSVI-PHE) has a bit better performance compared with Ensemble+.
The result of Ensemble+ w. OS ($M = 16$) and \HyperAgent w. OIS demonstrates enhanced deep exploration via optimistic value estimation, which is aligned with our theoretical insights.

In another dimension of comparison, As depicted in Figure~\ref{fig:deepsea_ois}(b), Ensemble+ w. OS ($M=16$) still fails to solve the large-size DeepSea.
This further showcases the exploration efficiency of \HyperAgent.

\subsection{Ablation studies on implementation settings and hyper-parameters}
\label{sec:abltaion_implementation}

\begin{figure}[htbp]
    \centering
    \begin{minipage}[h]{0.48\textwidth}
        \centering
        \includegraphics[width=0.95\linewidth]{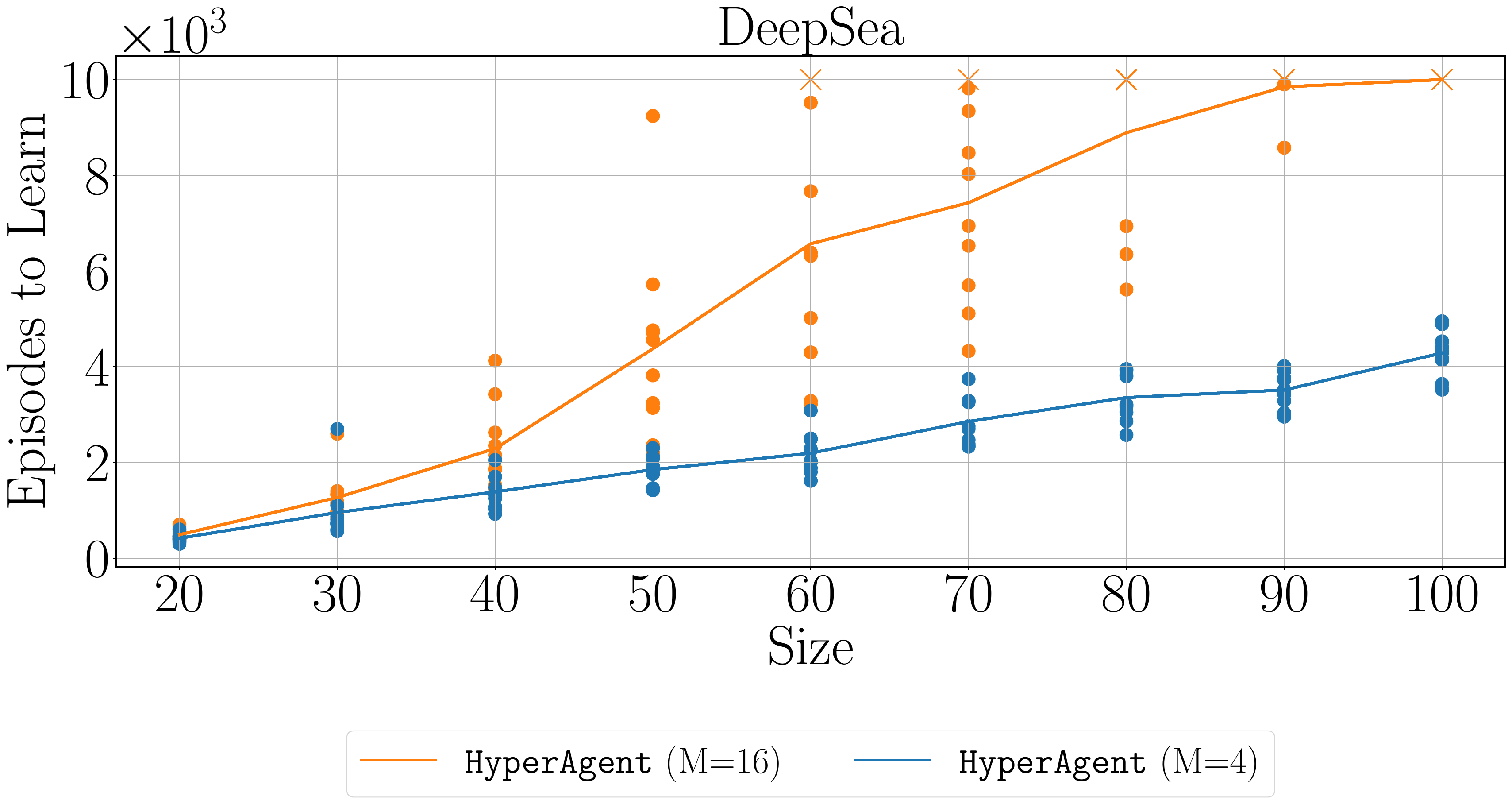}
        \caption{Results of \HyperAgent with large index dimension. In both settings, we set $\NpS = 20$, which fails to achieve satisfactory performance for $M=16$.}
        \label{fig:deepsea_m16} 
    \end{minipage}
    \hspace{1.0pt}
    \begin{minipage}[h]{0.48\textwidth}
        \centering
        \includegraphics[width=0.95\linewidth]{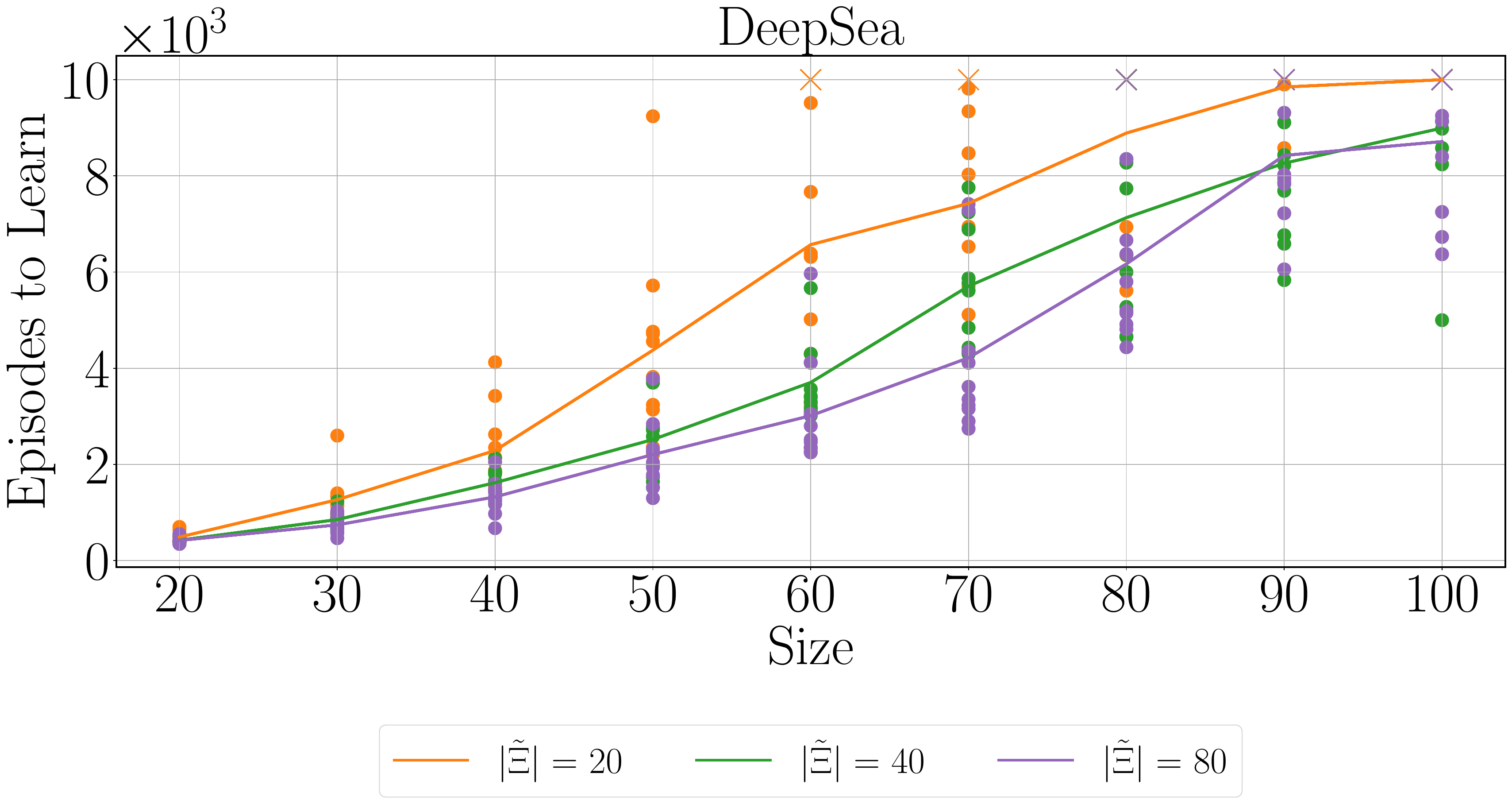}
        \caption{Results of \HyperAgent with index dimension $M=16$. We can increase the number of indices $\NpS$ to alleviate performance degradation,}
        \label{fig:deepsea_nps}
    \end{minipage}
\end{figure}

\paragraph{Sampling-based approximation.}
\HyperAgent by default uses Gaussian reference distribution $P_{\xi} =  N(0, I_M)$, which requires sampling-based approximation~\cref{eq:hyperfqi,eq:hyperfqi-sampled}.
Therefore, it is imperative to set the number of indices $\NpS$ of \cref{eq:hyperfqi-sampled} to be sufficiently large for a good approximation given index dimension $M$.
$\NpS=20$ falls short of delivering satisfactory results when $M = 16$ as illustrated in Figure~\ref{fig:deepsea_m16}. For this scenario, increasing the number of indices $\NpS$ can alleviate performance degradation as depicted in Figure~\ref{fig:deepsea_nps}, facilitating deep exploration empirically. 
To achieve accurate approximation, $\NpS$ may need to grow exponentially with $M$, but this comes at the cost of increased computation. To balance accuracy and computation, we have chosen $M = 4$ and $\NpS=20$ as the default hyper-parameters, which already demonstrate superior performance in Figure~\ref{fig:deepsea_baseline}. 

\begin{figure}[htbp]
    \centering
    \includegraphics[width=0.6\linewidth]{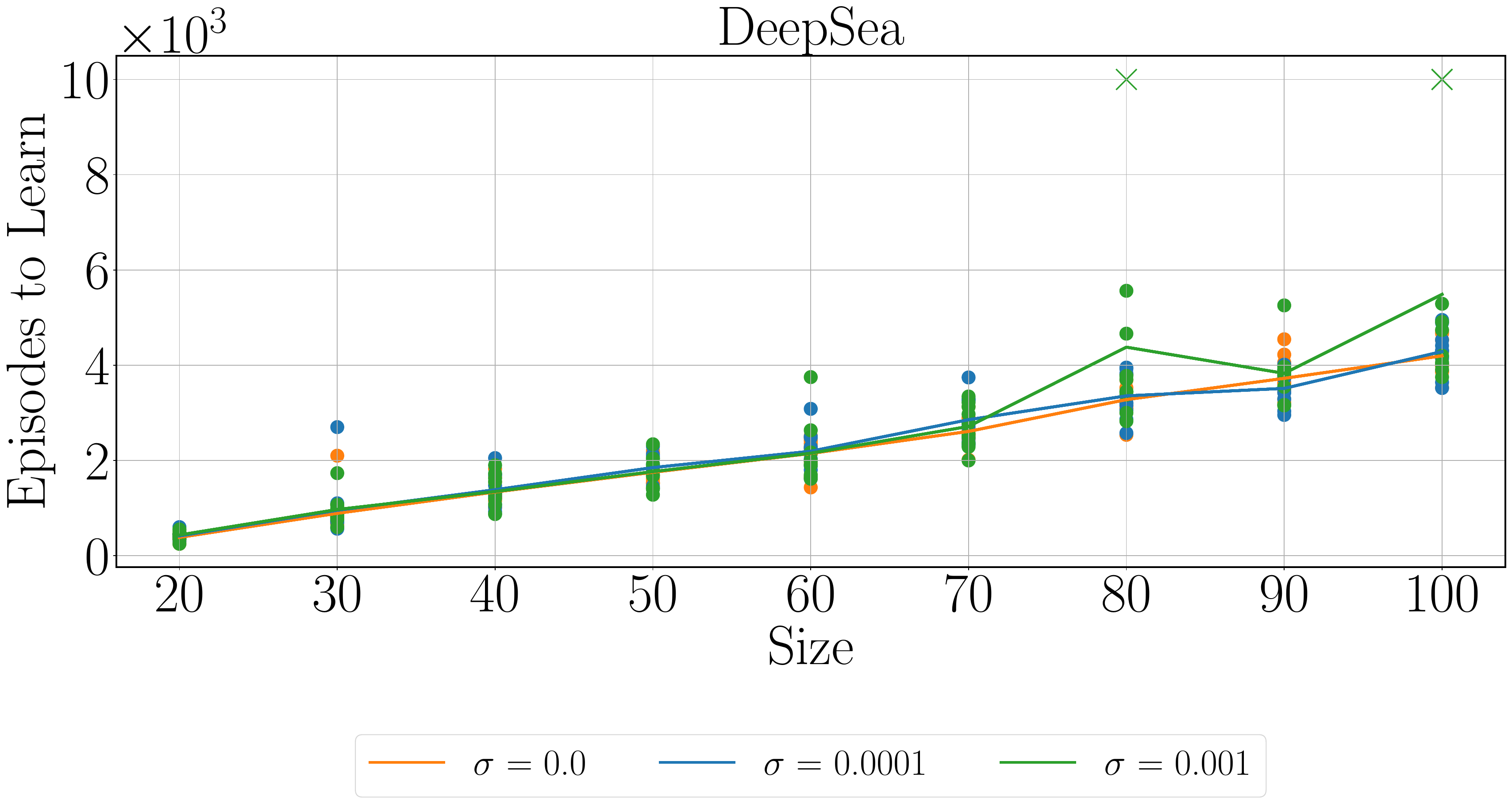}
    \caption{Ablation results on the $\sigma=0.0001$  of~\cref{eq:q-loss}. We set $\sigma=0.0001$ as the default setting.}
    \label{fig:deepsea_sigma}
\end{figure}

\paragraph{Ablation results for other hyper-parameters.}
In addition, we have also investigated the effect of the $\sigma$ of~\cref{eq:q-loss} on our method, as shown in Figure~\ref{fig:deepsea_sigma}. \HyperAgent is not sensitive to this hyper-parameter, and we have selected $\sigma=0.0001$ as the default hyper-parameter.

\subsection{Additional results with algorithms using hypermodel framework}
\label{sec:hypermodel_framework_diff}
\paragraph{Experiment investigating different network structures within the hypermodel framework.}
We perform an ablation experiment examining various network structures, (1) HyperModel~\citep{dwaracherla2020hypermodels}
(2) epinet~\citep{osband2023epistemic}, outlined in~\cref{appendix:difference} with the same hyper-parameters, same algorithmic update rule, same target computation rule and same action selection rule.

\begin{figure}[htbp]
    \centering
    \includegraphics[width=0.6\linewidth]{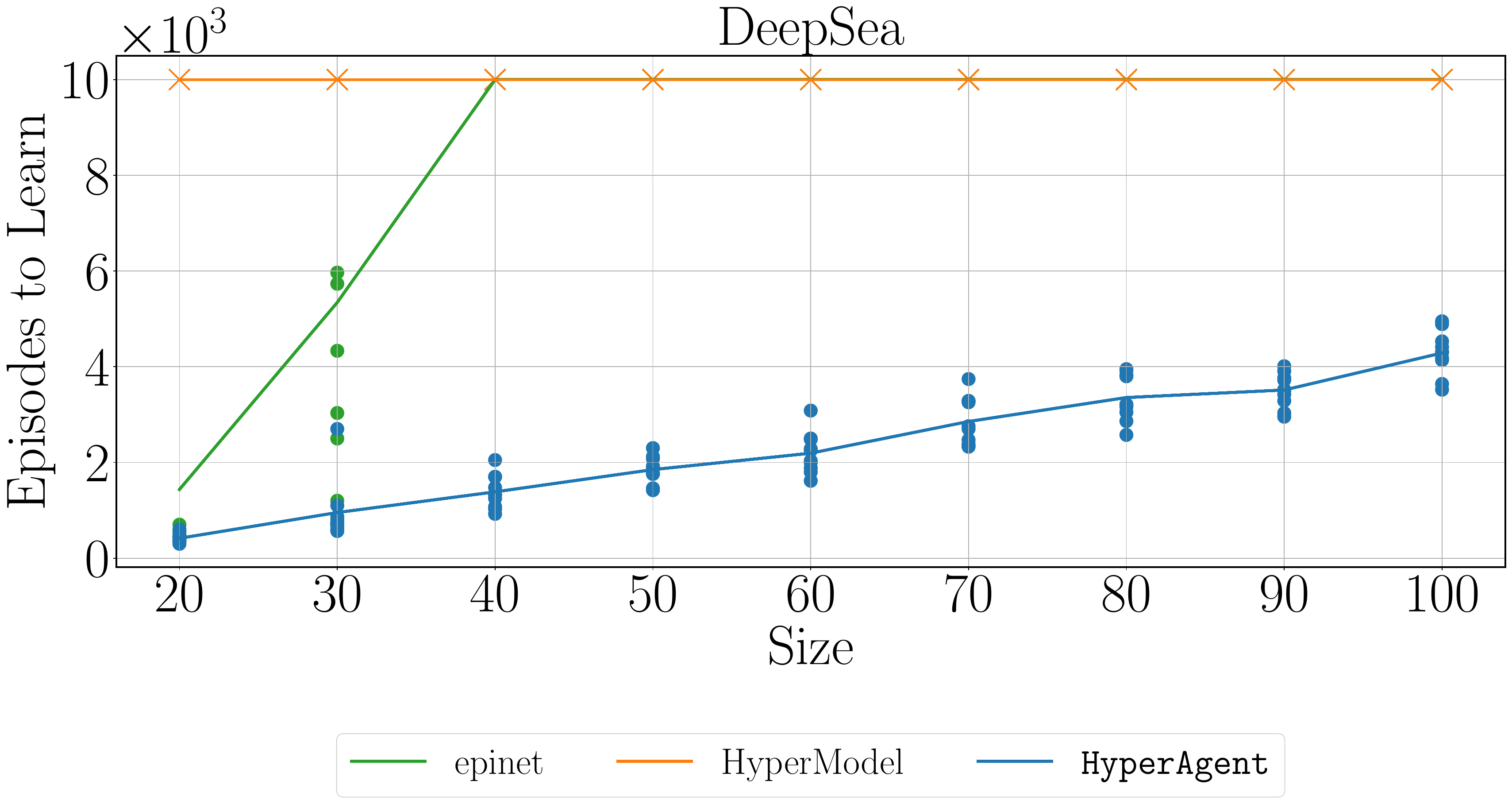}
    \caption{Results from various network structures using the hypermodel framework. \HyperAgent demonstrates superior data and computational efficiency compared to other methods.}
    \label{fig:deepsea_net}
\end{figure}

The comparison results are presented in Figure~\ref{fig:deepsea_net}. HyperModel is unable to solve DeepSea, even with a size of 20, while ENNDQN cannot solve DeepSea when the size exceeds 30.
Overall, \HyperAgent demonstrates superb efficiency and scalability, as it efficiently solves DeepSea instances of size $100^2$ that previous literature has never achieved.

\begin{figure}[htbp]
    \centering
    \includegraphics[width=0.6\linewidth]{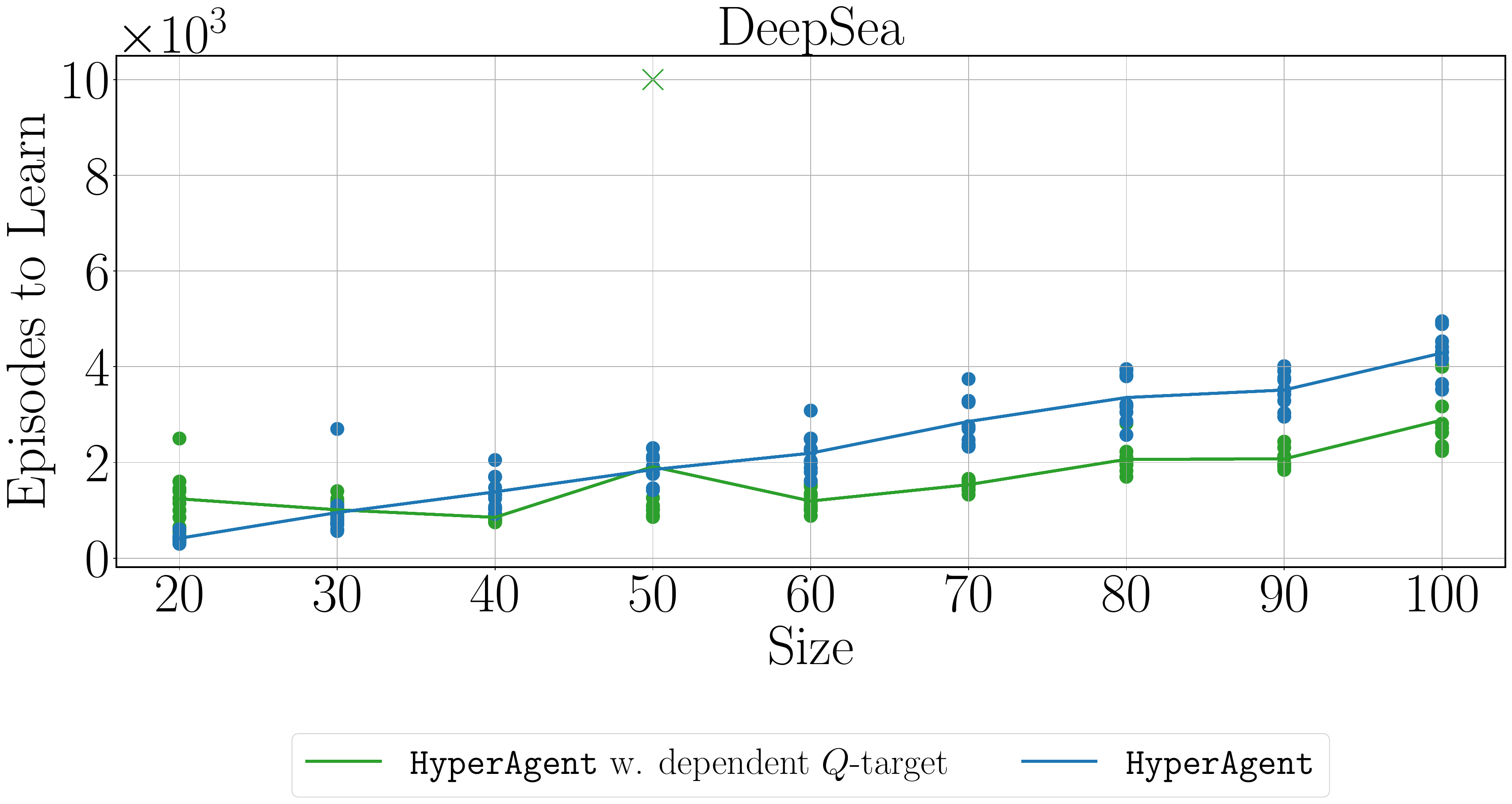}
    \caption{Ablation results on different $Q$-target computation methods. The method of \textit{dependent $Q$-target computation} can achieve better results inspiring future research.}
    \label{fig:deepsea_de}
\end{figure}

\paragraph{Ablation experiment on $Q$-target computation.}
Several methods, including HyperDQN, ENNDQN, and Ensemble+, employ the injected index to achieve posterior approximation and utilize the same index to compute the $Q$-target  as described in~\cref{appendix:difference}. 
They apply the same index for both main $Q$-value and target $Q$-value computation, which we refer to as \textit{dependent $Q$-target computation}. 
In contrast, \HyperAgent employs independent $Q$-target computation, where it independently samples different indices to compute the target $Q$-value, a strategy supported by theoretical analysis.
We contrast these two $Q$-target computation methods within the \HyperAgent framework and introduce the \HyperAgent w. dependent $Q$-target. 
As depicted in Figure~\ref{fig:deepsea_de}, \HyperAgent w. dependent $Q$-target demonstrates improved results, prompting further analysis of the underlying reasons in future research.

\clearpage
\section{Additional Results on Atari}
\label{appendix:additional_res_atari}
We demonstrated the efficiency of \HyperAgent in handling data and computation in Section~\ref{sec:exp_atari}.  Here, we present comprehensive results for each environment to further establish the superiority of our approach.

\paragraph{Detailed results on each game.}
We adhere to the evaluation protocol described in~\cref{appendix:eval_protocol} to obtain the best score achieved in each game with 2M steps of \HyperAgent, as presented in~\cref{tab:atari_score}.
The sources of the compared baseline results can also be found in~\cref{appendix:repo_baselines}.

\begin{table}[htbp]
    \centering
    \begin{adjustbox}{width=0.95\columnwidth}
        \begin{tabular}{l|rr|rrrrr}
            \toprule[1.2pt]
            Game           & Random  & Human   & DDQN$^\dag$ & DER     & Rainbow  & HyperDQN & \HyperAgent          \\
            \midrule \midrule
            Alien          & 227.8   & 7127.7  & 722.7   & 1642.2  & 1167.1   & 862.0    & \textbf{1830.2}   \\
            Amidar         & 5.8     & 1719.5  & 61.4    & 476.0   & 374.0    & 140.0    & \textbf{800.4}    \\
            Assault        & 222.4   & 742.0   & 815.3   & 488.3   & 2725.2   & 494.2    & \textbf{3276.2}   \\
            Asterix        & 210.0   & 8503.3  & 2471.1  & 1305.3  & 3213.3   & 713.3    & 2370.2            \\
            BankHeist      & 14.2    & 753.1   & 7.4     & 460.5   & 411.1    & 272.7    & 430.3             \\
            BattleZone     & 2360.0  & 37187.5 & 3925.0  & 19202.5 & 19379.7  & 11266.7  & \textbf{29399.0}  \\
            Boxing         & 0.1     & 12.1    & 26.7    & 1.7     & 69.9     & 6.8      & \textbf{74.0}     \\
            Breakout       & 1.7     & 30.5    & 2.0     & 6.5     & 137.3    & 11.9     & 54.8              \\
            ChopperCommand & 811.0   & 7387.8  & 354.6   & 1488.9  & 1769.4   & 846.7    & \textbf{2957.2}   \\
            CrazyClimber   & 10780.5 & 35829.  & 53166.5 & 36311.1 & 110215.8 & 42586.7  & \textbf{121855.8} \\
            DemonAttack    & 152.1   & 1971.0  & 1030.8  & 955.3   & 45961.3  & 2197.7   & \textbf{5852.0}   \\
            Freeway        & 0.0     & 29.6    & 5.1     & 32.8    & 32.4     & 30.9     & 32.2              \\
            Frostbite      & 65.2    & 4334.7  & 358.3   & 3628.3  & 3648.7   & 724.7    & \textbf{4583.9}   \\
            Gopher         & 257.6   & 2412.5  & 569.8   & 742.1   & 4938.0   & 1880.0   & \textbf{7365.8}   \\
            Hero           & 1027.0  & 30826.4 & 2772.9  & 15409.4 & 11202.3  & 9140.3   & 12324.7           \\
            Jamesbond      & 29.0    & 302.8   & 15.0    & 462.1   & 773.1    & 386.7    & \textbf{951.6}    \\
            Kangaroo       & 52.0    & 3035.0  & 134.9   & 8852.3  & 6456.1   & 3393.3   & 8517.1            \\
            Krull          & 1598.0  & 2665.5  & 6583.3  & 3786.7  & 8328.5   & 5488.7   & 8222.6            \\
            KungFuMaster   & 258.5   & 22736.3 & 12497.2 & 15457.0 & 25257.8  & 12940.0  & 23821.2           \\
            MsPacman       & 307.3   & 6951.6  & 1912.3  & 2333.7  & 1861.1   & 1305.3   & \textbf{3182.3}   \\
            Pong           & -20.7   & 14.6    & -15.4   & 20.6    & 5.1      & 20.5     & 20.5              \\
            PrivateEye     & 24.9    & 69571.3 & 37.8    & 900.9   & 100.0    & 64.5     & 171.9             \\
            Qbert          & 163.9   & 13455.0 & 1319.4  & 12345.5 & 7885.3   & 5793.3   & 12021.9           \\
            RoadRunner     & 11.5    & 7845.0  & 3693.5  & 14663.0 & 33851.0  & 7000.0   & 28789.4           \\
            Seaquest       & 68.4    & 42054.7 & 367.6   & 662.0   & 1524.7   & 370.7    & \textbf{2732.4}   \\
            UpNDown        & 533.4   & 11693.2 & 3422.8  & 6806.3  & 39187.1  & 4080.7   & 19719.2           \\
            \toprule[1.2pt]
        \end{tabular}
    \end{adjustbox}
    \caption{The averaged score over 200 evaluation episodes for the best policy in hindsight (after 2M steps) for 26 Atari games. The performance of the random policy and the human
        expert is from dqn\_zoo~\citet{dqnzoo2020github}.}
    \label{tab:atari_score}
\end{table}

\begin{figure}[htbp]
    \centering
    \includegraphics[width=0.8\linewidth]{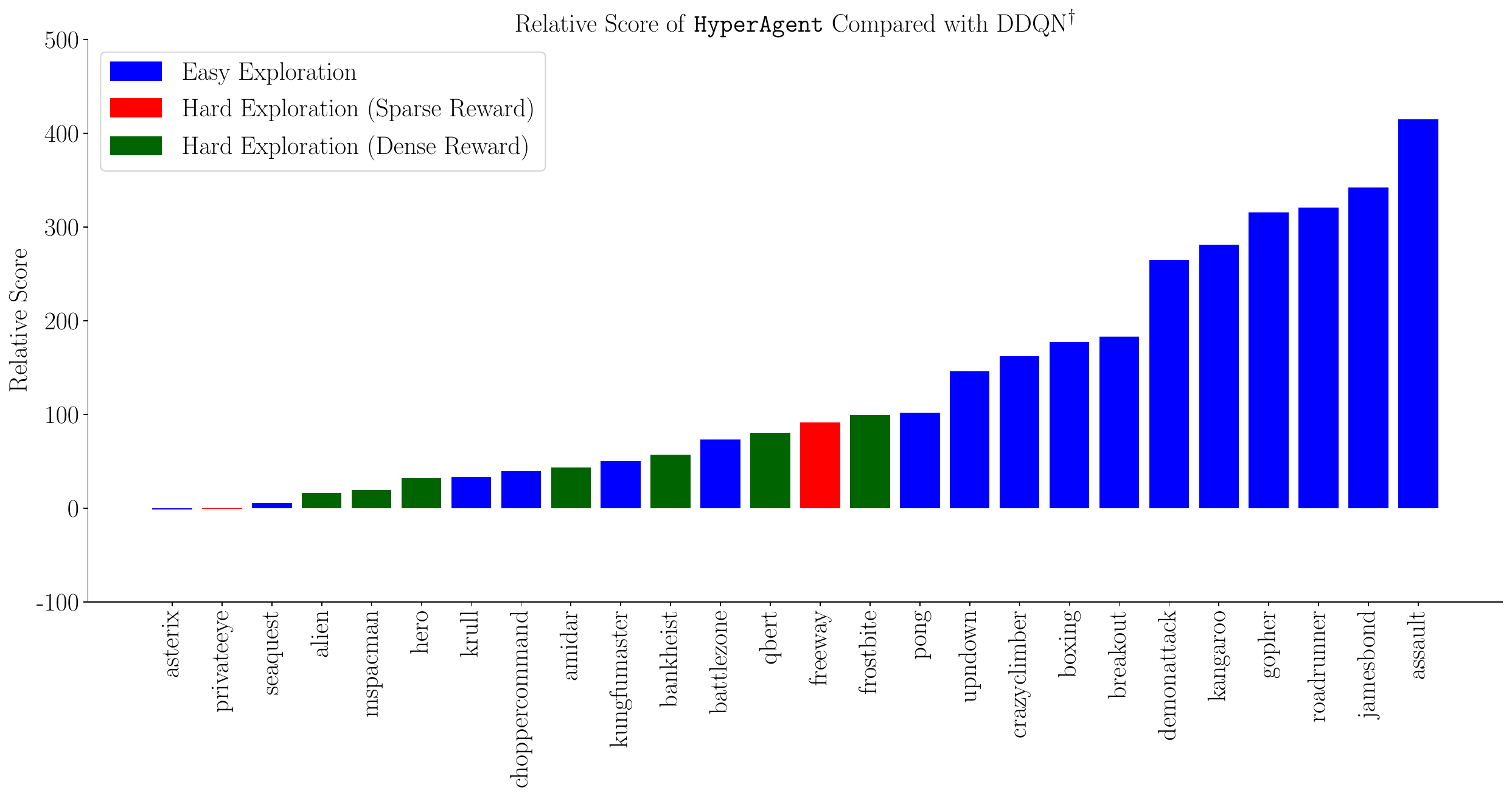}
    \caption{Relative improvement of \HyperAgent compared with DDQN$^\dag$}
    \label{fig:relative_hyperfqi_doubledqn}
\end{figure}

\begin{figure}[htbp]
    \centering
    \includegraphics[width=0.8\linewidth]{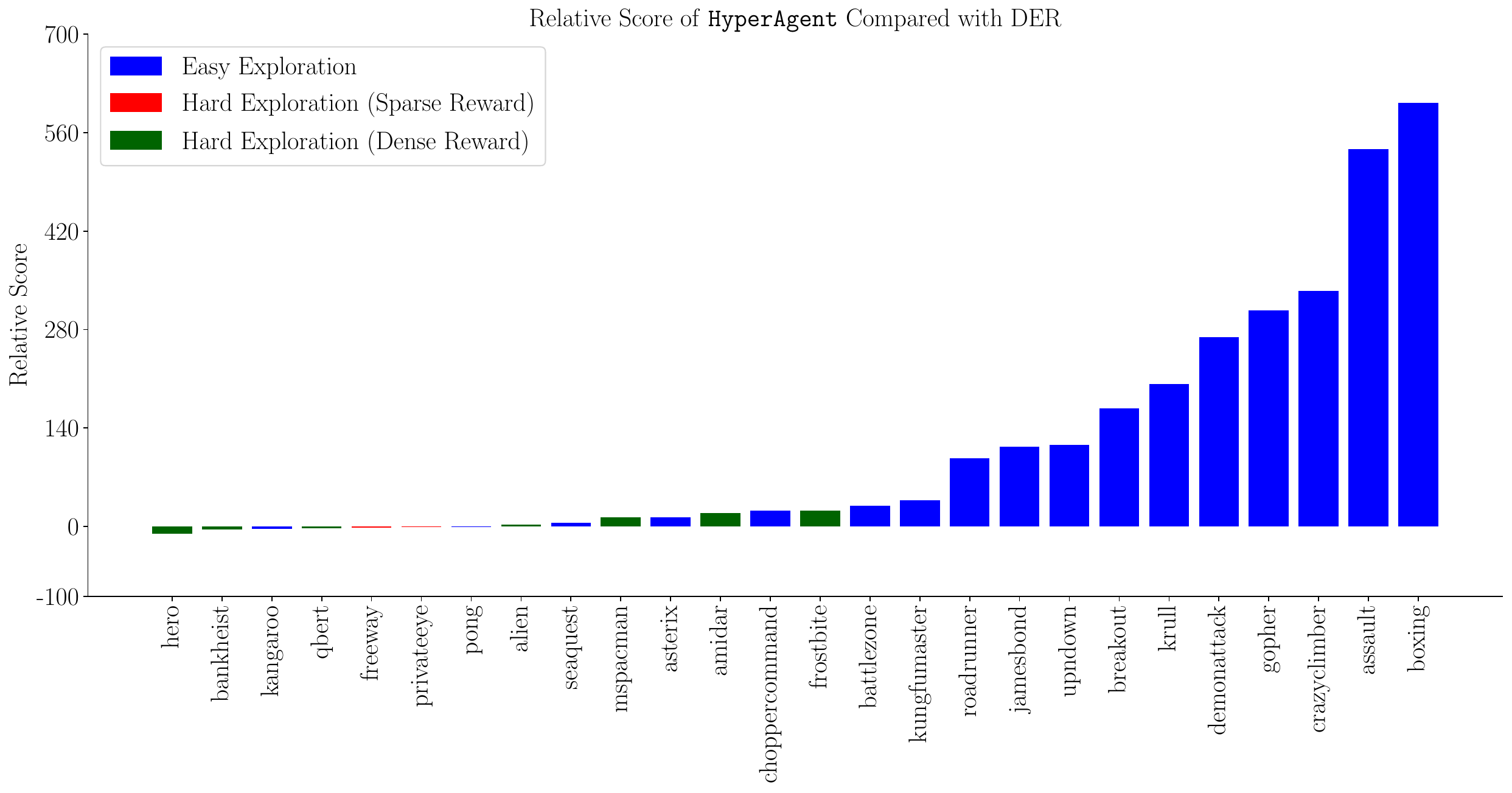}
    \caption{Relative improvement of \HyperAgent compared with DER}
    \label{fig:relative_hyperfqi_der}
\end{figure}

\begin{figure}[htbp]
    \centering
    \includegraphics[width=0.8\linewidth]{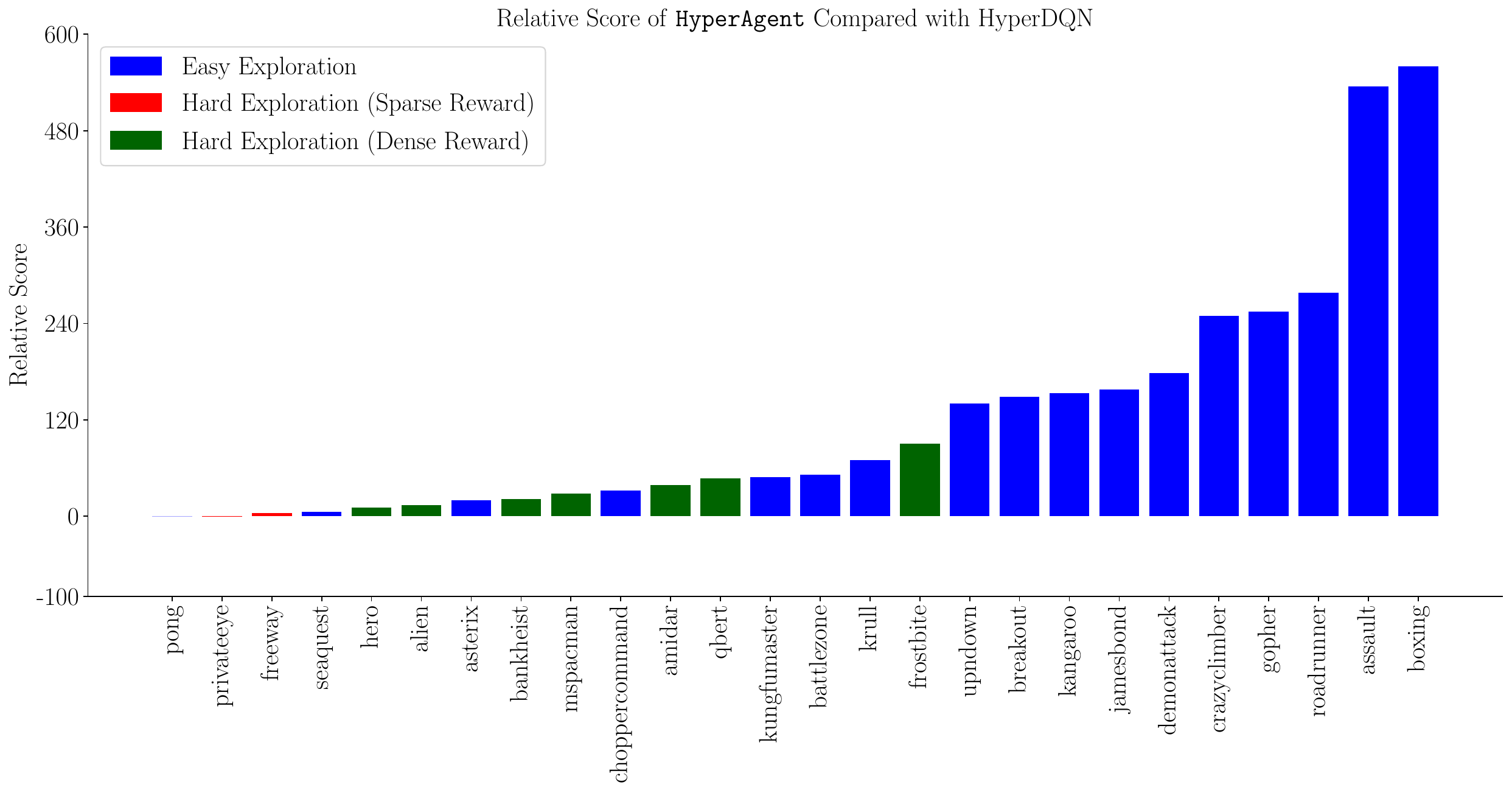}
    \caption{Relative improvement of \HyperAgent compared with HyperDQN}
    \label{fig:relative_hyperfqi_hyperdqn}
\end{figure}

\begin{figure}[htbp]
    \centering
    \includegraphics[width=0.8\linewidth]{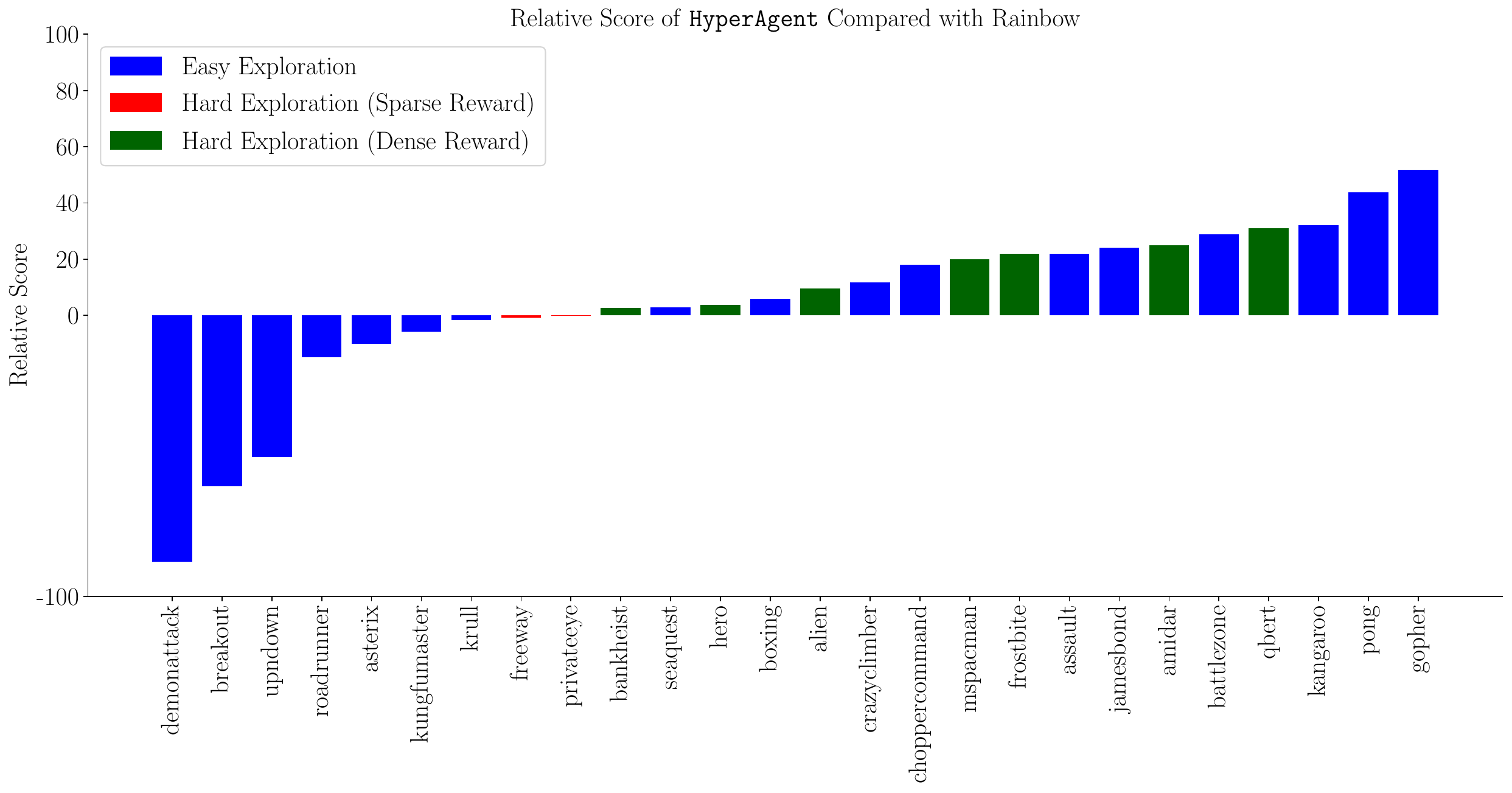}
    \caption{Relative improvement of \HyperAgent compared with Rainbow}
    \label{fig:relative_hyperfqi_rainbow}
\end{figure}

We also present the relative improvement of \HyperAgent in comparison to other baselines for each game, which is determined by the given following equation as per~\citep{wang2016dueling}:
$$\text{relative improvement} = \frac{\text{proposed} - \text{baseline}}{\max(\text{human}, \text{baseline}) - \text{human}}.$$
Our classification of environments into three groups, namely ``hard exploration (dense reward)", ``hard exploration (sparse reward)" and ``easy exploration", is based on the taxonomy proposed by~\cite{bellemare2016unifying}. The overall results are illustrated in~\cref{fig:relative_hyperfqi_doubledqn,fig:relative_hyperfqi_doubledqn,fig:relative_hyperfqi_der,fig:relative_hyperfqi_hyperdqn,fig:relative_hyperfqi_rainbow}.

\begin{figure}[htbp]
    \centering
    \includegraphics[width=0.85\linewidth]{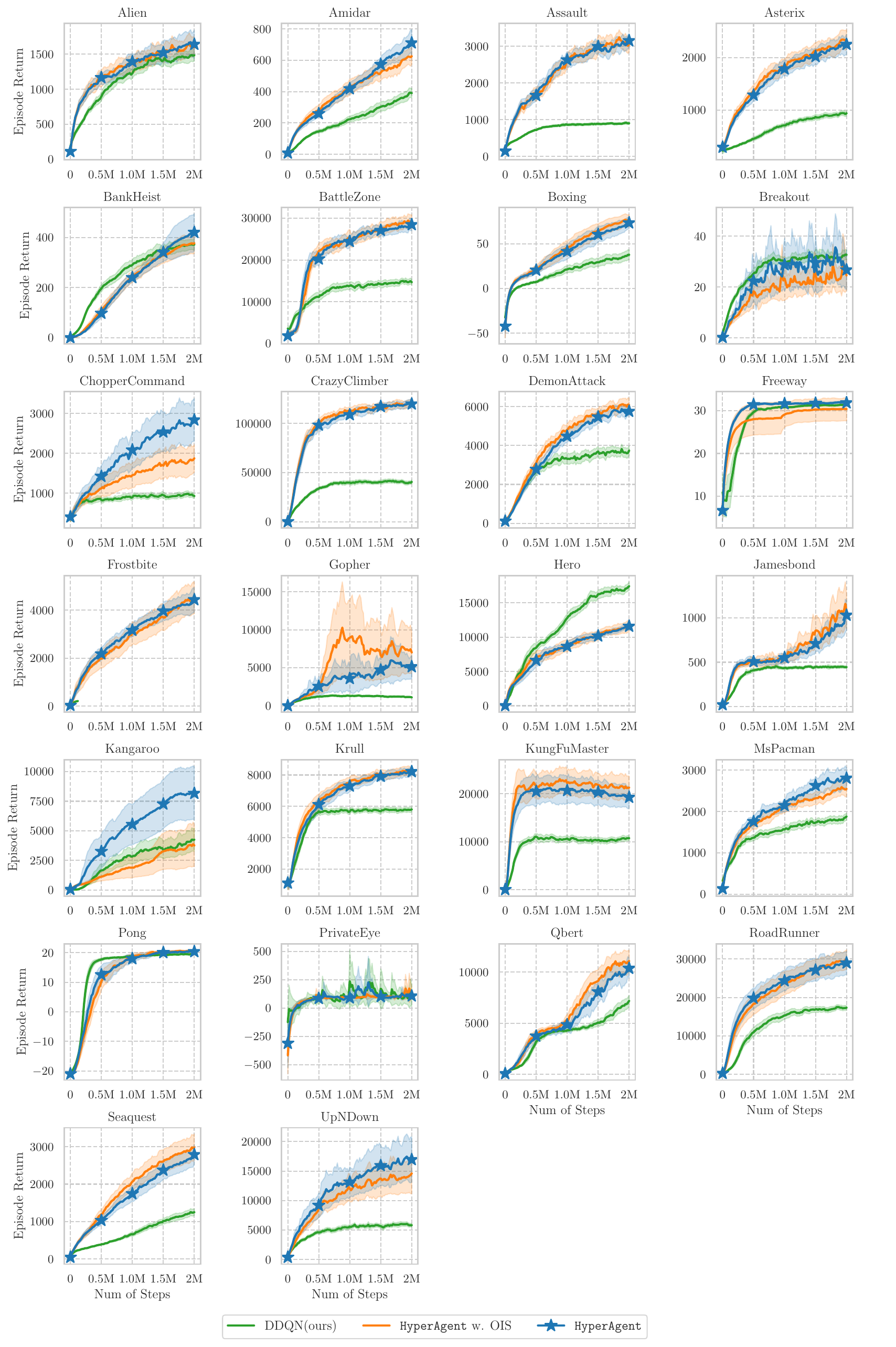}
    \caption{Learning curve for each game. 
    }
    \label{fig:atari_curve}
\end{figure}

\HyperAgent algorithm exhibits significant improvement compared to DDQN$^\dag$, DER, and HyperDQN in environments with ``easy exploration", and overall it performs better in all environments. This indicates that \HyperAgent has better generalization and exploration abilities. On the other hand, when compared to Rainbow, our algorithm performs better in environments which are in the group of ``hard exploration (dense reward)", demonstrating its superior exploration capabilities. 
In the case of Freeway, which belongs to the ``hard exploration (sparse reward)" group, both \HyperAgent and Rainbow achieve similar optimal scores (as shown in Table~\ref{tab:atari_score}). However, \HyperAgent demonstrates faster convergence, as evidenced in Figure~\ref{fig:atari_hardest}. Overall, \HyperAgent showcases better generalization and exploration efficiency than other baselines.

We also evaluate the OIS method across the 26 Atari games, as illustrated in Table~\ref{tab:atari_variant}. Our findings indicate that the OIS method does not generate significant differences in complex networks like Convolutional layers.

\begin{table}[htbp]
    \centering
    \begin{tabular}{lccc}
        \toprule[1.2pt]
        Method               & IQM               & Median            & Mean              \\
        \midrule \midrule
        \HyperAgent        & 1.22 (1.15, 1.30) & 1.07 (1.03, 1.14) & 1.97 (1.89, 2.07) \\
        HyperAgent w. OIS    & 1.15 (1.09, 1.22) & 1.12 (1.02, 1.18) & 2.02 (1.91, 2.16) \\
        \toprule[1.2pt]
    \end{tabular}
    \caption{Comparable results achieved using the OIS method in Atari games. The data in parentheses represent the 95\% confidence interval.}
    \label{tab:atari_variant}
\end{table}

Furthermore, we present the learning curve for each game in Figure~\ref{fig:atari_curve}. It is evident that \HyperAgent has demonstrated superior performance compared to DDQN(ours), attributed to the integration of hypermodel that enhances exploration.
Moreover, it is worth highlighting that the learning curve of \HyperAgent continues to rise in specific environments, indicating that it can achieve even better performance with additional training.

\begin{figure}[htbp]
    \centering
    \includegraphics[width=0.98\linewidth]{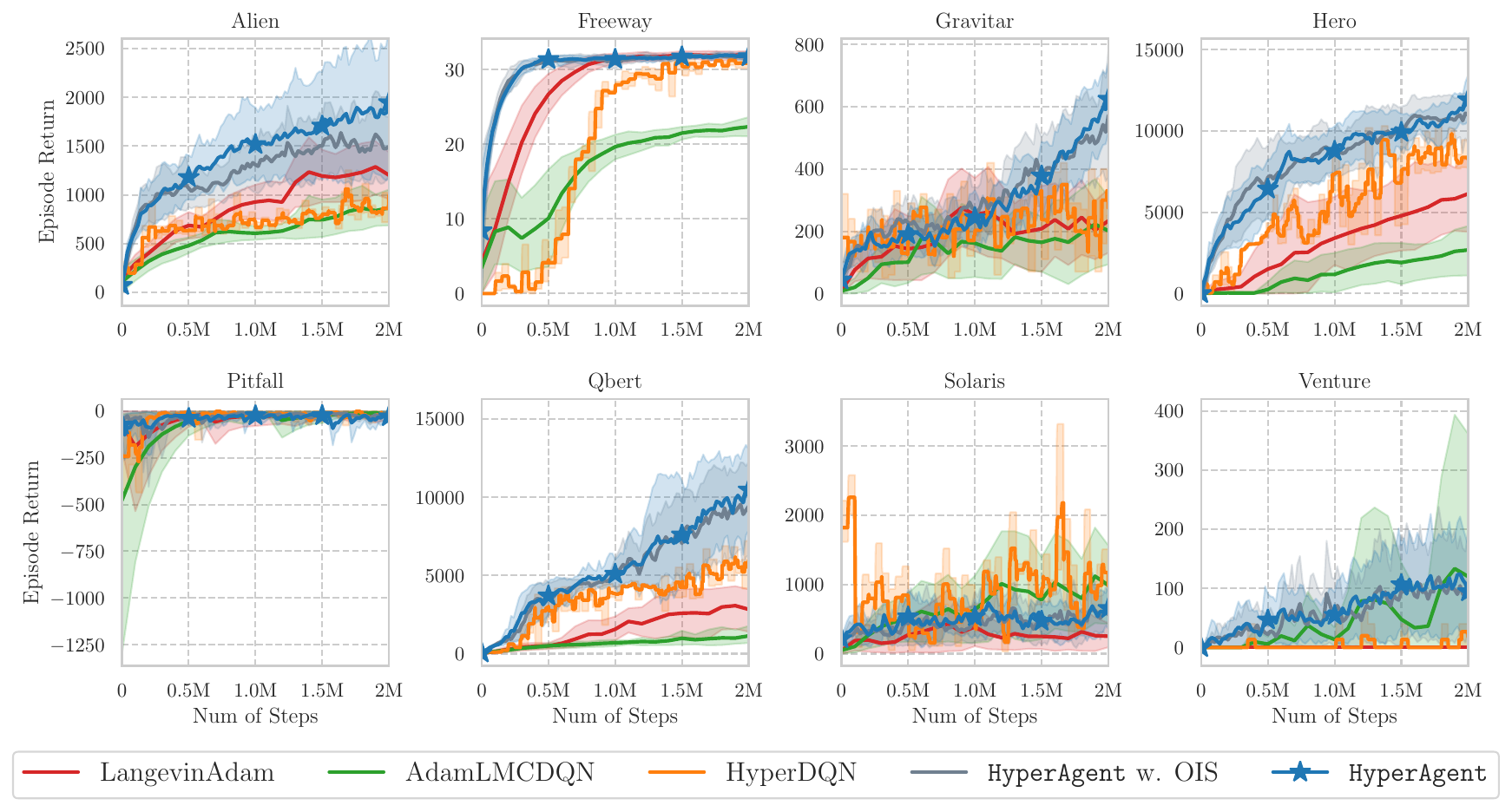}
    \caption{Comparative results on 8 hardest exploration games with more baselines.}
    \label{fig:atari_hardest_more}
\end{figure}

\paragraph{Additional results about exploration on Atari.}
To further demonstrate the exploration efficiency of \HyperAgent, we compare it with additional baselines, including AdamLMCDQN~\citep{ishfaq2024provable}, LangevinAdam~\citep{dwaracherla2020langevin}, HyperDQN~\citep{li2022hyperdqn} and our variant HyperAgent w. OIS.
As depicted in Figure~\ref{fig:atari_hardest_more}, both HyperDQN and \HyperAgent demonstrate improved results, indicating that applying hypermodel can lead to better posterior approximation and consequently enhance exploration.

\paragraph{Additional results on other 29 Atari games.}
To further showcase the robustness and scalability of \HyperAgent, we conduct experiments on all 55 Atari games using identical settings (see Table~\ref{tab:atari_setting}). 
We present the best score achieved in these 29 environments using 2M steps in Table~\ref{tab:atari_score_other}, with 5 seeds repeated for each environment.
\HyperAgent outperforms others in over half of these 29 environments, delivering top performance in 31 out of 55 Atari games.

\begin{table}[htbp]
    \centering
    \begin{adjustbox}{width=0.95\columnwidth}
        \begin{tabular}{l|rr|rrrrr}
            \toprule[1.2pt]
            Game           & Random  & Human   & DDQN$^\dag$    & Rainbow  & HyperDQN & \HyperAgent          \\
            \midrule \midrule
            Asteroids         & 210.0   & 47388.7 & 520.8 & 969.4 & 1044.7        & \textbf{1176.6}       \\
            Atlantis          & 12850.0 & 29028.1 & 5771.3        & 528642.7      & 370160.0      & \textbf{793851.9}       \\
            BeamRider         & 363.9   & 16926.5 & 464.4 & 10408.0       & 981.9 & 6749.2       \\
            Berzerk           & 123.7   & 2630.4  & 566.6 & 822.7 & 336.7 & \textbf{840.5}   \\
            Bowling           & 23.1    & 160.7   & 12.6  & 27.7  & 14.8  &   \textbf{61.3} \\
            Centipede         & 2090.9  & 12017.0 & 4343.2        & 5352.1        & 2071.5        &  4121.9 \\
            Defender          & 2874.5  & 18688.9    & 2978.4    & 25457.1   & 4063.3        &  21422.7   \\
            DoubleDunk        & -18.6   & -16.4 & -21.2 & -2.2  & -17.7 &   \textbf{-1.7}          \\
            Enduro            & 0.0     & 860.5 & 338.7 & 1496.6        & 201.0 &  1223.0   \\
            FishingDerby      & -91.7   & -38.7 & -78.1 & -22.4 & -74.3 & \textbf{-0.7} \\
            Gravitar          & 173.0   & 3351.4        & 2.6   & 372.0 & 300.0 & \textbf{629.5}     \\
            IceHockey         & -11.2   & 0.9      & -12.4 & -7.0  & -11.6 &     \textbf{-3.8}          \\
            NameThisGame      & 2292.3  & 8049.0   & 5870.3        & 9547.2        & 3628.0        &  5916.4    \\
            Phoenix           & 761.4   & 7242.6   & 3806.2        & 6325.0        & 3270.0        &   4941.9  \\
            Pitfall           & -229.4  & 6463.7        & -55.5 & -0.1  & -13.0 & -11.2          \\
            Riverraid         & 1338.5  & 17118.0       & 3406.2        & 5627.9        & 4233.3        &  \textbf{6896.3}    \\
            Robotank          & 2.2     & 11.9    & 7.8   & 22.4  & 3.1   &     \textbf{37.2}      \\
            Skiing            & -17098.1& -4336.9       & -22960.7      & -16884.8      & -29975.0      &   \textbf{-11654.4}      \\
            Solaris           & 1236.3  & 12326.7       & 390.2 & 1185.9        & 1173.3        & 941.3       \\
            SpaceInvaders     & 148.0   & 1668.7        & 356.1 & 742.2 & 425.3 & \textbf{762.2}  \\
            StarGunner        & 664.0   & 10250.0       & 346.3 & 11142.3       & 1113.3        & 6135.7     \\
            Tennis            & -23.8   & -8.3  & -10.1 & 0.0   & -17.0 & -1.4    \\
            TimePilot         & 3568.0  &5229.2        & 2204.6        & 3763.8        & 3106.7        & \textbf{6006.1} \\
            Tutankham         & 11.4    & 167.6   & 108.6 & 104.1 & 93.0  & \textbf{116.2}            \\
            Venture           & 0.0     & 1187.5        & 21.5  & 0.0   & 26.7  & \textbf{163.9}     \\
            VideoPinball      & 16256.9 & 17667.9       & 10557.5       & 49982.2       & 24859.2       & 29674.8        \\
            WizardOfWor       & 563.5   & 4756.5        & 275.7 & 3770.9        & 1393.3        & \textbf{4019.5}            \\
            YarsRevenge       & 3092.9  & 54576.9       & 10485.5       & 10195.6       & 4263.1        & \textbf{28805.5}     \\
            Zaxxon            & 32.5    & 9173.3        & 2.1   & 7344.1        & 3093.3        & \textbf{7688.1}          \\
            \toprule[1.2pt]
        \end{tabular}
    \end{adjustbox}
    \caption{The evaluated score of other 29 games from ALE suite. 
    }
    \label{tab:atari_score_other}
\end{table}

\end{document}